# DEEL
## DEpendable & Explainable Learning

White Paper

# Machine Learning in Certified Systems

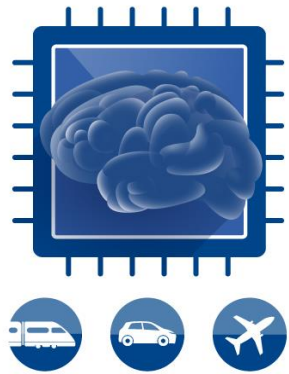

DEEL Certification Workgroup
IRT Saint Exupéry
March 2021
Ref - S079L03T00-005

**Contributors and affiliations**

- Hervé Delseny, Christophe Gabreau, Adrien Gauffriau (Airbus)
- Bernard Beaudouin, Ludovic Ponsolle (Apsys)
- Lucian Alecu, Hugues Bonnin (Continental)
- Brice Beltran, Didier Duchel, Jean-Brice Ginestet, Alexandre Hervieu, Ghilaine Martinez, Sylvain Pasquet (DGA)
- Kevin Delmas, Claire Pagetti (ONERA)
- Jean-Marc Gabriel (Renault)
- Camille Chapdelaine, Sylvaine Picard (Safran)
- Mathieu Damour (Scalian)
- Cyril Cappi, Laurent Gardès (SNCF)
- Florence De Grancey, Eric Jenn, Baptiste Lefevre (Thales)
- Gregory Flandin, Sébastien Gerchinovitz, Franck Mamalet, Alexandre Albore (IRT Saint Exupéry)

*The authors would like to thanks all those who contributed to the discussions and acculturation during the workshops of the ML Certification Workgroup, and particularly…*

The DEEL project core team : Francois Malgouyres, Jean-Michel Loubes (IMT), Edouard Pauwels, Mathieu Serrurier (IRIT), David Bertoin, Thibaut Boissin, Mikaël Capelle, Guillaume Gaudron, Rémi Verdier [on leave], David Vigouroux (IRT), Mélanie Ducoffe, Jayant Sen-Gupta (Airbus), Pierre Blanc-Paques (Airbus-DS), Thomas Soumarmon (Continental), Frederic Boisnard, Bertrand Cayssiols, Raphael Puget (Renault), Jérémy Pirard [On leave], Agustin Martin Picard (Scalian), Quentin Vincenot (Thales Alénia Space).

External invited speakers : Eric Poyet (Continental), Bertrand Iooss (EdF) , Fabrice Gamboa (IMT), Jérémie Guiochet (LAAS), Arnault Ioualalen (Numalis), Yann Argotti, Denis Chaton (Renault).

Occasional participation: Guillaume Vidot (Airbus), Gilles Dulon, James Bezamat (Safran), Jean-Christophe Bianic (Scalian).

The white paper has been realised in the frame of the DEEL[1] research program under the aegis of IRT Saint Exupéry. We would like to thank the industrial and academic members of the IRT who supported this project with their financial and knowledge contributions.

We would also like to thank the Commissariat Général aux Investissements and the Agence Nationale de la Recherche for their financial support within the framework of the Programme d'Investissement d'Avenir (PIA).

This project also received funding from the Réseau thématique de Recherche Avancée Sciences et Technologies pour l'Aéronautique et l'Espace (RTRA STAe).

---

[1] DEEL is an international research program on Dependable and Explainable Learning. It is operated by IVADO, IRT Saint Exupéry, CRIAQ, and ANITI. It benefits from the support of Commissariat Général aux Investissements (CGI) and the Agence Nationale de la Recherche (ANR) through the French Programme d'Investissement d'Avenir (PIA), the support of Conseil de Recherche en Sciences Naturelles et en Génie du Canada (CRSNG), as well as the MEI, Ministère de l'économie et de l'innovation du Québec. Visit https://www.deel.ai.

**Document history**

| Version | Modifications | Author | Date |
|---------|---------------|--------|------|
| 1.0 | Creation | Certification Workgroup | 2020/06 |
| 2.0 | All sections | Certification Workgroup | 2021/02 |

*Version 2 of this document has been improved significantly thanks to the careful reviews of: Lauriane Aufrant (DGA), Martin Avelino (Airbus), Guy Berthon (Thales), Calixte Champetier, Alexis de Cacqueray (Airbus), Emmanuelle Escorihuela (Airbus), Louis Fabre (Airbus), Rodolphe Gelin (Renault), Bertrand Iooss (EDF), Frédéric Jurie (Safran), Jean-Claude Laperche (Airbus), Juliette Mattioli (Thales), Emmanuel Pratz (Airbus), Alain Rossignol, Guillaume Soudain (EASA), Nicolas Valot (Airbus).*

*The content of this white paper does not engage the responsibility of the reviewers or theirs companies.*

# DEEL

**DEpendable & Explainable Learning**

## DEpendable & Explainable Learning
### DEEL PROJECT

IRT Saint Exupéry and ANITI in Toulouse, IVADO and CRIAQ in Montreal, have joined forces to build with their partners a "startup laboratory" on the theme of explainable and dependable artificial intelligence: DEEL (DEpendable & Explainable Learning).

It brings together manufacturers from the aeronautics, space, automotive and academic sectors, both French and Canadian.

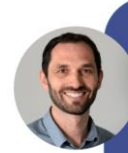
**GRÉGORY FLANDIN**
Program Director Artificial Intelligence for Critical Systems
Head of DEEL
gregory.flandin@irt-saintexupery.com
+336 77 05 68 70

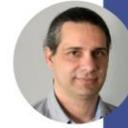
**FRANCK MAMALET**
Machine Learning Researcher
Head of Certification Working Group
franck.mamalet@irt-saintexupery.com

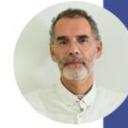
**ERIC JENN**
Research Engineer Embedded Systems
Head of Certification Working Group
eric.jenn@irt-saintexupery.com

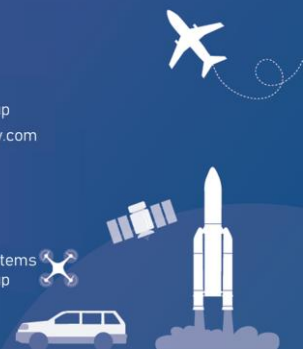

## ABOUT DEEL

DEEL is a team of 60 people, with a wide range of profiles including internationally renowned academic researchers, data scientists and thematic engineers specialized in operational safety and certification. This integrated team is located in Toulouse and Québec, with the objective of responding to scientific challenges of an industrial inspiration.

**It enables manufacturers to secure their critical systems by obtaining guarantees on the performance of artificial intelligence functions.**

## OBJECTIVES

- To create internationally recognized, visible, and identified scientific and technological knowledge on the explainability and robustness of artificial intelligence algorithms in an automatic and statistical learning context.
- To create conditions within the project itself for the rapid transfer of work to its industrial partners through the development of skills and the provision of tools.
- To make the scientific community and the general public aware of the emblematic scientific challenges built with the partners.

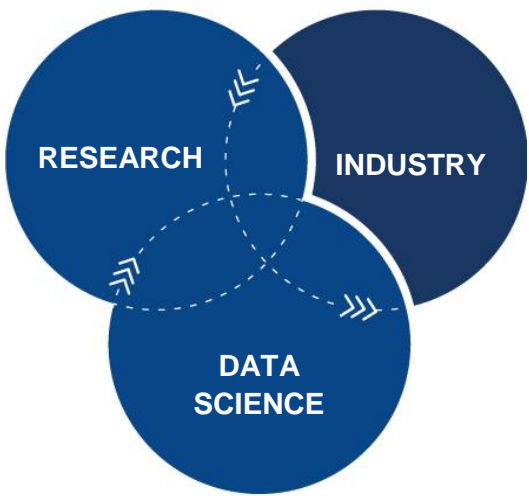

RESEARCH — INDUSTRY — DATA SCIENCE

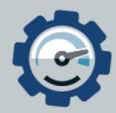 **START UP** Sept. 2018   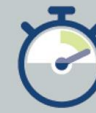 **DURATION** 5 years

**ACTUAL BUDGET** +€30M    **27 PARTNERS**

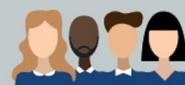 **60** ENGINEERS, EXPERTS, GEEKS, RESEARCHERS & PHDS

IVADO · IRT SAINT EXUPÉRY · CRIAQ · ANITI · Institut intelligence et données

# Table of Contents

> DISCLAIMER: This White Paper about "Machine Learning in Certified Systems" presents a snapshot of a work that is still in progress. Faced to the span and complexity of the problem, we have had to make choices among the topics to develop, to bound the depth of the analysis, and to leave out some "important details". Nevertheless, we consider that, even though they are preliminary, the results given in this paper are worth being shared with the community.

# 1 Introduction

## 1.1 THE ML CERTIFICATION WORKGROUP

Machine Learning (ML) seems to be one of the most promising solution to automate partially or completely some of the complex tasks currently realized by humans, such as driving vehicles, recognizing voice, etc. It is also an opportunity to implement and embed new capabilities out of the reach of classical implementation techniques. However, ML techniques introduce new potential risks. Therefore, they have only been applied in systems where their benefits are considered worth the increase of risk. In practice, ML techniques raise multiple challenges that could prevent their use in systems submitted to certification[2] constraints. But what are the actual challenges? Can they be overcome by selecting appropriate ML techniques, or by adopting new engineering or certification practices? These are some of the questions addressed by the ML Certification[3] Workgroup (WG) set-up by the Institut de Recherche Technologique Saint Exupéry de Toulouse (IRT), as part of the DEEL Project.

In order to start answering those questions, the ML Certification WG has decided to restrict the analysis to *off-line supervised learning techniques*, i.e., techniques where the learning phase completes before the system is commissioned (see §2.1). The spectrum of ML techniques is actually much larger, encompassing also unsupervised learning and reinforcement learning, but this restriction seems reasonable and pragmatic in the context of certification. The scope of the analysis is more precisely defined in §1.3.

The ML Certification WG is composed of experts in the fields of certification, dependability, AI, and embedded systems development. Industrial partners come from the aeronautics[4], railway,[5] and automotive[6] domains where "trusting a system" means, literally, accepting "to place one's life in the hand of the system". Some members of our workgroup are involved in both the AVSI (Aerospace Vehicle Systems Institute), the SOTIF (Safety Of The Intended Functionality) projects, and the new EUROCAE (European Organisation for Civil Aviation Equipment) WG-114 workgroup.

Note that the DEEL ML Certification workgroup is definitely not the only initiative dealing with trust and machine learning. However, it is singular for it leverages on the proximity of the members of the DEEL "core team" that gathers, at the same location, researchers in statistics, mathematics and AI coming from partner laboratories (IMT,[7] IRIT,[8] LAAS[9]), and specialists and data scientists coming from the IRT and the industrial partners.

---

[2] We use the term "certification" in a broad sense which does not necessarily refer to an external authority.
[3] In this document, "ML certification" will be used in place of "Certification of systems hosting ML techniques"
[4] Represented by Airbus, Apsys, Safran, Thales, Scalian, DGA, and Onera.
[5] Represented by SNCF.
[6] Represented by Continental and Renault.

## 1.2 OBJECTIVES OF THIS WHITE PAPER

This White Paper targets the following objectives:

- Sharing knowledge
- Identifying challenges for the certification of systems using ML
- Feeding the research effort.

### 1.2.1 Sharing knowledge

Standards like DO-178C/ED-12C, EN50128, or ISO 26262 have been elaborated carefully, incrementally, and consensually by engineers, regulators and subject matter experts. We consider that the same approach must apply to ML techniques: the emergence of new practices must be "endogenous", i.e., it must take roots in the ML domain, and be supported by data science, software, and system engineers. In order to facilitate this emergence, one important step is to share knowledge between experts of different industrial domains (automotive, aeronautics, and railway). This is a prerequisite to create a common understanding of the certification stakes and ML technical issues.

### 1.2.2 Identifying challenges for the certification of systems using ML

Once a common ground has been established between experts, the next step is to identify challenges considering conjointly the characteristics of ML techniques and the various certification objectives. To identify why ML techniques are *that different* from classical techniques (i.e., non-ML) from the perspective of certification is one of the objectives of this White Paper. Nevertheless, two significant differences are worth being mentioned right now. First, ML algorithms, based on learning, are fundamentally stochastic whereas classical solutions are (essentially) deterministic. Second, the behaviour of ML components is essentially determined by data, whereas the behaviour of classical algorithms is essentially determined by instructions. These differences and their impacts on our capability to certify systems are further elaborated in the rest of this document.

### 1.2.3 Feeding the research effort

The physical and organizational proximity of the "core team" of the DEEL project with the workgroup is a singularity compared to other initiatives. In practice, members of the core team, i.e., data scientists and mathematicians from industry and academia, participate actively to the workgroup meetings. Those discussions feed AI and Embedded Systems researchers with accurate and sharp challenges directly targeted towards certification objectives.

## 1.3 SCOPE OF THE DOCUMENT

The scope of this document is defined by the following restrictions:

- **We only consider off-line supervised learning** (see §2.1 for details).

---

[7] Institut de Mathématique de Toulouse : https://www.math.univ-toulouse.fr/.
[8] Institut de Recherche en Informatique de Toulouse : https://www.irit.fr/.
[9] Laboratoire d'Analyse et d'Architecture des Systèmes : https://www.laas.fr/public/.

- Focus is on ML subsystems and algorithms. System level considerations are only addressed in section "Resilience" (§4.3).
- We consider the use of ML algorithms in systems of all criticality levels, including the most critical ones.
- We consider systems with or without human in the loop.
- Issues related to implementation, i.e., how ML models are realized in software, hardware or a combination thereof are not specifically addressed in the document. However, we do know that they raise various concerns about, for instance, the effects of floating-point computations, the determinism of end-to-end response times, etc.
- Issues related to maintainability are also marginally addressed.
- The study is neither restricted to a specific domain nor to a specific certification framework. If focus is somewhat placed on large aeroplanes when considering the aeronautical domain, results are applicable to all other domains of aviation including rotary wing aircrafts, UAVs, general aviation, Air Traffic Management (ATM) systems, or other ground systems such as simulators.
- Cybersecurity is not addressed.

<u>Important notice on terminology</u> (Fault, Error and Failure): Fault and errors are given different definitions according to the domain and the references, but they agree on the fact that they are the more or less direct cause of a system failure, i.e., a situation where the system does not longer perform its intended function. In this white paper, and according to Laprie [1], we will mainly use the term "fault" to designate "the adjudged or hypothesized cause of a modification of a system state that may cause its subsequent failure". However, in paragraphs where this causality chain is not the main topic, we may use faults and errors interchangeably (or the term "fault/error"). Besides, the word "error" has another signification in theory of Machine-Learning, and specific footnotes will point out sections where this definition holds.

## 1.4 ORGANIZATION OF THE DOCUMENT

Section 1 is a brief introduction to the document.

Section 2 provides a general overview of the context of ML certification. It introduces the Machine Learning techniques and details the industrial needs and constraints.

Section 3 describes the analyses performed by the workgroup, in order to identify the ML certification challenges.

Section 4 presents the main ML certification challenges from an industrial and academic point of view. Note that subsections 4.2 to 4.8 have been written by different subgroups and can be read independently.

Section 5 concludes this White Paper and summarizes the main challenges identified.

# 2 Context

## 2.1 MACHINE LEARNING TECHNIQUES

"Machine learning is a branch of artificial intelligence (AI) […] which refers to automated detection of meaningful patterns in data. [It covers] a set of techniques that can "learn" from experience (input data)." [2]. Machine Learning uses techniques coming from the fields of AI, statistics, optimization, and data sciences.

For the categories of Machine Learning (ML) techniques considered in this document, "experience" comes under the form of a set of "examples" compiled into a "training dataset". In practice, learning provides a system with the capability to perform the specific tasks illustrated by those "examples". Usually, ML is used to perform complex tasks difficult to achieve by means of classical, procedural, techniques, such as playing go, localizing and recognizing objects on images, understanding natural languages, etc.

Figure 1 gives a simple example of such a task: the ML component classifies its inputs (characterized by their features) based on a parameterized ML model. The objective of the ML approach is to learn these parameters in order to be able to take (or infer) decisions.

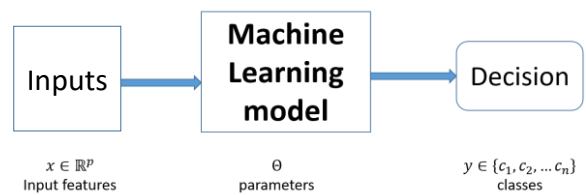

*Figure 1. Using ML to perform a classification operation*

Machine Learning covers a large variety of applications, methods, and algorithms. In the rest of this section, we will first give some elements of classification of ML techniques, according to the way data are collected (the *protocol*), and the learning strategies. We present a short taxonomy of the main Machine Learning algorithms and finish with an overview of intrinsic errors[10] introduced by ML approaches.

### 2.1.1 Elements of classification of ML techniques
#### 2.1.1.1 Protocol for data collection

The protocol used to collect data is the first important criterion of classification. Many criteria, such as the type of data sources, the type of sampling methods, etc., should be taken into account to precisely characterize a protocol. However, for the sake of simplification, we will consider here only two main dimensions:

- **Passive vs active acquisition**, i.e., data acquisition independent (resp. dependent) of the ML decisions taken during the learning phase.
- **Offline vs online learning**, i.e., learning relying on a "frozen" training set, or learning relying on data acquired during operation.

#### 2.1.1.2 Learning strategies

In the literature, we commonly find four types of learning strategies:

---

[10] In this paragraph, the term error has no link with Laprie's definitions given in §1.3. It corresponds to the standard definition in Machine Learning theory.

- **Supervised learning:** With a labelled training dataset $S = \{x_i, u_i\}$ $x_i \in X$ (input space), and $u_i \in U$ (decision space), the system learns a function $h_\theta$, of parameters $\theta$, as close as possible to the target function, or distribution. $U$ is discrete for classification problems and continuous for regression or density estimation problems.
- **Semi-supervised learning:** The system learns using both a small amount of labelled data $\{x_i, u_i\}$ and a large amount of unlabeled data $\{x'_i\}$ in the training dataset S.
- **Unsupervised learning:** The system learns underlying structure of data using an unlabeled dataset $S = \{x_i\}$.
- **Reinforcement learning:** The system learns by interacting with its environment, in order to maximize long-term reward. Exploration of the input space is determined by the actions taken by the system (active acquisition).

In this document, we focus on **passive offline** and **supervised learning**, as it seems to be a reasonable first step towards certification.

### 2.1.2 Taxonomy of Machine Learning techniques

Table 1 gives a short list of the most common ML techniques. It indicates their main advantages and drawbacks. Each kind of ML technique will rely on one or several hypothesis function space(s) ($h_\theta$), and one or several exploration algorithms (not listed in this document) to minimize a loss function on the training dataset (more details and mathematical formulations in §4.2.3). Much literature can be found on each technique, and for readers that need more details, we can advise to refer to [2].

*Table 1. Taxonomy of ML techniques*

| Techniques | Applications | Pros | Cons |
|---|---|---|---|
| **Linear models**: Linear and logistic regressions, support vector machines | classification, regression | Good mathematical properties (proof), easily interpretable | Linear constraints induce a poor performance in complex or high dimension problems |
| **Neighbourhood models**: KNN, K means, Kernel density | classification, regression, clustering, density estimation | Easy to use | Relies on a distance that can be difficult to define in high dimension problems |
| **Trees**: decision trees, regression trees | classification, regression | Fast, applicable on any type of data | Unstable, unable to cope with missing data |
| **Graphical models**: naive Bayesian, Bayesian network, Conditional Random Fields (CRF) | classification, density estimation | Probabilistic model (confidence in outputs), cope with missing data | Difficult to extend to continuous variables; complex to train |
| **Combination of models**: Random Forest, Adaboost, gradient boosting (XGboost) | classification, regression, clustering, density estimation | Efficient, with any type of data | Lots of parameters, not efficient on high dimension data (such as image) |
| **Neural networks (NN)**, Deep Learning (DNN) | classification, regression | Easily adaptable, effective on high-dimensional data | Many parameters, computation complexity |

### 2.1.3 Biases and errors in ML

In the previous sections, we have seen that in order to learn the ML model parameters for a given application, we will have to choose: the protocol, the input space and the dataset, the hypothesis function space ($h_\theta$), the exploration algorithm of $h_\theta$ space (local or global, gradient descent, etc.), and the loss function and performance metrics (error rate, mean square error, log-likelihood, etc.).

Each choice will introduce biases, but those biases are actually inherent to the learning process. Given these biases, the literature points out three types of errors[11] induced by a ML mechanism (see Figure 2):

- **Approximation error:** this error is the distance between the target function $f$ (unknown) and the closest function in the hypothesis function space $h_{\theta^*}$. This error is correlated to the choice of the hypothesis function.
- **Estimation/Generalization error:** this error is the distance between the optimal function $h_{\hat\theta}$ achievable given the training dataset and the closest function $h_{\theta^*}$. This error is correlated to the choice of the training dataset that is only a sample of the input space.
- **Optimization error:** this error is the distance between the function found by the optimization algorithm $h_{\theta'}$, and the optimal function $h_{\hat\theta}$. Given the dataset, the optimization algorithm might not reach the global optimal function $h_{\hat\theta}$ but only find a local minimum $h_{\theta'}$ (or just a stationary point).

Note that these errors, intrinsic to the ML process, will come in addition to more common errors such as input or rounding errors.

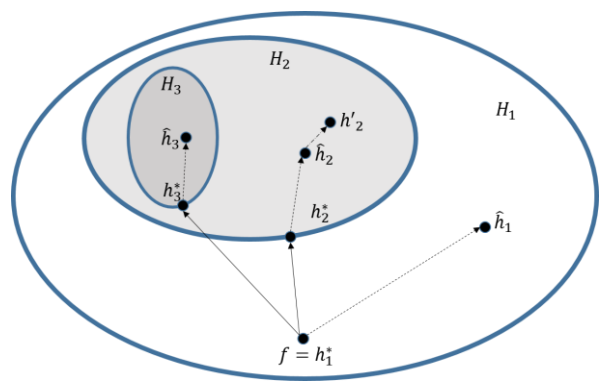

*Figure 2. Errors in Machine Learning: the H1 ellipse represents the set of all functions in which we can find the target (unknown function). H2 and H3 are examples of hypothesis function spaces (H2 is more expressive than H3). $h_i^*$ are the best candidates in each function space (approximation error), $\hat{h}_i$ are the best candidates given the training dataset (estimation error), and $h_i'$ are the elements learnt (optimization error).*

## 2.2 THE NEEDS FOR MACHINE LEARNING

After this quick introduction to Machine Learning techniques, and before addressing the certification of systems using those techniques, it may be worth saying a few words about why we are considering integrating ML in those systems.

---

[11] In this paragraph, the term error has no link with Laprie's definitions given in §1.3. It corresponds to the standard definition in Machine Learning theory. See §4.2.3 for more details and mathematical formulations.

Technically, ML solves problems that were generally considered intractable, such as the processing of natural language, the recognition of objects, and more generally the extraction of complex correlations in huge datasets and their exploitation in real-time. The spectrum of possible applications is huge, from giving autonomy to cars, aeroplanes, trains, or UAVs, offering new man-machine interaction means, to predicting failures, possibly leading to an improvement of the overall safety of systems. Let us briefly consider the needs for Machine Learning for automotive, railway, and aeronautics systems.

### 2.2.1 Automotive

Since its creation, the automotive industry has always been a fertile ground to create or integrate innovative solutions. Machine Learning techniques are no exception to the rule.

Data are now massively produced everywhere inside companies (engineering department, manufacturing, marketing, sales, customer services…). Combining these data and new Machine Learning techniques is a means to revisit existing processes, identify new needs, and create new products to satisfy them.

In the marketing field, ML techniques may be used to leverage information collected in showrooms in order to improve the knowledge about customers and better satisfy their needs. Such applications, having no impact on safety and a limited impact on business, are strong opportunities with low risks.

In the field of predictive maintenance, ML techniques are well suited to identify abnormal patterns in an ocean of data. In addition, for a given car, they can account for the driving conditions and the driver profile. Again, these techniques will probably not have a direct impact on the safety of drivers and passengers because safety rules regarding car maintenance remains applicable to avoid hazardous situations. Nevertheless, they can be useful to prevent customers from facing unpleasant situations.

In the manufacturing field, those techniques can also improve and reduce the cost of quality control. Until now, this control is done by a human operator who is in charge of identifying all potential manufacturing defects. Such defects may simply affect the aspect or the comfort of the vehicle, but they may also have a strong impact on its safety. This is for instance the case of the welding points of the car's body. Ensuring the conformity of welding point requires large efforts and costs due to the imperfection of human verification. Using a camera coupled with a machine-learning based computer vision algorithm to detect welding points not complying with specifications would thus be of great interest. In this case, guarantees must be given that the ML system is at least as performant as the human it replaces. Therefore, in the context of manufacturing, Machine Learning Algorithms could have a positive impact on global safety of the product if we can bring guarantees regarding their performance.

Last but not least, Machine Learning techniques conditions the emergence of Automated Driving.

Figure 3 presents the 5 levels of automation usually considered by the international automotive community.

Level 0, not shown on the figure, means "no automation".

Level 1 and 2 refer to what is usually called ADAS ("Advanced Driving Assistance Systems"), i.e. systems that can control a vehicle in some situations, but for which the driver is always supposed to be able, at any time, to take back control of the vehicle. Moreover, the driver remains fully responsible for driving the vehicle.

Starting from level 3, delegation for driving is given to the vehicle under certain circumstances, but at this level, the driver is still supposed to be able to take back control of the vehicle at any time, when required by the system, but in a defined delay.

At last, Levels 4 and 5 give more and more driving delegation to the vehicle without monitoring of driver on a specific use case (level 4) or on all use cases (level 5).

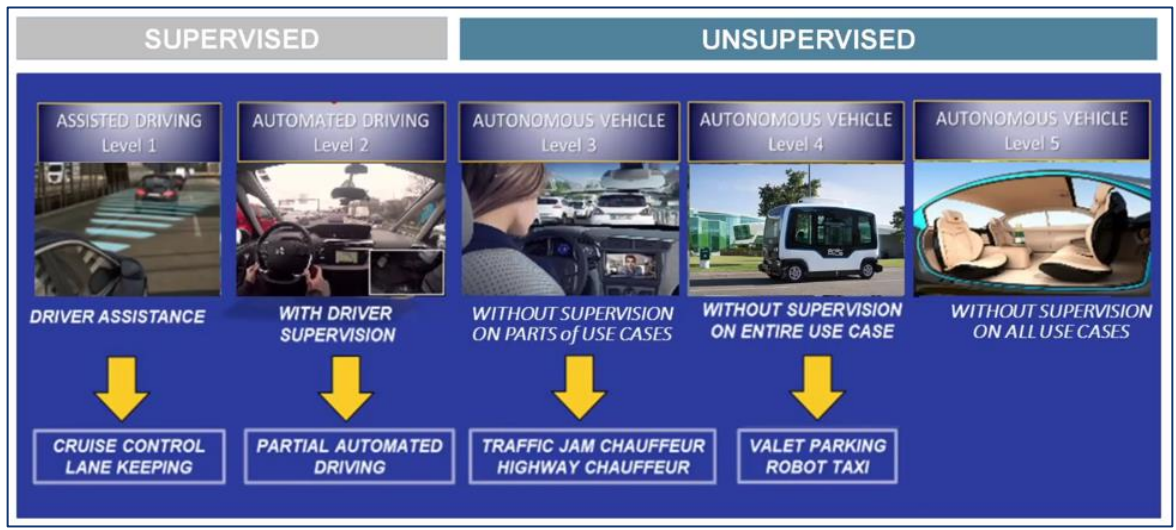

Figure 3. The five levels of automation (picture taken from the SIA conference « Les applications du véhicule autonome, nécessités et limites »)

Automated Driving (AD) is a well-known field of application for Machine Learning. In particular, Deep Learning (DL) and convolutional networks applied to computer vision have enabled to develop efficient camera-based perception systems and have contributed to the rise of ADAS and Autonomous driving applications. Thank to these techniques, the vehicle is now able to perceive its environment better than ever. However, there are still severe obstacles to reach the level of dependability required for a system that can take control of a motor vehicle.

As ADAS and Automated Level 1 and 2 functions impose that the driver remains responsible for driving, one could think that safety is not at stake for these levels of automation. This concerns, for instance, functions heavily depending on camera-based perception systems using ML techniques such as AEB (Advanced Emergency Braking System) that are now under regulation, ACC (Active Cruise Control), or LKA (Lane Keeping Assist). However, having dependable Machine Learning algorithms to reduce false positives and false negatives is already a very strong concern.

Nevertheless, the most critical situations are obviously observed for Levels 3 to 5. In these situations, driving responsibility being delegated to the vehicle, at least for a while and under certain circumstances, performance guarantees must be provided to demonstrate the safety of the automated function.

Providing such guarantees is not only a long-term concern. For instance, a first Level 3 system regulation has already been adopted in June 2020 at UNECE for the safe introduction of ALKS (Automated Lane Keeping Assist System) by 2021. Even if the condition of use are still restrictive ("[…] on roads where pedestrians and cyclists are prohibited and which, by design, are equipped with a physical separation that divides the traffic moving in opposite directions. […] operational speed of ALKS systems [limited] to a maximum of 60 km/h […]"[12]), this new regulation makes concrete the need for dependable Machine Learning and certification procedures.

Beyond the objective of reducing road casualties thanks to AD (human factor is the first contributor to road accidents) the benefits of this new technology are potentially large:

- Reducing traffic and pollution by creating dedicated traffic lanes for AD vehicles. Indeed, when associated with connectivity (between vehicles and infrastructures) AD can significantly reduce safety distance required between vehicles and so increase traffic flow.
- On demand Transport in Urban context.
- AD Shuttle service complementary to existing transport network.
- AD Shuttle in rural areas.
- Urban Logistics for last kilometer delivery.
- Logistics in confined areas (Airport, Port, Mines …).
- Valet Parking.
- etc.

From a technological perspective, ML in automotive concerns:

- **Sensor fusion.** The difficulty of integrating large amounts of heterogeneous, multi-source, data may be addressed by Machine Learning techniques. Examples range from fusing video data with radar, LiDAR, infrared, ultrasound, geographical, meteorological, etc. ML may also be used to develop so-called virtual sensors, *i.e.* numerical models that can reliably infer physical quantities from other sensor data. These virtual sensors can either replace or be used as backups of their physical counterparts, depending on whether the subject of cost or robustness is considered a priority.
- **Dimensionality reduction.** Selecting both the right (i.e. relevant) content and amount of information needed to be exchanged and processed from the vastness of available data, in order to detect or predict a certain state or object, is always a great concern for the data driven process, as it has a direct impact on the reliability of the system and its operational costs.
- **Time series forecasting.** As most of the vehicle-related data have embedded a strong temporal component, many problems can be formulated as (multi-variate) time series predictions. Forecasting examples include prediction of traffic jams, parking availability, most probable route, short-term speed horizon, road conditions according to weather, etc.
- **Anomaly detection.** Detecting abnormal cases or patterns in the data can highlight early errors or even prevent the occurrence of events with serious consequences. Anomaly detection is a hot and challenging research topic (usually due to the rare occurrence of the patterns) with a strong practical impact. Examples of applications range from fault detection in manufacturing processes to predictive

---

[12] See https://www.unece.org/info/media/presscurrent-press-h/transport/2020/un-regulation-on-automated-lane-keeping-systems-is-milestone-for-safe-introduction-of-automated-vehicles-in-traffic/doc.html

- maintenance or to the detection of abnormal driver behaviours (e.g. to ensure a required level of attention in semi-autonomous driving).
- **Optimization.** The objective is to optimize specific quantities, such as the time or costs of certain phases of the car assembling pipeline, the trip time or the delivery time (routing), the energy consumption of (individual or fleet of) vehicle(s), the pollutant emissions, etc.
- **Planning and reasoning.** Perceptual capacities concern almost exclusively of decision-making on very short time scales. They are a necessary yet not sufficient component of any intelligent system. Planning and reasoning are usually required to be implemented on the higher-levels of the perception-action loop of any proper autonomous agent. Historically, these problems have been addressed by symbolic AI techniques, expert systems, and planning theory. However, Machine Learning is expected to gain momentum in this area in the years to come.

From a purely technological point of view, we may also differentiate among the possible loci of the Machine Learning components, namely whether they are embedded, cloud-based, or a combination of the two (or other hybrid technologies).

### 2.2.2 Railway

Natural language processing, artificial vision, time series prediction and more generally artificial intelligence technologies are widely applied in non-critical systems in the railway industry (services, maintenance, and customer relations).

As a case study, Rail operations at SNCF comprise managing 15,000 trains a day before, during, and after their commercial service; defining timetables; rotating rolling stock and supervising personnel; providing travel information and recommending itineraries; and managing traffic flows into, out of, and at stations. With passenger and freight volumes continually growing, transport offers expanding to keep pace with demand, and infrastructures sometimes being at full capacity locally, operating a rail system is becoming increasingly complex. Travelers expectations are changing too: they want quicker responses, more personalized service, and greater reliability.

Under these circumstances, the operation of rail systems is evolving, and the use of artificial intelligence offers new ways of optimizing the capabilities and raising the operational performance of systems, and improving the overall quality of service.

AI is already used in various operations, such as real time traffic resource and passenger flows optimization, response to unforeseen events with projected evolution, decision-making based on predictive simulations, comparison and recommendation of solutions, etc. AI, and more specifically ML, is not meant to replace rail operators, but to help them making better and faster decisions by analyzing a much larger set of parameters than the human mind can process, with the required level of safety. For efficient interactions with operators, data visualization and new man/machine interfaces will be crucial.

ML techniques may also be used in safety critical systems. Safety critical tasks can be divided in two categories:

- Tasks performed by machines (signalling, speed control, train and network supervision and regulation, etc.).
- Tasks performed by human operators (driving, visual control of train before operation, checking various elements, etc.).

Today, tasks of the first category are performed by systems that do not rely on ML. Their behaviour is completely deterministic and certified by the European railway authorities. However, ML could be profitable here by giving the capability to process data near the sensors that produce them, thus avoiding network latency times and gaining in reactivity. Note that beyond the specific issues raised by ML, processing data closer to sensors and actuators will also have to face the classical problems of space constraints, electromagnetic compatibility issues (EMC), communication, etc.

The second category, safety critical human operator tasks, is perfectly covered by a number of detailed and precise regulations, and by a three-year safety cycle for operators. However, as humans are obviously not perfectly deterministic, ML may also be a way to raise the overall safety of the system.

So, ML can be used to improve the realization of both categories of tasks. However, the focus is currently mainly on human operator tasks that could be supported or replaced by systems embedding ML algorithms. Finally, this boils down to moving these tasks from the second to the first category, going from safety critical human task to safety critical automatic systems, with all the consequences on the certification process that this new kind of automated tasks may introduce.

The most obvious example of this trend is the autonomous train.

**Autonomous train**

Why making train autonomous? The main objective with autonomous trains is to operate more trains on existing infrastructure by optimizing speeds and enabling traffic to flow more smoothly. More trains on the same stretch of track means the possibility of transporting more people and goods. For example, with the extension of the RER E commuter line to Nanterre, in the western Paris region the frequency will increase from 16 to 22 trains an hour between Rosa Parks and Nanterre in 2022. Autonomous trains operating at optimal speeds and connected through a common control system, rail service will be more responsive and resilient to varying conditions. For customers, this will mean improved regularity and punctuality. Energy consumption will also decrease with autonomous trains. An onboard system will be able to calculate the most energy-efficient running for a given track route. Here, too, the impact will be very significant in view of the railway group's €1.3-billion annual energy bill.

Of course, autonomous trains will go into commercial service only when all conditions to ensure total safety have been met: this means a level of safety which will be at least as high, if not higher than, today's (i.e. GAME principle described in §2.3.2.1). Capacity, regularity, lower energy consumption: all these benefits will add up to improved competitiveness for rail transport.

Driverless railway mobility is not a new thing. The first automatic metro was introduced in the 80's and ATO (Automatic Train Operation) and now ATS (Automatic Train Supervision) systems are widely used all over the world. However, these systems fall into the first category described above: no ML inside and a system completely certified, end to end. This of course comes with clear constraints: the necessity to operate in a controlled environment, and the requirement of a completely new infrastructure and new trains.

However, the goal with the newest autonomous train projects is to achieve a new level of autonomy by dealing with open environment (obstacle detection for example is mandatory).

For some projects, this has to be done without any change in the network infrastructure. In this context, helping or replacing the driver shall be seen as a direct transfer of human tasks to the machine: the latter has to operate the train according to the external signals, it has to detect any abnormal behaviour, react to any unplanned event in environment, deal with weather conditions, etc.

To achieve these goals, railway industry needs to gain confidence on systems embedding ML techniques and find out ways to certify those systems.

### 2.2.3 Aeronautics

Most of the uses of Machine Learning described in the previous two sections are also relevant for Aeronautics. In this section, the specifics of aeronautics are highlighted, but the level of details is limited, in order to avoid redundancies with the previous sections.

In aeronautics, many functions are already automated: a modern aircraft can be automatically piloted in almost all phases of flight, even in adverse environmental conditions. Nevertheless, the pilot remains today the first decision-maker on board, and the last resort in case of abnormal situation.

Machine Learning (and more broadly Artificial Intelligence) has multiple potential applications in safety-critical aeronautical systems such as

- Support the pilots in their tasks.
- Provide new operational capabilities.
- Optimize aircraft functions by augmenting physical sensing (virtual sensors).
- Reduce aircraft operating costs thanks to ML-based health monitoring and predictive maintenance.
- Enable unmanned operations.
- Etc.

ML is indeed a powerful tool to augment the capabilities of a human. ML can help listen to the Air Traffic Control instructions, monitor the systems, and detect a conflict or an abnormal behaviour. As systems are getting more and more complex, ML can also help the pilot to analyse the situation, and can highlight the most relevant information. These capabilities can be classified in three main categories:

- **Perception** (computer vision, natural language processing), for instance to increase the situational awareness of the pilot.
- **Anomaly detection**, for instance to help the pilot identify any malfunction, failure, or discrepancy.
- **Recommendation and decision making** (planning, scheduling, reasoning), for instance to provide the pilot with options on how to conduct and optimize the flight.

These new capabilities can also be useful on the ground, where ML techniques could equally support the work of Air Traffic Controllers.

If these techniques prove to be effective, they could then pave the way to new operational capabilities, such as automated take-off, Single Pilot Operations (one pilot supported by a virtual copilot), and overall an increased level of automation. ML could also support the rise of Unmanned Aircraft Systems (UAS) operations, and, on a longer term, autonomous commercial air transport (including freight transport).

This evolution will not be limited to the commercial aviation. It will also have a significant effect on the general aviation (including Search and Rescue operations), military operations, etc.

## 2.3 CERTIFICATION PRACTICES

This section provides an overview of the current and upcoming certification practices in the automotive, railway, and aeronautical domains. This section does not intend to provide a detailed description of all applicable certification standards and regulations, but rather aims at enabling readers that are not familiar with certification to understand the basic principles of certification in each domain.

### 2.3.1 Automotive

#### 2.3.1.1 Overview

There is no certification for cars equipment but, when a new vehicle model is placed on the market, it must conform to a set of norms at the national and international levels. Besides this vehicle type certification, industrial standards are also adopted on a voluntary basis, at least for the commercial advantages that the adherence to these standards potentially offers. The main examples for our topic of interest are:

- **Euro-NCAP** (European New Car Assessment Program): it defines a list of safety equipment that the car may have. It delivers also a notation (expressed with "stars") to any new vehicle model depending on how much its equipment covered by this standard meet the requirements.
- **ISO 26262**: this standard "applies to all activities during the safety lifecycle of safety-related systems comprised of the hardware components (named E/E for Electrical/ Electronic) and the software components" of the automobile vehicle. The general characteristics of this standards are:
    - It proposes mitigation means against risks coming from E/E failures.
    - It considers systematic failures from software and hardware and random failures from hardware.
    - On the software part, the main principles are:
        - Strong management of requirements and their traceability.
        - Cascade of different levels of requirements.
        - High importance of the verification of the implementation of these requirements through different levels of test.
        - Configuration management of all elements of the software lifecycle.
    - However, software analysis differs since the functional requirements are differentiated from the safety ones, which implies to carry out an independence analysis ensuring their respective consistency.

#### 2.3.1.2 Towards new practices

During the ISO 26262 revision process (2016-2018), the necessity for a complementary approach to the component "failure-oriented approach" that was considered in the ISO 26262 emerged. The idea was that some limitations of an equipment could lead to the violation of some safety goals *without any failure in any system of the vehicle*.

These limitations concern the *intention of a function that is not achieved in some conditions*, not the realization of the function. Therefore, these limitations cannot be mitigated by the application of the ISO 26262, which is focused on the occurrences or consequences of *failures*.

To address this problem, a workgroup was formed to deal specifically with the *safety of the intended function (SOTIF)*. The result of this work is the new ISO standard ISO PAS 21448.

The SOTIF classifies operational scenarios according to their impact on the safety (safe/unsafe) and the *a priori* knowledge one may have concerning their occurrence in operation (known/unknown[13]). This is represented on Figure 4.

As any other methodology aimed at ensuring safety, the main objective of the SOTIF is to maximize or maintain the "safe areas", i.e. the proportion of operational scenarios leading to a safe situation, and minimize the "unsafe area", i.e. the proportion of operational scenarios leading to an unsafe situation while keeping a reasonable level of availability of the system under design.

The originality of the SOTIF lies in the fact that it addresses explicitly the case of the *unknown* and *unsafe* operational scenarios. It proposes an iterative approach to reduce the occurrence of these situations "as much as possible with an acceptable level of effort".

The SOTIF is complementary to the ISO 26262. The ISO 26262 proposes means to handle the design and random errors; the SOTIF extends the ISO 26262 by considering the effects of the "the inability of the function to correctly comprehend the situation and operate safely [… including] functions that use Machine Learning", and the possible misuse of the system.

The specific limitations of the ML algorithms, including their capability "to handle possible scenarios, or non-deterministic behaviour", typically fits in the scope of the SOTIF.

The SOTIF proposes a process to deal with those situations. In particular, it proposes an approach to verify the decision algorithms (see §10.3 of the SOTIF) and evaluate the residual risk. For instance, Annex C proposes a testing strategy of an Automatic Emergency Braking (AEB) system. An estimation of the probability of rear-end collision in the absence of AEB is obtained from field data. Considering that the probability of occurrence of the hazardous event shall be smaller than this probability, the number of kilometres of data to be collected (for the different speed limits) in order to validate the SOTIF can then be estimated. The process may be iterated up to the point where the estimated residual risk is deemed acceptable.

---

[13] We come back on the concept of known/unknown, known/known, etc. in Section 4.2.3.1.

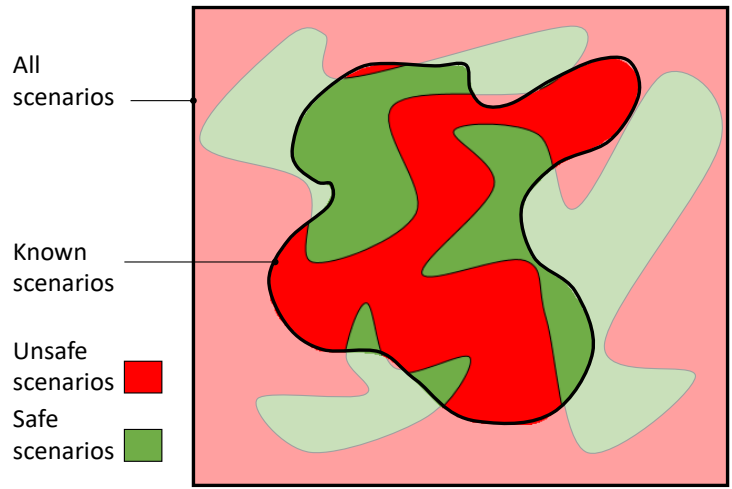

*Figure 4. Operational scenarios classification as per SOTIF*

This approach *allows but manages* the performance limitations inherent to ML. In particular, it explicitly allows unknown situations for which no learning data will be provided during the learning phase but provides a methodological framework to reduce them and their impact on safety.

### 2.3.2 Railway
#### 2.3.2.1 Overview

In accordance with European regulations, the railway system is broken down into *structural subsystems* (infrastructure, energy, control-command and signalling, *r*olling *stock)* and *functional* subsystems (operation and management of the traffic, including the organization, maintenance, including the organization).

The standards given in Table 2 are applicable to the hardware and software parts of systems/sub-systems/equipment. Before an application for Safety approval can be considered, an independent safety assessment of the system/sub-system/equipment and its Safety Case shall be carried out to provide additional assurance that the necessary level of safety has been achieved. The resulting Safety Assessment Report must explain the activities carried out by the safety assessor to determine how the system/sub-system/equipment (hardware and software) has been designed to meet its requirements. It can also specify some additional conditions for the operation of the system/sub-system/equipment.

In addition to the strict application of standards, European regulation 402/2013/EC requires railway operators to implement a "common safety method relating to the assessment and evaluation of risks" to all modification, whether technical, organizational or operational, of the railway system. The Common Safety Method (CSM[14]) describes how the safety levels, the achievement of safety targets and compliance with other safety requirements should be fulfilled.

---

[14] Common Safety Methods: the CSM is a framework that describes a common mandatory European risk management process for the rail industry and does not prescribe specific tools or techniques to be used

The risk acceptability of the system under assessment shall be evaluated by using one or more of the following risk acceptance principles:[15]

1. **Compliance with a recognised and context-specific code of practice** (see Table 3) to deal with conventional risks that are not necessary to explicitly re-identify if there is no technical or functional innovation or evolution of the environment and context.
2. **Comparison with a "similar reference system".**[16] This principle existed previously to the CSM in France as GAME (*Globalement Au Moins Equivalent*): it is possible to accept a new system if it is shown that the deviations from the existing system, having the same functionalities and operating in a similar environment to the system to which it is compared, do not introduce new risks or increase existing risks.
3. **Explicit risk analysis** to demonstrate quantitatively or qualitatively that the risk is kept low enough to be acceptable.

In the end, the Certification Authority is based on the opinion of an expert: This independent expert will evaluate the subject and the means of evidence for the safety demonstration in order to obtain the confidence. If this condition is not met, additional evidences may be required.

---

[15] They ensure that the objective of maintaining the level of security imposed by the legislation is respected. Article 5 of the 19/03/2012 French decree states: "Any changes concerning a system or under the system including the national rail network operated, such as the integration of a new system, implementation of new technologies or changes in the organization, procedures, equipment or sets of equipment included in rail infrastructure, rolling stock or operations, are carried out in such a way that the overall level of safety of the national rail network is at least equivalent to that existing before the development considered."

[16] The reference system may be a subsystem that exists elsewhere on the railway network, similar to that which is planned after the change, provided its functions and interfaces are similar to those of the sub-system evaluated and placed under similar operational and environmental conditions.

*Table 2. Simplified representation of requirements for railway certification*

| | Interoperability Directive (EU) 2016/797 | Safety Directive (EU) 2016/798 | |
|---|---|---|---|
| **System** | Set of TSI,[17] NNSR/TR[18] (Notified National Safety Rules / Technical Rules, EN 50126, EN 50128, EN 50129) | CSMs and NNSR/TR[18] (Notified National Safety Rules /Technical Rules), EN 50126, EN 50128, EN 50129) | Law |
| **HW/SW** | Standards called by TSI | 50126, 50128/50657, 50129 | Rules, Methods, Standards |
| **Mechanical** | Standards called by TSI | EN 15566, EN 15551, EN 16286-1, EN 15085-5, … | |
| **Fire** | Standards called by TSI | EN 45545 | |
| **Etc.** | … | … | |

*Table 3. Overview of the main railway standards*

| Requirements | Objectives |
|---|---|
| EN 50126-1:2017<br>EN 50126-2:2017 | "Railway Application – The Specification and Demonstration of Reliability, Availability, Maintainability and Safety (RAMS) – Part 1: Generic RAMS Process" & "Part 2: System Approach to Safety" |
| EN 50128:2011 | "Railway applications – Communication, signalling and processing systems – Software for railway control and protection systems" |
| EN 50657:2017 | "Railways Applications – Rolling stock applications – Software on Board Rolling Stock" |
| EN 50129:2018 | "Railway applications – Communication, signalling and processing systems – Safety related electronic systems for signalling" |
| EN 45545 (part 1 to 7) | "Railway applications – Fire protection on railway vehicles" |
| UIC 612<br>ISO 3864-1<br>EN 13272 | "DRIVER MACHINES INTERFACES FOR EMU/DMU, LOCOMOTIVES AND DRIVING COACHES" |
| EN 15355:2008<br>EN 15611:2008<br>EN 15612:2008<br>EN 15625:2008<br>UIC 544-1 oct 2004<br>EN 15595:2009 | "Railway applications – Braking" |
| … | … |

---

[17] Technical Specifications for Interoperability: TSI are specifications drafted by the European Railway Agency and adopted in a Decision or Regulation by the European Commission, to ensure the interoperability of the trans-European rail system

[18] Notified National Safety Rules /Technical Rules: NNSR/TR National rules refer to all binding rules adopted in a Member State, regardless of the body which issues them which contain railway safety or technical requirements, other than those laid down by Union or international rules, and which are applicable within that Member State to railway undertakings, infrastructure managers or third parties

### 2.3.2.2 Towards new practices

The CENELEC EN-50128 standard recognises that default prediction (trend calculation), defect correction, and maintenance, and monitoring actions are supported very effectively by Artificial Intelligence-based systems in various parts of a system. This does apply exclusively to SIL0 as it is not recommended for higher SIL (Safety Integrity Level) levels.

There is currently no specific work at the French and European level to introduce any breakthrough concerning AI certification. Instead, continuity of the process of publishing new TSIs is favoured. This is illustrated by the latest Loc & Pas TSI, which deals with "innovative solutions".

Indeed, to allow technological progress, it may be necessary to introduce innovative solutions not complying with defined specifications (i.e. that do not meet TSI specifications or to which TSI evaluation methods cannot be applied). In this case, new specifications and / or new evaluation methods associated with these innovative solutions must be developed and proposed.

Innovative solutions can involve the "Infrastructure" and "Rolling Equipment" subsystems, their parts and their interoperability components. If an innovative solution is proposed, the manufacturer (or his authorized representative established in the E.U.) lists the discrepancies with the corresponding provision of the TSI attached and submits it to the European Commission for analysis. Then, the Commission may consult with EU Agency for Railways and, if necessary, with relevant stakeholders. Finally, if the Commission approves the proposed innovative solution, the functional and interface specifications and evaluation method required to enable the use of this innovative solution are developed and integrated into the TSI during the process review. Otherwise, the proposed innovative solution cannot be applied.

Pending the revision of the TSI, the approval given by the Commission is seen as a statement of compliance with the essential requirements of the Directive (E.U) 2016/797 and can be used for the evaluation of subsystems and Projects.

## 2.3.3 Aeronautics

### 2.3.3.1 Overview

The certification framework for an airborne equipment that includes hardware and software is summarized in Figure 5. This Figure, which concerns large aeroplanes, is based on FAA and EASA processes but it would be similar for other countries and authorities.

From the applicable CS25 document (Certification Specification and Acceptable Means of compliance for Large Aeroplanes), there is a link to ED-79A/ARP4754A and ED-135/ARP4761 that provide a list of recommended practices for safety at function, aircraft, and system levels. For analysis at component level, depending of the system's nature, documents such as ED-80/DO-254 (for hardware), ED-12C/DO-178C (for software) and ED-14G/DO-160G (for environmental conditions) can be used.

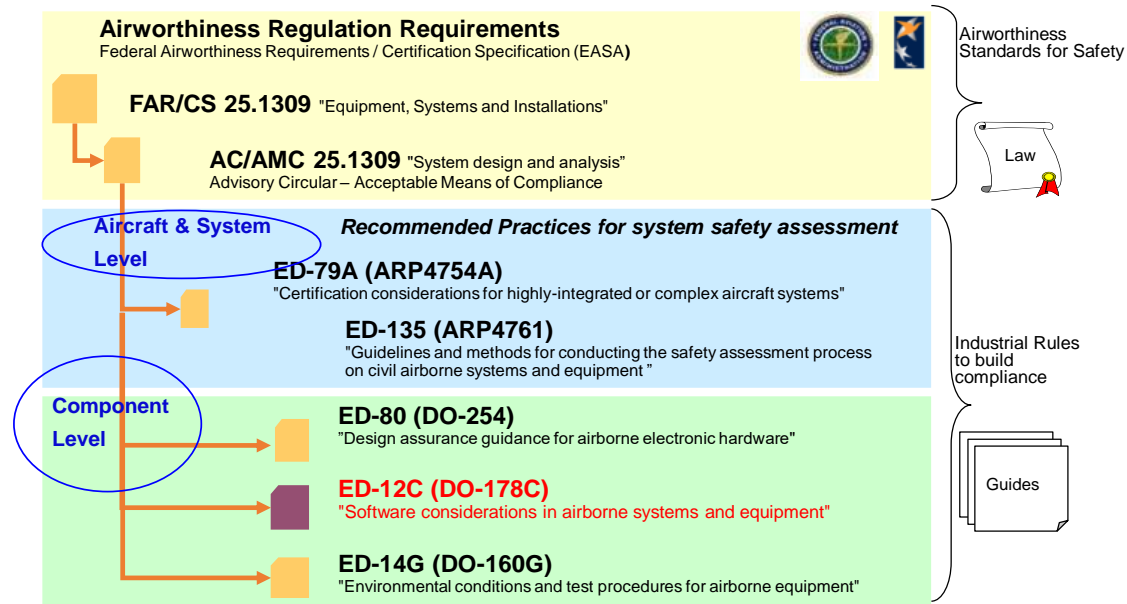

*Figure 5. Aeronautics certification framework: example of large aeroplanes (partial view)*

This framework includes several layers:

- **High-level airworthiness objectives** described in the certification regulations (in particular CS 25.1309 for any system or equipment onboard an aeroplane, but also CS 25.1301 for intended function and CS 25.1302 for human factors).
- **Detailed industry standards** recognized by the certification authorities as acceptable means of compliance to the objectives. In particular:
    - Recommended practices for system certification (ED-79A / ARP4754A and system safety assessment (ED-135 / ARP4761).
    - Development assurance standards for Hardware (ED-80 / DO-254) and Software (ED-12C / DO-178C).
    - Standard for environmental tests on airborne equipment (ED-14G / DO-160G).

Certification of airborne equipment that includes software and hardware in the aeronautical domain usually relies on the application of these regulations and standards. Although these regulations and standards are intended to be independent from any specific technology, they have been written at a time when Machine Learning was not as mature and promising as today. Consequently, they do not cover well some of the specificities of this technology, leading to the various challenges for certification identified later in this White Paper.

These standards overall rely on a typical development process for airborne equipment that includes the following steps:

- Aircraft function development.
- Allocation of aircraft function to system.
- System description and architecture.
- Safety assessment and allocation (Functional Hazard Assessments).
- Requirements capture (all requirements, including safety ones).
- Requirements validation.

- Implementation.
- Verification.
- Safety demonstration.
- System integration, verification and validation.

The ARP4754A standard also defines a development assurance process to establish confidence that a system has been developed in a sufficiently disciplined manner to limit the likelihood of development errors that could impact aircraft safety.

It must be noticed that before thinking about a system architecture, the analysis must start by the definition of the main functions and then the sub-level functions. Then, a Functional Hazard Assessment is realized (and later updated) in order to characterize the severity associated to the loss of a function or "Failure Condition". The severity, ranging from NSE "No Safety Effect" to CAT "Catastrophic", is used to determine the architecture of the system (e.g., a simple or a redundant architecture). The Development Assurance Level (DAL) of the function is also assigned based on the severity, and the possible system architecture mitigations. The more severe the Failure Condition, the greater the DAL (E to A).

It is now widely accepted that that existing standards may not be fully applicable to systems embedding ML components, and even if applicable, they may not comply with the actual intent of certification. Therefore, an evolution of the underlying development process is proposed in §4.1, in order to introduce possible adaptations of these standards to allow for ML-based products certification.

### 2.3.3.2 Towards new practices

*Overarching properties*

As part of its effort towards "Streamlining Assurance Processes", the Federal Aviation Administration (FAA) launched in 2016 an initiative called "Overarching Properties". The objective of this initiative is to develop a minimum set of properties such that if a product is shown to possess all these properties, then it can be certified. As of 2019, the three overarching properties retained are:

1. **Intent**. The defined intended functions are correct and complete with respect to the desired system behaviour.
2. **Correctness.** The implementation is correct with respect to its defined intended functions, under foreseeable operating conditions.
3. **Innocuity.** Any part of the implementation that is not required by the defined intended behaviour has no unacceptable safety impact.

These properties are, by construction, too abstract to constitute an actionable and complete means of compliance for certification. In practice, they shall be refined to be applicable, leaving an opportunity to establish a specific set of properties for the certification of ML systems.

*Abstraction layer*

On the European side, and in order to evaluate more flexible and efficient ways for software and hardware qualification, the EASA (European Union Aviation Safety Agency) proposed the concept of "Abstraction layer".

The objective is to propose an abstraction layer "above the existing standards" (DO-178C/ED-12C and DO-254/ED-80), that captures the properties required for certification, independently from the technology and process used. Unlike the FAA with the "Overarching properties", the EASA opted for a bottom-up approach in order to leverage the experience gained on the existing standards.

The Abstraction layer will probably not constitute an actionable and complete means of compliance for ML certification. If used for ML certification, the Abstraction Layer evaluation framework may have to be supplemented to assess any new methodology specific to the ML development.

The current and future structure of the certification framework, together with the Overarching Properties and Abstraction Layer initiatives, are depicted in Figure 6.

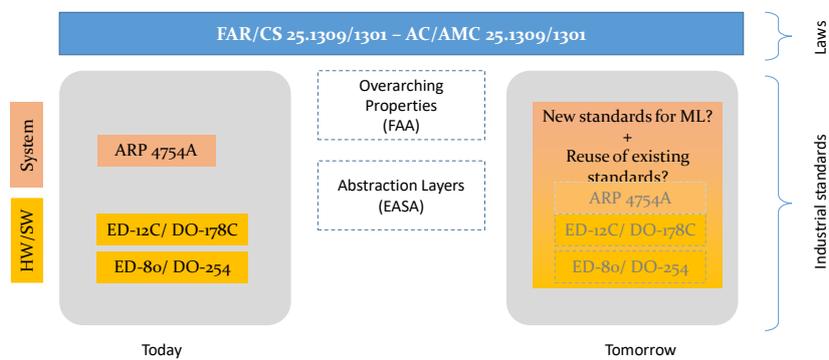

*Figure 6. FAA's Overarching properties and EASA's abstraction layer*

# 3 Analysis

The aim of this White Paper is to point out some of the most important challenges for the certification of systems embedding ML. In this section, we briefly describe the approach used by the ML Certification Workgroup to identify these challenges. The challenges themselves are presented in Section 4.

## 3.1 CERTIFICATION OBJECTIVES, CHALLENGES, AND HIGH-LEVEL PROPERTIES

A *certification objective* is a property which satisfaction contributes to show that a system meets the applicable certification specifications (e.g., "the algorithms are accurate", "the architecture is consistent", etc.). A *challenge* is a particular difficulty to demonstrate that a system meets those specifications.

So, why is Machine Learning raising specific challenges for certification?

Machine Learning is data-driven, i.e. the behaviour of a ML-based system is learnt from data. With respect to classical systems, ML requires new development activities such as data collection and preparation (cleaning, labelling, normalization…) and model training (model selection, hyper-parameter tuning…). Existing development assurance standards used for certification (e.g. DO-178C/ED-12C, ISO26262, EN50128) are not data-driven and do not consider these ML-specific development activities. Therefore, existing certification objectives are not sufficient to address the ML-based systems.

Facing this lack of applicable certification objectives, the workgroup started its work from a simple question:

> *What properties should a ML-based system possess to be certified?*

To answer this question, the working group decided to establish a list of properties which, if possessed by a ML technique, was considered to have a positive impact on the capability to certify a ML-based system using this technique. Those properties, called High-Level Properties (HLPs), carry the ML certification objectives.

## 3.2 HIGH-LEVEL PROPERTIES

High-Level Properties have been considered, collectively, to be necessary (but not sufficient) for the development of ML in the industrial domains represented in the workgroup.

The list of HLPs considered in this White Paper is given hereafter. The definitions proposed have been discussed among members, considering the literature and the existing standards.

- **Auditability:** *The extent to which an independent examination of the development and verification process of the system can be performed.*
- **Data Quality:** *The extent to which data are free of defects and possess desired features.*
- **Explainability:**[19] *The extent to which the behaviour of a Machine Learning model can be made understandable to humans.*

---

[19] Many terms such as Interpretability and Transparency can be found in the literature. §4.6 will give the definitions used by the Working Group.

- **Maintainability:** *Ability of extending/improving a given system while maintaining its compliance with the unchanged requirements.*
- **Resilience:** *Ability for a system to continue to operate while an error or a fault has occurred.*
- **Robustness:** (Global) *Ability of the system to perform the intended function in the presence of abnormal or unknown inputs* / (Local) *The extent to which the system provides equivalent responses for similar inputs.*
- **Specifiability:** *The extent to which the system can be correctly and completely described through a list of requirements.*[20]
- **Verifiability:** *Ability to evaluate an implementation of requirements to determine that they have been met (adapted from ARP4754A).*

## 3.3 HIGH-LEVEL PROPERTIES AND CERTIFICATION

In this document, HLPs are essentially introduced to identify the objectives and challenges raised by the certification of ML-based systems. Later, those HLPs might be integrated, explicitly or implicitly, in a ML certification framework. This framework will follow and/or combine the following approaches for certification:

- **Process-based certification:** the completion of a predefined development assurance process during the development of a product is the assurance that this product complies with the requirements laid down in the certification basis.
  Most of the standards currently used for certification are "process-based". In this approach, HLPs are demonstrated as part of the full process. For example, "Specifiability" may be demonstrated at the specification activity; "Verifiability" and "Robustness" may be demonstrated as part of the verification activity, "Resilience" may be demonstrated as part of the safety assessment activity, etc.
- **Property-based certification:** the demonstration by the applicant that a predefined set of properties is met by a product is the assurance that this product complies with the requirements laid down in the certification basis. In this approach, the HLPs may be some (or all) of the properties to be demonstrated.
  Even though there is a growing interest for this approach, one of the difficulties is to prove that the selected set of properties completely covers the desired certification objectives. A clear consensus on an acceptable set of properties has not been reached yet, but "property-based" certification is in line with the Overarching Properties initiative, and it remains an option considered for the future.

Those two approaches are not exclusive and can be combined. Whatever the certification framework chosen, we believe that the ability to demonstrate the HLPs will be a key enabler for ML-based system certification. Depending on the system and the certification strategy, the HLPs to be demonstrated, as well as the depth of demonstration may vary, but in any case, this list of HLPs summarizes the main challenges of ML certification.

## 3.4 HIGH-LEVEL PROPERTIES ANALYSIS APPROACHES

Once the High-Level Properties selected, the next step is to identify the challenges in meeting these HLPs. Towards that goal, we have considered three complementary approaches:

---

[20] Requirement: an identifiable element of a function specification that can be validated and against which an implementation can be verified.

- An approach based on the analysis of a typical ML-system development process.
- An approach based on the identification of non-ML techniques, already implemented in certified systems, raising similar challenges to those encountered on ML techniques (called *similarity analysis).*
- An approach based on the selection of ML-techniques that do not raise some or all of the previous challenges (called *backward analysis*).

Those approaches are briefly introduced hereafter.

### 3.4.1 ML component development process analysis

This approach aims to analyse a typical ML development process and to determine the effect of the actions and decisions taken by a ML designer on the HLPs. It is aimed at pointing out the steps in the process where faults[21] having an impact on the HLPs are most likely to be introduced.

Note that, by construction, the results produced by a ML component will generally be "erroneous" due to statistical nature of the learning process. Here, a *fault* in the ML component design process is a human-made action (including choices) that could have an impact on the overall safety of the system.

As shown partially[22] on Figure 7, this analysis shall cover all engineering activities. For an implementation based on neural networks, for instance, it shall include the specification of the function, the acquisition of the datasets, the choice of the initial values of the weights, the choice of the activation function or optimization method, the choice of the implementation framework, etc.

---

[21] Fault, error and failure correspond to Laprie's definition (see §1.3 and §4.2.2.1)
[22] For sake of conciseness, implementation steps are hidden in this figure.

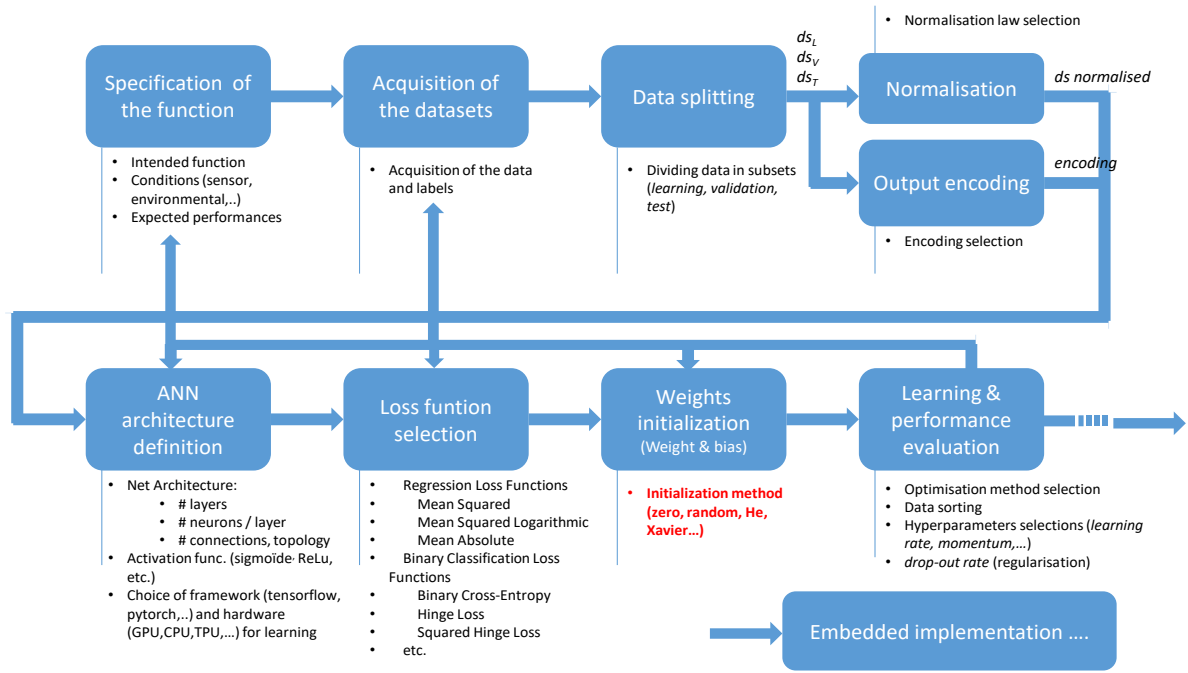

*Figure 7. Simplified supervised Machine Learning Workflow*

The diagram shows the main phases along with some of the actions taken by the designer. For instance, during the "weight initialization phase", the designer has to select one particular method among those available. In this phase, a fault may be introduced by choosing an inappropriate method (for instance, using an "all-to-zero" initialization, see Figure 8).

This example is obviously trivial, but it illustrates the principle of the analysis: identify where faults can be introduced and their consequences on the system's behaviour, and define appropriate prevention, detection and elimination, and/or mitigation means. These means involve development process activities (i.e., provide guidance to prevent the fault to be introduced, provide appropriate verification means to activate and detect the fault/error, etc.) and system design activities (to mitigate the effect of the fault/error).

Unfortunately, finding the consequence of a fault on the system safety was intractable in practice, due to the opacity and complexity of ML training phase. So, we moved from the final safety properties to the intermediate HLPs.

Figure 8 illustrates how the engineering choices could impact a given HLP (robustness). This systematic analysis is aimed at identifying the effects of ML design choices on confidence they can give on ML component through the HLPs.

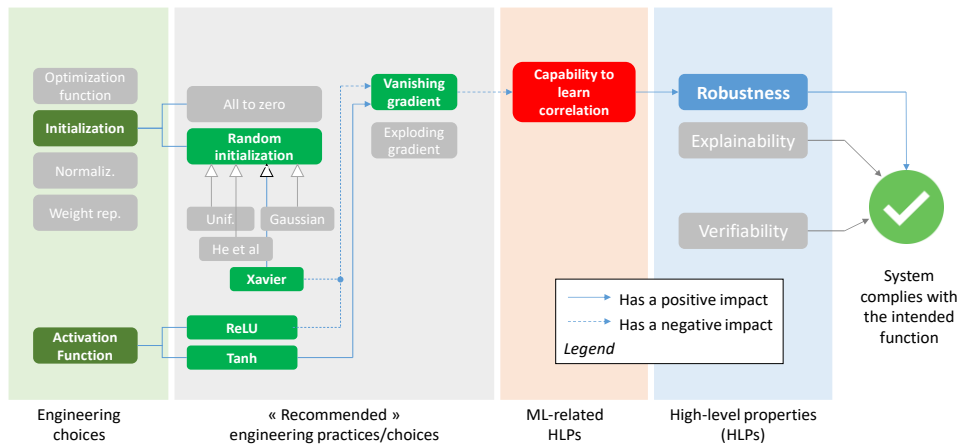

*Figure 8. From High-Level Properties to Engineering choices*

### 3.4.2 Similarity analysis

Another approach is to search for situations where non-ML systems showing characteristics "similar"[23] to those encountered on ML systems have already been successfully certified.

Hereafter, we consider three specific cases:

- Kalman filters
- Computation of worst-case execution times
- Complex COTS processors.

#### 3.4.2.1 Kalman filters

Kalman Filtering (KF) is a predictive filtering technique used, for instance, to compute the position and altitude of robots or aircrafts. KF algorithms are "similar" to ML techniques in the sense that (i) they implement a statistical estimation process, (ii) their behaviour depend strongly on empirical data (the covariance matrix) and on hypotheses on the inputs (the distribution of the noise), (iii) they produce outputs that are associated with an estimation of the result quality (an confidence ellipsoid). Note that KF algorithms also show strong differences with ML: their input space is usually small (e.g., the position and velocity of a mobile); they embed a model of the physical system (to perform the prediction) ; errors used by KF algorithms are estimated on the basis of physical models representing upper bounds of actual errors. Anyway, as for ML, their dependency on data (point (ii) above) makes verifying the correctness of the algorithm *implementation* not sufficient. A combination of mathematical demonstrations and tests is actually required to assess the intended function (including precision, stability, etc.).

In practice, the use of KF in aircraft positioning systems is covered by the DO-229 standard [3] which gives very detailed recommendations about validation of this algorithm (type and number of tests), on the basis of a precise (yet statistical) knowledge of the system's environment (satellite positioning errors, gravity model, ionospheric error models, etc.). Hence, confidence is obtained by applying a function-specific standard. For ML-based systems, this approach would mean establishing a specific standard for each function.

---

[23] "Similar" is taken in an abstract sense.

Work is still needed to determine to what extent such statistical approaches could be applied to ML-based systems. At least, the computing power available thanks to supercomputers and clouds makes safety demonstration based on "massive testing" a solution to be explored for ML certification. However, computing power is not sufficient, and a convincing testing approach also requires robust data collection and test processes to assure effective statistical representativeness.

### 3.4.2.2 Worst-Case Execution Times

Statistical analysis is also a method for the estimation of Worst-Case Execution Times (WCET) of software components deployed on the very complex processors and System-on-Chips used today in embedded systems [4].

Can these statistical techniques be transposed to ML based systems?

Here again, practices cannot be easily transposed to the ML domain: those hardware components are not strictly considered as "black boxes" and statistical analysis is strongly supported by analyses of the hardware and software (see e.g., CAST-32A [5] and soon to be published AMC 20-193).

### 3.4.2.3 Hardware COTS

Explainability is a challenge for ML system, but it is already a challenge for the complex software or hardware systems deployed today [6]. So, could current practices be applied to ML systems?

Let us consider again the case of the complex COTS processors used on embedded systems. As already noticed, complexity of those components is usually so high that they can partly be seen as black-boxes. So, how do current certification practices address this issue?

In aeronautics, the development of complex hardware is covered by the DO-254/ED-80 with some additional guidance given in AMC 20-152A [7]. Section 6 of [7] addresses the specific case of Off-the-Shelf devices (COTS), including complex processors. The selection, qualification and configuration management of those components is addressed in the "Electronic Component Management Process" (ECMP) which shall consider, in particular, the *maturity of the component* and the *errata of the device*. As stated in Appendix B of [7], maturity can be assessed by the "time of service of the item", the "widespread use in service", "product service experience", and the "decreasing rate of errata". This approach is perfectly adapted to general purpose components used at a large scale in critical *and* non-critical devices (mobile devices, home appliances, etc.), and for a long time, as it is the case for processors. This will certainly be very different for a ML component dedicated to a specific task.

The "input domain" of a processor is also well defined: a processor implements an Instruction Set Architecture (ISA) that specifies very strictly how the software will interact with the processor. The existence of a well-defined interface, shared by all applications, makes it "easier" to leverage on field experience to get confidence on the component. Again, the variability of the operational domain may be significantly higher for ML components, making confidence based on field experience harder to justify.

Finally, software verification activities will also contribute to gaining confidence on the component. As stated in [8, Sec. 9.2], "The development assurance of microprocessors and of the core processing part of the microcontrollers and of highly complex COTS microcontrollers (Core Processing Unit) will be based on the application of ED-12B/DO-178B to the software they host, including testing of the software on the target microprocessor/microcontroller/ highly complex COTS microcontroller". Stated differently, the hardware is considered to be (partially) tested along with the software it executes.

So, even though complexity and a certain lack of transparency seems to be common issues of both COTS processors and ML components, solutions and recommendations applicable to the former seem to be hardly applicable to the latter.

### 3.4.3 Backward analysis

The "backward analysis" takes the problem the other way round by considering *first* the ML solutions possessing some "good" properties for certification, and *then* determining to which problems they can be applied, with what guarantees, and under what limits. We illustrate this approach with decision trees and NNs.

#### *3.4.3.1 Decision trees and explainability*

The property of explainability has been introduced in the previous section and is developed in §4.6. Explanations may be needed by different actors and for different reasons. For instance, a pilot may need explanations to understand the decision taken by its ML-based co-pilot in order to be aware of the current situation and take control of the system if needed. An engineer may need explanations to investigate the origin of a problem observed in operation or after an accident.[24] Explanations may also be required to understand and then to act or react, but also to gain confidence on the system that takes decisions.

All ML techniques are not equivalent with respect to explainability. Let us consider for instance the case of decision trees.

Decision trees predict the label associated with an input taking a series of decisions based on distinctive features [2]. They can be used for different types of applications such as classification or regression problems [9]–[11].

For decision trees, providing an explanation simply consists to expose the decisions taken, from the root to the leave of the decision tree. This property enables to provide a full description of the decision process after learning, and a comprehensive set of Low-Level Requirements for software implementation. Clearly, demonstrating the HLP *explainability* is easier for decision trees than for, for instance, NNs.

#### *3.4.3.2 Reluplex and provability*

Formal verification methods (FM) such as deductive methods, abstract interpretation, or model checking, aims at verifying that a model of a system satisfies some property on a mathematical basis. Grounded on mathematics, these methods bring two fundamental properties, namely soundness and completeness, which can hardly be achieved by other means such as testing or reviews. FM are already considered as viable verification methods at item level by authorities (see DO-333/ED-216 supplement to the DO-178C/ED-12C).

In the current context, the objective is to demonstrate the compliance of an ML implementation to its specification in all possible situations, without explicitly testing the behaviour of the system in each of them. Completeness is very difficult to achieve by testing for problems with a very large dimensionality (as it is often the case with problems solved by ML). Whatever the testing effort, it will only cover an infinitesimal part of the actual input space. Confidence in the test could be improved drastically if *equivalence classes*[25] can be defined but, unfortunately, it is not yet very clear how those classes can be defined in the case of, for instance, NNs. Being able to apply a formal verification technique on a ML design would represent a significant improvement over testing.

---

[24] E.g., to ensure the "continued airworthiness" in aeronautics [11, Sec. AMC 25.19].

So, are there ML techniques amenable to formal verification?

The answer is yes: formal methods have already been considered in the domain of Machine Learning [12]. For instance, the Reluplex method [13] is used to verify properties on NN with ReLU activation functions. It is based on the classical Simplex algorithm, extended to address ReLU constraints. As for the Simplex itself, the technique gives satisfying results in practice despite the problem being NP-complex. It has been applied successfully on NN of moderate size (8 layers of 300 nodes each, fully connected).

Reluplex has been used on an NN-based implementation of an unmanned collision avoidance system (ACAS-Xu), a system generating horizontal manoeuvre advisories in order to prevent collisions. The ML implementation of the ACAS Xu is a small memory-footprint (3Mb) alternative to the existing implementation based on lookup tables (2Gb).[26]

This technique provides formal guarantees on properties that can be expressed using linear combinations of variables, which means that only a subset of the possible functional properties can be verified that way. For instance, on the ACAS-Xu, safety properties such as "no unnecessary turning advisories", "strong alert does not occur when intruder vertically distant", etc. have been verified. Reluplex has also been applied to demonstrate the $\delta$-local and global robustness.[27] Such formal proof, providing demonstration for HLPs *robustness* and *verifiability*, may be alleviating some of the requirements put on the learning phase.

## 3.5   FROM HIGH-LEVEL PROPERTIES TO CHALLENGES

Based on the identification of the HLPs and the three types of analyses performed (ML development process analysis, similarity analysis and backward analysis), the certification workgroup has identified seven major challenges for ML certification:

1. Probabilistic assessment
2. Resilience
3. Specifiability
4. Data Quality and Representativeness
5. Explainability
6. Robustness
7. Verifiability

The close relationship between HLPs and challenges appears clearly in this list, where we find six HLPs (2 to 7) among the seven main challenges identified. The other HLPs

---

[25] An equivalence class is defined with respect to some fault model so that "any test vectors in a class will discover the same fault".
[26] The ACAS Xu was too complex for manual implementation. The strategies considered to reduce the size of the data required to implement the functions are described in [14].
[27] A NN is $\delta$-locally robust if for two inputs $x$ and $x'$ such that $\|x - x'\| \leq \delta$ the network assigns the same label to $x$ and $x'$.

PAGE 29

can often be linked to one of these seven main challenges. These main challenges are presented in details in the next Section.

# 4 Challenges of Machine Learning in Certified Systems

Challenges have been organized in seven main categories: Probabilistic assessment (§4.2), Resilience (§4.3), Specifiability (§4.4), Data Quality and Representativeness (§4.5), Explainability (§4.6), Robustness (§4.7), and Verifiability (§4.8).

Before detailing them individually, we propose to show where those challenges mainly arise in the general Machine Learning development process depicted on Figure 9.

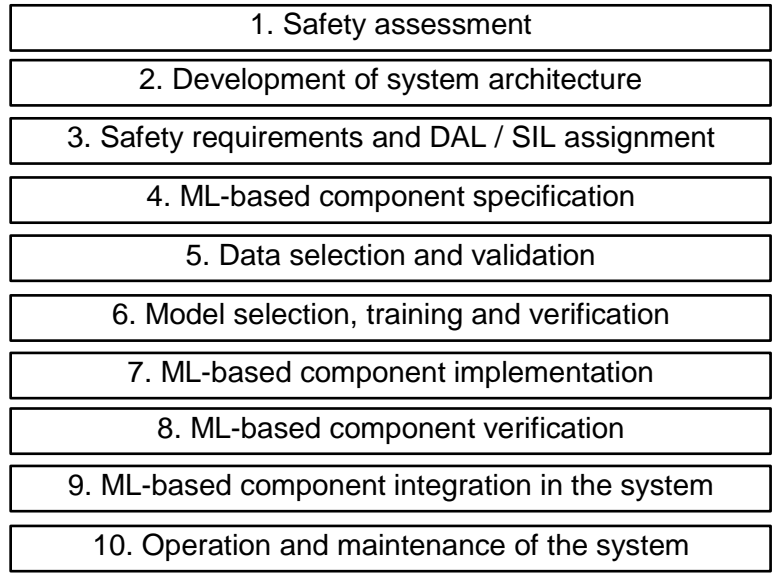

1. Safety assessment
2. Development of system architecture
3. Safety requirements and DAL / SIL assignment
4. ML-based component specification
5. Data selection and validation
6. Model selection, training and verification
7. ML-based component implementation
8. ML-based component verification
9. ML-based component integration in the system
10. Operation and maintenance of the system

Figure 9. General (simplified) development process

## 4.1 CHALLENGES IN A TYPICAL ML DEVELOPMENT PROCESS

In accordance with the scope of the document presented in §1.3, the ML development process described here will be limited to *off-line supervised learning techniques.* Taking into account online learning or unsupervised/reinforcement learning, would require various modifications of this process that are not discussed thereafter.

This process is extremely simplified. In particular, it leaves no room for loops, reiterations, etc., In addition, for sake of simplicity, each challenge has been allocated to a unique phase of the process although some of them, such as *explainability* and *verifiability,* could actually concern multiple phases.

### 4.1.1 Context

One can anticipate that ML algorithms use will vary from the design of a single part of a system function to the design of the whole system itself. The ML model development, as part of the full system, will require the system expertise and should therefore be addressed at system level. In addition, without being mandatory, it is likely that resilience of ML-based systems will require mitigations at system level. For these reasons, the process presented here starts at system level, and is not limited to component or item level.

In the following, we consider that part of the intended function of the system is realized by a ML component (Figure 10).

Before going into more details, it is reminded that the goal of certification process is to give confidence, based on a set of structured evidences, on the capability of the system to correctly and safely perform its intended function.

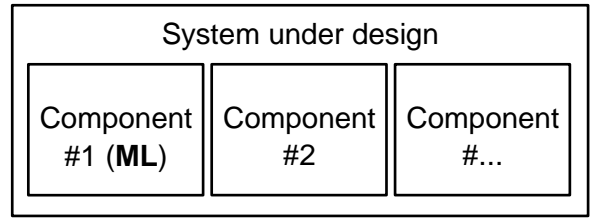

*Figure 10. ML Components within a system under design*

### 4.1.2 Safety assessment

The safety assessment aims at identifying the failure[28] conditions of the system, associated hazards, their effects, and their classification (e.g. no safety effect, minor, major, hazardous, catastrophic). It enables identifying top-level safety requirements for the whole system. The safety assessment is then refined and updated throughout the system development process together with the system architecture. The probabilistic nature of ML (see §2.1) raises specific issues on the way system safety requirements are verified.

| **MAIN CHALLENGE #1: PROBABILISTIC ASSESSMENT** |
|---|
| *For safety critical systems, quantitative safety analysis is used to assess properties such as "the probability of a catastrophic event of an aircraft shall be lower than $10^{-9}$ per flight hour". Similarly, Machine Learning techniques rely on mathematic practices that include statistics and probabilities. Nevertheless, despite their similarities, the two domains often employ different definitions and interpretations of key concepts. This makes difficult the endeavor of establishing a safety assessment methodology for ML-based systems.* |

### 4.1.3 Development of system architecture

Several system architectures may be used to implement the same intended function. The choice of a specific architecture is made considering the top-level safety requirements (see previous section), as well as other technical, economic, and programmatic constraints. For each architecture, the safety requirements allocated to the ML-based component and its interfaces with other components must be considered with specific care. Due to the probabilistic nature of Machine Learning, resilience, defined as the ability for a system to continue to operate while a fault/error has occurred, is a primary concern when designing the system architecture.

| **MAIN CHALLENGE #2: RESILIENCE** |
|---|
| *Resilience is crucial to ensure safe operation of the system. Resilience typically raises challenges regarding the definition of an "abnormal behaviour", the monitoring of the system at runtime, and the identification of mitigation strategies. With Machine Learning, these challenges are made more complex because of the usually wider range of possible inputs (e.g. images), the difficulties to adopt classical strategies (e.g., redundancy with dissimilarity), and the ML-specific vulnerabilities.* |

---

[28] See definition in §4.2.2.1.

### 4.1.4 Safety requirements and DAL / SIL assignment

Once the system architecture is defined, the safety requirements and Design Assurance Level / Safety Integrity Level (DAL / SIL) can be assigned to the ML-based component. Design assurance aims at limiting the likelihood of development faults, in order to fulfill the requirements.

The safety requirements are derived from the safety allocation and the system architecture. Safety requirements may be qualitative (e.g. no single failure) or quantitative (probability of erroneous output, probability of loss, etc.). The DAL / SIL corresponds to the level of rigor of development assurance tasks. It is chosen in relation to the severity of the effects of the failures of the component on the system. For example, if a development fault could result in a catastrophic failure condition, then the greatest DAL / SIL is selected for this component.

### 4.1.5 ML-based component specification

At this stage, requirements have been allocated to the ML-based component. This activity concerns safety requirements, functional requirements, customer requirements, operational requirements, performance requirements, physical and installation requirements, maintainability requirements, interface requirements, and any additional requirements deemed necessary for the development of the ML-based component.

As shown on Figure 11, those requirements have to be further refined and completed up to the point where they allow the development of the ML-based component. Some top-level requirements specify general expectations, whereas other level requirements specify implementation details. In particular, in the case of supervised learning, data could be considered as detailed requirements of the intended behaviour of the ML-based component. Similarly, the structure of the ML model, its parameters and hyper-parameters could also be considered as detailed requirements for the ML-based component.

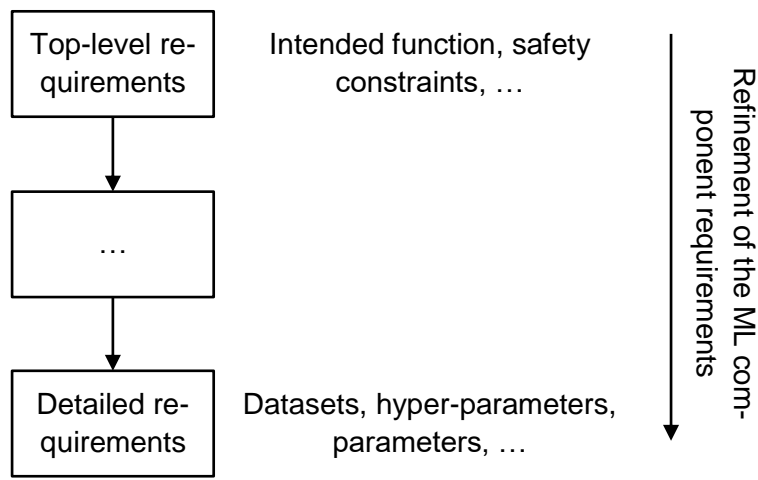

Figure 11. Refinement of the specification

| **MAIN CHALLENGE #3: SPECIFIABILITY** |
|---|
| *Specifiability is the ability to describe the intended function of a system in terms of functional, safety, operational, and environmental aspects. This practice allows engineers to designate the target of the development process and to demonstrate that this target has been hit. Nowadays, this is a pillar to build acceptably safe systems. Because ML techniques are often used to address problems that are by nature hard to specify, they raise specific challenges to include them in safe systems, the question of trust being one of these challenges.* |

### 4.1.6 Data selection and validation

The final behaviour of the ML-component is largely determined by the selected data. The goal of this activity is to select data with the expected quality attributes (representativeness, lack of undesired bias, timeliness, partitioning, accuracy, integrity, consistency, completeness…). Once data are selected, data validation consists in verifying that the desired quality attributes are indeed present. Validation can be performed by systematic check of certain attributes, sampling, cross-check, etc.

| **MAIN CHALLENGE #4: DATA QUALITY AND REPRESENTATIVENESS** |
| --- |
| *Machine Learning-based systems rely on the exploitation of information contained in datasets. Therefore, the quality of these data, and in particular their representativeness, determines the confidence on the outputs of the ML-based components.*<br>*The qualification of a dataset with respect to properties related to quality can be particularly complex, and depends strongly on the use case.* |

### 4.1.7 Model selection, training and verification

Together with data selection, model selection and training are crucial to the correct design of the ML-based component. There are many models available, and many ways to train a model. Certification standards should impose neither a specific model nor a specific training technique. The focus should rather be on the properties, such as explainability and robustness that the model must possess after training. Other properties such as maintainability, auditability, etc. could also be checked at this stage.

The depth of demonstration of these properties can vary depending on the requirements. If these properties are required for the overall safety demonstration, then in-depth demonstration is necessary. However, if no requirement stems from the safety assessment and component specification, then the ML model could remain a "black box", without explainability and/or robustness demonstration.

Some verification activities can be performed directly on the model, before implementation. If it is the case, it should be demonstrated that the results of these verification activities are preserved after implementation (see §4.1.9 for more details on the verification activities).

| **MAIN CHALLENGE #5: EXPLAINABILITY** |
| --- |
| *The opacity of ML models is seen as a major limitation for their development and deployment, especially for systems delivering high stake decisions. Quite recently, this concern has caught the attention of the research community through the XAI (eXplainable Artificial Intelligence) initiative which aims to make these models explainable. The ongoing investigations highlight many challenges which are not only technical but also conceptual. The problem is not only to open the black box but to establish also the purpose of explanations, the properties they must fulfill and their inherent limits, in particular in the scope of certification.* |

| **MAIN CHALLENGE #6: ROBUSTNESS** |
| --- |
| *Robustness is defined as the ability of the system to perform the intended function in the presence of abnormal or unknown inputs, and to provide equivalent responses within the neighbourhood of an input (local robustness). This property, which is one of the major stakes of certification, is also a very active research domain in Machine Learning. Robustness raises many challenges, from the definition of metrics for assessing robustness or similarity, to out-of-domain detection, and obviously adversarial attacks and defence.* |

### 4.1.8 ML-based component implementation

At this step of the product lifecycle, the behaviour of the ML-based component is fully defined and can be implemented.

The three main implementation phases, i.e., hardware production, software coding, and hardware / software integration (see Figure 12) being similar to those performed on traditional systems, existing development standards can be applied. Therefore, the implementation of ML-based components is not addressed in this White Paper. Nevertheless, the workgroup is aware that a particular attention shall be taken on the implementation errors that may be introduced during this phase: incorrect representation of the computed weights on the target, insufficient memory or processing resources, accuracy and performance degradation with respect to the model used during learning, reduction of the arithmetic precision, etc.

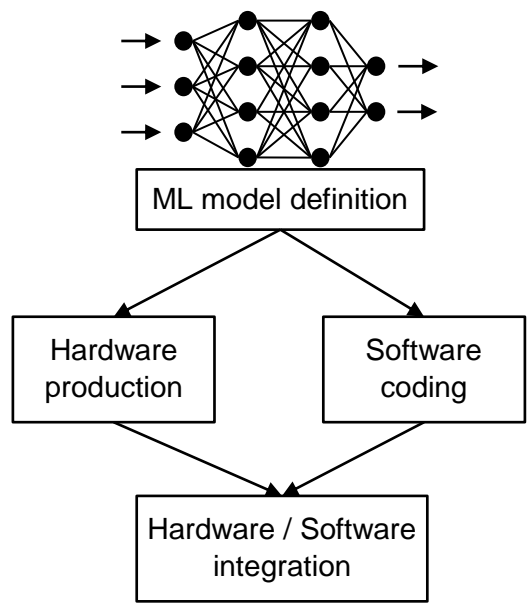

Figure 12. ML-based component implementation

### 4.1.9 ML-based component verification

The purpose of this step is to check that the requirements are indeed fulfilled by the ML-based component. Various verification strategies can be adopted, depending on the verification means available:

- Testing (including massive testing and adversarial testing)
- Formal methods
- Service experience
- Etc.

Verification shall be performed on the integrated ML-based component, but "verification credits" can also be taken from verification activities performed on the model before imple-mentation or performed at item level (software / hardware). Whatever the strategy, verification is done against the requirements captured during the ML-based component specification (including innocuity[29]), so verification and specification challenges are tightly linked.

> **MAIN CHALLENGE #7: VERIFIABILITY**
>
> *A large and ever-growing set of methods and tools, ranging from formal mathematical methods to massive testing, is available to verify ML-based systems. However, understanding the applicability and limits of each of them, and combining them to reach the verification objectives raises many challenges...*

### 4.1.10 ML-based component integration

Once the ML-based component has been designed, implemented, and verified, it can be integrated in the system under design, and system-level verification activities can be performed. Note that those activities can also contribute to the verification of the ML-based component (cf. §4.1.9), and reciprocally.

### 4.1.11 Operation and maintenance

After entry into service, the system is operated and maintained. Specific challenges (in addition to the previous ones) may be raised by these phases of the life-cycle, including the ability to maintain, repair, and update the ML-based component. These aspects shall be taken into account as part of the development of the system, but they are not addressed in this document.

### 4.1.12 Summary and possible variations in this process

As stated in the beginning of this section, our development process is highly simplified. In practice, the steps are not strictly executed from top to bottom. In particular, the safety assessment is not performed once for all at the beginning, but it is updated all along the lifecycle and finalized only once the system is finished. In addition, we do not show transverse activities that play a significant role in the development process such as requirement validation, configuration management, certification liaison, etc. Finally, all the steps in this process are not mandatory.

Depending on the certification strategy, the effort of demonstration spent at each step of the process may vary, just like the "potentiometers" introduced in EASA trustworthy building blocks [15]. For example, if the top-level requirements stemming from step 1, step 2 and step 3 can be completely demonstrated through formal proof in step 8, it may be acceptable to alleviate all or part of the intermediate steps. On the contrary, if verification methods in step 8 are weak, then additional effort should be made for those intermediate steps in order to provide confidence in the development. More generally, the aim of this process is to ensure that the ML-based component is correctly and safely performing its intended function: if this can be achieved without performing all the steps described above, some of them could become optional.

The next subsections further detail each of the seven main challenges identified above.

---

[29] "Any part of the implementation that is not required by the defined intended behaviour has no unacceptable safety impact.", see §2.3.3.2.

## 4.2 MAIN CHALLENGE #1: PROBABILISTIC ASSESSMENT

> DISCLAIMER: this section presents the first conclusion of a work in progress. It may contain several inaccuracies, inconsistencies in vocabulary and in concepts, and surely lacks references.

### 4.2.1 Introduction

Safety critical systems must comply with dependability requirements. Such compliance is typically achieved by validation and verification activities but also by a control of the development process. A highly critical function such as the flight control system of an aircraft is considered to have catastrophic consequences in case of malfunction. Thus, regulation impose for a large transport aircraft that (i) no single failure shall lead to the loss of the function and (ii) the function shall not be lost with a rate greater than $10^{-9}$/h. These dependability requirements are propagated down to software components. This is done on the basis of current standards (e.g. DO178C) for "classical" software.

We argue that ML-based software should not be treated as "classical" software. New safety analysis methodologies must be developed to assess the capability of such systems to comply with safety requirements. Thus, the fundamental question raised by this statement is: **(Q1) How to develop safety approaches and analyses on systems including ML-based applications?**

Both fields, i.e. Machine Learning and safety analysis, rely on statistical tools and probabilistic reasoning to assess the behaviour of a system. Nevertheless, despite their similarities, as we will present in the following paragraphs, the two domains often employ different definitions and interpretations of key concepts. So, **(Q2) Can we adapt probabilistic safety assessment approaches to assess the dependability of ML-based systems?**

Besides, one of the motivations for this subject is that it has a direct impact on the engineering. Indeed, from the level of risks of the systems to develop (see definitions below), the current norms and standards define 1) an applicable design level (DAL,[30] ASIL,[31] Software Level,[32] etc.) and 2) a set of rules to design and develop the system, the rigor of which depends on this level. The underlying principle is that the stricter the rules are, the smaller the likelihood to introduce faults during the engineering should be. Rules are introduced so that the likelihood of the remaining faults introduced during the system design can be considered as an acceptable risk.

Considering that the uncertainties of a ML model is contributing to the level of risk of the system to which it belongs, it is important to identify at which level this contribution is done, and to consider conjointly the ML model faults/errors, the engineering faults, and the global system risk. As shown on Table 4, three cases can be distinguished, each one determining a specific engineering effort. This strong dependency between the ML uncertainties and the risk level motivates a formal definition of each case; this is addressed in the next section.

---

[30] Development Assurance Level, defined in ARP 4754A
[31] Automotive Safety Integrity Level, defined in ISO-26262
[32] defined in DO-178/ED-12

Table 4. Contribution of the ML to the system risk level

| Case | Situation for engineering the model | System level consideration |
|---|---|---|
| ML uncertainties level is **quite higher** than the acceptable risk for the system | In this case, the model uncertainties level itself cannot ensure the acceptable level of risk for the system. The reduction of engineering faults by the application of the standards is not relevant. | A mitigation mean, independent of the MLM shall be introduced at system level, ensuring that the uncertainties of the model are correctly "filtered" |
| ML uncertainties level is **of the same order of magnitude** than the acceptable risk for the system | In this case, all the sources of faults have to be considered. The engineering must conform to the standards to ensure that the engineering faults are under a good level of control. The mix of the faults/errors due to the ML uncertainties level with the engineering faults must be at the level of the accepted risk of the system. | The MLM can be in direct line with the FE[33] (taking into account additionally that the double or triple faults protections demanded by the standards) |
| ML uncertainties level is **quite lower** than the acceptable risk for the system | As the engineering faults have a higher contribution, these only must be considered, applying the existing standards. | |
| *MLM : Machine Learning Model | | |

### 4.2.2 Reminder on safety

#### 4.2.2.1 Some vocabulary

As a reminder, see §1.3, we use Fault/Error/ Failure definitions from Laprie [29], which might differ from specific domain standard definitions (for instance ED-135 / ARP4761):

Failure: An event that occurs when the delivered service deviates from correct service.

Error: The part of the system state that may cause a subsequent failure: a failure occurs when an error reaches the service interface and alters the service.

Fault: The adjudged or hypothesized cause of an error (for instance an incorrect action or decision, a mistake in specification, design, or implementation).

#### 4.2.2.2 Uncertainties and risk

To develop a safe system, the main objective is to ensure that there will be no failure – i.e., the effect of some error – with an unacceptable safety impact under foreseeable conditions.

All experimental measurements can be affected by two fundamentally different types of observational errors: random errors and systematic errors.

Random errors refer to distinct values obtained when repeated measurements of the same attribute or quantity are taken. Random errors include for example the effect of Single Event Upset (SEU) on electronic components. They are stochastic and unpredictable by nature (i.e. uncertainty about the measurement cannot be further reduced). In this context, uncertainty is seen as an inherent part of the measuring process. In some cases, statistical methods may be used to analyse the data and determine the distribution of observations and its attributes (mean, variance, mode...).

---
[33] Feared Event

Systematic errors are not determined by chance but are introduced by a limit in the measuring process. They are predictable (although not always easily detectable), in the sense that the measurement always produces the same value whenever the experiment is performed in the same way. For these reasons, systematic errors are sometimes called statistical biases.

In practice, most measurements are affected by both types of errors and separating or analyzing them may prove difficult or costly.

System design standards address both types of errors, but they typically make a clear distinction between how they affect hardware and software components. For example, ISO 26262 defines random hardware errors[34] as errors "that can occur unpredictably during the lifetime of a hardware element, and that follow a probability distribution." The same standard defines systematic errors as error "related in a deterministic way to a certain cause that can only be eliminated by a change of the design or of the manufacturing process, operational procedures, documentation or other relevant factors". While this standard addresses both types of errors in the case of hardware components design, it focuses exclusively on systematic errors in the case of software design.

However, all ML-based modelling are stochastic by nature and therefore affected by random errors. Moreover, as this modelling is usually data-driven, it is implicitly affected by data measurement errors, therefore by both types of errors. ML acknowledges both types of errors, which usually translates to the bias-variance problem. The bias error is accounted by erroneous assumptions in the learning algorithm. The variance is the error incurred from sensitivity to small fluctuations in the training dataset. In order to minimize the expected generalization error, all ML modelling must solve a bias-variance trade-off.

According to these considerations, we argue that in their current form, software design standards do not apply to ML-based software. Consequently, we argue that new norms must be formulated in order to address in a rigorous fashion both types of errors in a ML-based system design.

<u>Definition (ARP4761 Event)</u>: An occurrence which has its origin distinct from the aircraft, such as atmospheric conditions (e.g., wind gusts, temperature variations, icing, lighting strikes), runaway conditions, cabin and baggage fires, etc.

<u>Definition (ARP4761 reliability)</u>: The probability that an item will perform a required function under specified conditions, without failure, for a specified period.

The system analysis studies the potential safety impacts of system failures, giving them a level of severity/criticality. This level of severity is assigned to the failure that provokes the consequence, for instance with an FMEA (failure mode and effects analysis).

<u>Definition (ARP4754A / ARP4761 Risk)</u>: The combination of the frequency (probability) of an occurrence and its level of severity.

---

[34] ISO 26262 uses the term of Failure instead of Error. We intentionally uses the term error for term consistency in this White Paper (see §1.3 and previous definitions)

A risk is associated to a maximum *likelihood of occurrence,* depending on its severity. Those values are decomposed in regions, as represented in Figure 13.

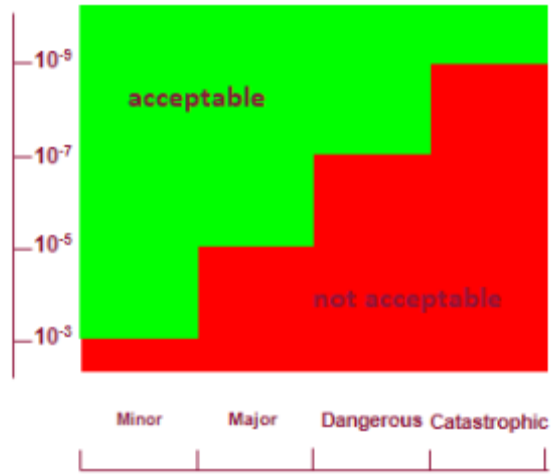

*Figure 13. Acceptable risk in safety [inspired by ARP4754A/ED-79A]. Y axis is failure rate per flight or driving hour*

When a failure rate objective is assigned to a system, the design of the system shall be done in a way that enforces that this failure rate is not exceeded. To ensure that, an analysis of the system is performed based on its components part list, and using usual reliability prediction standards (e.g., MIL HDKB 217F, UTE-C 80-811).

### 4.2.2.3 Probabilities

As identified by the author of [16], according to the safety assessment activity, the interpretation of the probabilistic measures can be *frequentist* or *Bayesian*.

The frequentist interpretation of probabilities assumes that events are likely to occur during repeated experiments. Thus, the assessment of the probability of rare or even unique events is in practice intractable with this interpretation of the probabilities. The frequentist interpretation is classically used to estimate the failure rate of components based on repeated and large-scale solicitation tests [17].

The Bayesian interpretation of probabilities encompasses the modelling of uncertainty (about a subjective belief for instance), that is by definition a notion that cannot be objectively measured by its frequency (since it is subjective). The Bayesian interpretation can be used for instance to extrapolate the failure rate of a component from a well-known context to a new one.

For the author of [16], if the considered event is likely to occur, the frequentist interpretation is preferred over the Bayesian one since it provides an objective estimation of the likelihood of the event. However, for rare events (i.e. highly severe failure conditions), the frequentist approach cannot be used. The classic safety assessment methods (like FTA or FMEA) allow computing some probabilistic quantification over such rare events thanks to assumptions about the propagation of component failures within the studied system. Thus, the probabilities measures can be interpreted as an uncertain proposition based on the subjective knowledge of the analyst and the probabilities of component failures (generally evaluated using a frequentist approach). At our best knowledge, the classical safety assessment does not quantify the uncertainties concerning the accuracy of the assessment results. More pragmatic ways have been chosen to deal with uncertainties for instance:

1. The deliberate introduction of a "pessimism" bias in the assessments.
2. The restriction of the assessment goal to the "foreseeable conditions" [18].
3. The continuous revision of the safety assessment thanks to the lessons learned on incidents and accidents.
4. The diversification of the safety analyses performed on the same system (FMEA, FTS, CMA, …).

### 4.2.3 Reminder on Machine Learning

#### 4.2.3.1 Uncertainty and risk

Uncertainty is usually defined as "imperfect or unknown information about the state of a system or its environment". Uncertainty can occur at various levels and raises two fundamental questions:

- *How to* estimate *uncertainty?*
- *How to* operate *with uncertainties? i.e. how to combine multiple uncertain observations to infer statements about the world?*

Answers to those questions are given respectively in the scientific field of "uncertainty quantification" and "uncertainty propagation".

The ISO standard 31000 (2009) and ISO Guide 73:2002 define risk as the "effect of uncertainty on objectives". So, a risk characterizes a *known* event (i.e., an event that may occur in the future) for which either the occurrence or the consequences or both are *uncertain*, i.e. *unknown*. But this constitutes only a strict subset of the full uncertainty spectrum: we generally consider also the *known knowns* (where all the elements, i.e. the event, its occurrence and its consequences are all *known* or can be inferred or predicted with very high accuracy), the *unknown knowns* (e.g. body of knowledge that had been forgotten or ignored), and the *unknown unknowns*.

Among all these classes, the last one is raising the biggest challenges since the occurrence, the consequence, and the *nature of the event itself* are uncertain. This concept is usually illustrated by the example of the *black swan*: prior to seeing (i.e. acknowledging the existence of) any black swan, it may be impossible to conceive the existence of a black swan. In this condition, not much can be said about its supposed occurrence or consequences under such circumstances...

Even though the class of unknown unknowns is infinite, given our prior knowledge about the world, we may argue that most unknown events are only theoretical, i.e. not foreseeable. Under such hypothesis, general safety objectives can be cast as an optimization problem: *maximizing the known knowns while minimizing the unknown unknowns*.

Any optimization efforts incurring costs, in practice, the theoretical optimization problem translates to a practical problem of finding a suitable trade-off between the safety requirements and the design and operational costs required to achieve them.

*4.2.3.2 Prediction goals of Machine Learning*

In the following sections, we come back to fundamental concepts used in Machine learning that were introduced in §2.1.3: loss, risk and empirical risk. We focus our attention to the common supervised learning setting.

Suppose we already have access to an offline dataset $(X_1, Y_1), \ldots, (X_n, Y_n)$, where $X_i$ denotes the attributes vector associated with the $i$-th example, and where $Y_i$ is the associated label. Depending on the application at hand, the variables $X_i$ and $Y_i$ can be discrete or continuous.

Informally, the goal is to predict what the value of $Y$ will be, given a new value of $X$. Depending on the application, we may be satisfied with a single prediction (the "likeliest one"), a set of predictions that "most likely" contain the true value of $Y$, or prediction(s) with "likelihood" score(s).

We start with the first goal (outputting a single prediction) and recall classical concepts in supervised machine learning: loss function, risk, and empirical risk.

In the sequel $\hat{f}$ denotes a *predictor* (or *machine learning model*) that maps any new input $X$ to a prediction $\hat{f}(X)$. The function $\hat{f}$ is typically chosen so as to fit the training dataset $(X_1, Y_1), \ldots, (X_n, Y_n)$ perfectly or purposely imperfectly (see below).

*4.2.3.3 Error[35] and loss function*

A discrepancy between the true outcome $Y$ and its prediction $\hat{f}(X)$ is interpreted as an error. Errors can be measured in several ways: as a binary value, noted $1_{\{\hat{f}(X) \neq Y\}}$ which equals one when the prediction is incorrect or zero when the prediction is correct, or as a measure of "how far" the prediction is from the expected outcome (e.g., $|\hat{f}(X) - Y|$ or $(\hat{f}(X) - Y)^2$).[36]

The function that measures the error, denoted $L(\hat{f}(X), Y)$, where $L: \mathbb{R} \times \mathbb{R} \to \mathbb{R}_+$ is called the *loss function*. The loss function, which is chosen by the ML designer, plays an important role in machine learning. We will address that point soon but, before, let us first consider the relation between the *loss function* and the *risk*.

*4.2.3.4 From Loss to Risk[37] (or generalization error)*

One of the main goals of supervised machine learning is to build predictors $\hat{f}$ that have very low *generalization error*, i.e., that perform well on new unseen data. This notion, which coincides with the statistical notion of *risk*, is defined as follows [2]:

> Given a joint distribution of data and labels $P_{X,Y}$, a measurable function $g$ and a nonnegative loss function $L$, the risk of $g$ is defined by $R(g) = \mathbb{E}_{P_{X,Y}}\big(L(g(X), Y)\big)$.

---

[35] In this paragraph, the term error has no link with Laprie definitions described in §1.3 and §4.2.2.1.
[36] Note that in the special case of binary classification (i.e., when both $Y$ and $\hat{f}(X)$ lie in $\{0,1\}$) all the errors $1_{\{\hat{f}(X) \neq Y\}}, : |\hat{f}(X) - Y|$ and $(\hat{f}(X) - Y)^2$ are equal.
[37] This notion of (statistical) risk has no relation with the safety risk define in §4.2.2.1.

Put differently, the risk of a predictor $g$ is the average error ("expected error") over all possible new input-label pairs $(X, Y)$ drawn at random from the distribution $P_{X,Y}$.

The *Bayes risk* is the smallest risk among all possible predictors:

$$R^* = \inf_g R(g)$$

The Bayes risk is an ideal target that can serve as a theoretical benchmark to quantify how "good" is a predictor $\hat{f}$. This leads to the concept of *excess risk,* i.e., the difference between the risk of $\hat{f}$ and the Bayes risk:

$$R(\hat{f}) - R^*$$

The excess risk can usually be decomposed into multiple terms, e.g., approximation error, estimation error, optimization error.

Unfortunately, both the risk and the excess risk are unknown in practice. Indeed, computing any risk $R(g)$ would require to know exactly the distribution $P_{X,Y}$, while in practice we typically only have access to samples from that distribution.

The only measurable risk is what is called the empirical risk.

### 4.2.3.5 From Risk to Empirical Risk

The *empirical risk* is the average loss over the dataset $(X_1, Y_1), (X_2, Y_2), \dots (X_n, Y_n)$:

> Given a dataset $(X_1, Y_1), (X_2, Y_2), \dots (X_n, Y_n)$, a measurable function $g$, and a non-negative loss function $L$, the empirical risk is defined by
> 
> $$\hat{R}(g) = \frac{1}{N} \sum_{i=1}^{N} L(g(X_i), Y_i)$$

If the $(X_i, Y_i)$ are independent random observations from the same joint distribution $P_{X,Y}$,[38] then the empirical risk $\hat{R}(g)$ is a "good" approximation of the true but unknown risk $R(g)$.

Typical machine learning models $\hat{f}$ are obtained by approximately minimizing $\hat{R}(f_w)$ over a class of functions $f_w$ (e.g., neural networks with weights $w$). An imperfect fit to the dataset (via, e.g., regularization techniques) may be preferred to favour simple models that may perform better on new unseen data.

How close the empirical and true risks $\hat{R}(f_w)$ and $R(f_w)$ are is key to understand how well a model learned from a specific training set will perform on new unseen data. In statistical learning theory, *generalization bounds* are guarantees typically of the form:

> With probability at least $1 - \delta$ over the choice of $(X_1, Y_1), (X_2, Y_2), \dots (X_n, Y_n) \overset{i.i.d.}{\sim} P_{X,Y}$ for all weights $w$
> 
> $$R(f_w) \leq \hat{R}(f_w) + \epsilon(n, w, \delta)$$

---

[38] We say in this case that *the dataset $(X_1, Y_1), (X_2, Y_2), \dots (X_n, Y_n)$ is drawn i.i.d. from $P_{X,Y}$.*

These bounds imply that a model $f_w$ with small empirical risk $\hat{R}(f_w)$ also has a small risk $R(f_w)$, that is, it performs well on new unseen data.

Examples of $\epsilon(n, w, \delta)$ or related quantities are given by VC-bounds, Rademacher complexity bounds, PAC-Bayesian bounds, among others. It is however important to note that most historical bounds do not explain why deep learning models perform well in practice (the term $\epsilon(n, w, \delta)$ is large for practical values of n). Improving such bounds towards distribution-dependent or structure-dependent generalization bounds for neural networks is the focus of renewed theoretical interest.

### 4.2.3.6 The importance of the loss function

The choice of the loss function $L$ during the design of the ML algorithm is very important for several reasons.

First, the loss function is used to measure the prediction performance of the machine-learning model $\hat{f}$. As the objective of the learning process it to maximize this performance (or minimize the loss), the loss function formalize the learning goal and different loss functions may correspond to different goals. For instance, choosing a loss function that penalizes harmful prediction errors may be pertinent if the ML-based component is used in a safety-oriented system.

Other technical aspects must also be considered when choosing the loss function:

- The computational power required to minimize the loss on a given training set depends on the loss function.
- The ability of a machine learning model to generalize (i.e., to perform well on previously unseen data) and its robustness to distribution changes could depend on the loss function.

### 4.2.3.7 From loss function to performance metrics

We outline below some classical and simple examples of performance criteria that are asymmetric and could provide hints on how to design a safety-relevant loss function.

In the context of hypothesis testing (which can also apply to binary classification, where $\in \{0,1\}$), machine learning terminology makes a clear distinction between two types of errors. The type I error is the rejection of a true null hypothesis (also known as a false positive). For binary classification, such an error occurs when $Y = 0\ but\ \hat{f}(X) = 1$. The type II error is the non-rejection of a false null hypothesis (also known as a false negative). For binary classification, such an error occurs when $Y = 1\ but\ \hat{f}(X) = 0$.

Other metrics/indicators can be used along with a loss function. Such indicators depend on the targeted problem, and can highlight or hide some behaviours (as every statistical measure). For example, for classification tasks, Precision, Recall and F1-Score are often used to assess the performance of an algorithm [19] and [20] (see Figure 14).

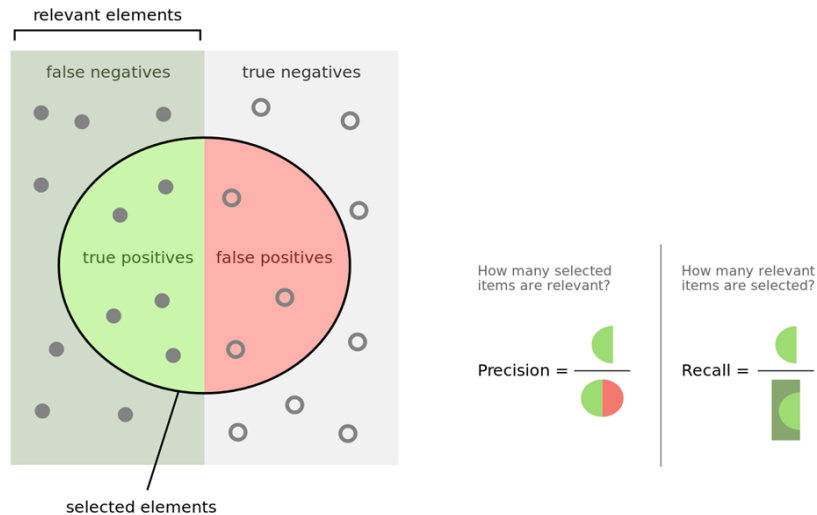

*Figure 14. Basic metrics in AI models (from Wikipedia).*

In short, for a classifier, when optimizing Precision, designers try to avoid being wrong when identifying true cases (in order to reduce the occurrence of false positives). Whereas, when optimizing Recall, designers try to be very good at discriminating true cases from others (in order to minimize the situation of missing a true case). Optimizing F1-score ($F_1 = \frac{2 \cdot precision \cdot recall}{precision + recall}$) means you do not want to choose Recall over Precision or the other way and find the best trade-off.

#### 4.2.3.8 From dataset to reality

ML metrics are computed on a particular dataset, which means that the performance measure is only an empirical assessment of the true performance of the algorithm (named training error, $Err_{train}$). In fact, the true performance of the algorithm, named *operational performance* (or also test error, $Err_{test}$) will be probably very different from the empirical one, and this performance will be obviously worse as the real-life diversity challenges the model ability to "infer" a good result when dealing with unseen cases.

#### 4.2.3.9 Machine learning and probabilities

*Probabilities in the ML world*

As stated above, ML modeling is stochastic by nature and in most cases can be recast in probabilistic terms. The probabilistic assessment of ML models crucially depends on various factors, such as the design of the experiment (e.g. training/validation), the eventual input probabilities (priors) and their interpretation.

It is rather easy to mechanically apply transformations to some numerical data (e.g. softmax) to further comply with probability calculus axioms ('non negative' and 'unit measure'). However, without careful considerations, once performed such operations, the probabilistic interpretation given to the final results may be very incorrect. For instance, it is well known that you can find adversarial example with high softmax output, see Figure 18).

ML modeling aims at building function approximations f such that f(x) approximates y, where x and y are respectively input and output data samples. The function f is parametrized by some set of parameters z that will be inferred from data. By defining a likelihood function this problem can be recast as an optimization problem. The ML literature proposes two major formulations of it, depending on the nature we assign to the parameters z. If we posit that the 'true' parameters are single values, the problem can be formulated as a maximum likelihood estimation (MLE) one. Alternatively, if we treat the 'true' parameters as random variables, the problem is stated as maximum a posteriori estimation (MAP). We thus emphasize that the probability assessment of any ML model depends on these choices (by means of which we find the aforementioned frequentist or Bayesian interpretations of probabilities).

Finally, we recall that most classical results in probability theory are deduced from the 'law of large numbers'. However, in many practical scenarios this provides only a very crude approximation of the studied phenomena and some hypotheses necessary for a sound probabilistic assessment may not hold (see for example the "independent and identically distributed" (i.i.d.) hypothesis).

In light of the above considerations, we argue that we should proceed with great care whenever we perform a probabilistic assessment of ML models. Without the necessary precautions, it may provide very misleading conclusions. This is particularly important in the case of phenomena characterized by very rare events, which are of special interest for any sound risk assessment.

*Calibrated probabilities*

Let us consider an arbitrary ML classifier capable of associating a probability distribution to its outputs (labels). Intuitively this probability may be interpreted as the confidence of the classifier's decision. Unfortunately, unless particular care is taken, this interpretation is usually wrong. As argued previously, it is rather easy to transform some arbitrary distribution as to satisfy in principle the axioms of probability calculus (e.g. applying a softmax). Nevertheless, this will not guarantee that the calculated probability will conform to some true observations. An alternative method, named conformal prediction [21], aims to provide such alignment. Conformal prediction uses experience to determine precise levels of confidence in new predictions. Some of the advantages of such a method are that it can be applied to any ML model capable of outputting a probability distribution and that it does not require re-training the model.

This process is similar to a calibration process, in this case the aim being the calibration of predicted probabilities to some observed probabilities. As in typical calibration procedures, this methodology requires a different dataset (calibration dataset) used solely for this purpose (i.e. distinct from the training and validation datasets).

*Probabilistic properties*

As a general statement, probabilistic assessment aims at providing probabilistic bounds for certain numerical quantities that may be deemed critical from a safety perspective. Sometimes, these quantities cannot be measured directly, and the designer may seek to ensure certain properties for the system / model (e.g. robustness). In most of such cases, providing strict bounds for such quantities / properties is generally an intractable or impractical problem, especially when no internal knowledge of the application domain or the model is taken into account. In some contexts, imposing soft constraints on the desired objectives may transform the problem into an easier one (e.g. proving robust guarantees with $(1-\epsilon)$ probability). We think that such probabilistic formulations can be appropriate to address certain properties to better support safety-aware decisions.

#### 4.2.4 Mixing the two worlds

*4.2.4.1 Uncertainty and risk*

**Issue 1.1:** How to estimate uncertainty?

As presented previously, the safety risk is usually assessed by taking into account two factors: (i) the probability of a critical event occurrence (e.g. a failure); (ii) the expected harm resulting from that event. Estimating each factor raises difficulties that are not specific to but enhanced in the case of ML-based systems (unknown/unknown):

1. How to identify *all* the possible events (not only failures)?
2. How to estimate the probability of occurrence of *each* event?
3. How estimate the harm associated with *each* event?

Safety solutions can be seen as a trade-off between the design and operation costs to handle uncertainties on the one hand and the potential harmful consequences of the system's behaviour on the other hand.

**Issue 1.2:** How can we relate the different notions of probability in the ML and safety domains?

As shown earlier in this section, ML models are evaluated by various metrics usually obtained by testing the model on validation dataset(s), e.g. the error rate in the case of a classification task. Given the fundamentally stochastic nature of any ML model, its outputs will be affected (at least partially) by random errors. As argued in §4.2.2.2, ML-based components are affected by both systematic and random errors. If the current software design standards (e.g., ISO26262 part 6) were to explicitly address the latter type of errors, the (acceptable) error rate of the model should be part of the system specifications, similarly to the specifications of hardware components.

As required by the secondary question (Q2), stated in the introduction of this section, this proposition mandates us to provide the necessary means to translate the performance of the model (evaluated on a dataset) to the operational performance of the system (evaluated during its operation).

*4.2.4.2 Loss function*

**Issue 2:** Usually, in ML, the loss functions do not distinguish the misclassifications depending on their safety effect. Let us consider the example of Pedestrian Detection in images from cars' front camera. A misclassification occurs when a pedestrian is not detected (which in terms of safety is catastrophic as the person could be injured) or when a pedestrian is wrongly detected (which in terms of safety is solely major, as it could lead to a useless abrupt braking).

**Result 2:** The author of [22] proposes a formal and domain-agnostic definition of the risk, identified as the central notion in safety assessment. According to this paper, the main objective of safety is risk minimization, whatever the explicit definition of these risks (usually defined in terms of mortality). The risk here is the expected value of the "cost of harm", highlighting that the exact cost is not known but its distribution can be available. The author assumes that it exists a function able to quantify the harm of a mis-prediction in a given context ($L: X \times Y^2 \rightarrow \mathbb{R}$). The "safety" of a given model can then be probabilistically assessed by providing the harm expectation of a model h as $\mathbb{E}[L(X; h(X); Y)]$. A safe model is thus a model minimizing such harm expectation and can be obtained by integrating the harm measure in the loss function.

**Limitation 2:** This definition of the risk is a bit different from the one proposed in the aeronautic standard ARP4754A/ED-79A, which relies on the notion of severity and probability (or frequency). Perhaps, an alternative definition of the risk incorporating the "formal aspect" (i.e. definable as a function) of [22] but linked to the notion of severity would be a good starting point. Such a definition would enable to express classical safety requirement such as "safety reserve" (safety margin, safety factor).

### 4.2.4.3 From dataset to reality

**Issue 3:** With ML, probabilities are assessed on the dataset and not on the real environment thus cannot be used, as is, as safety indicators.

**Result 3:** Fortunately, for binary classification with 0-1 loss, Vapnik and Chervonenkis and other researchers have stablished the following theorem (VC theorem) that bounds the probability of having an operational error[39] greater than the sum of the empirical error and the generalization error ($\delta$). $\delta$ depends on the number of samples ($N$) used during training, and the complexity of the model ($VC$), and the confidence level $\varepsilon$ :

$$P(Err_{test} \leq Err_{train} + \delta) \geq 1 - \varepsilon \text{ where } \delta = f(VC, N, \varepsilon)$$

The model complexity $VC$ – or VC dimension – is unfortunately really hard to evaluate for most ML models (this is an ongoing research). For instance, the VC dimension for a 10-12-12-1 Multi-Layer Perceptron is the number of network parameters, here 301.

An example of upper bound of $\delta$ is given by the following relation:

$$\delta \leq C_1\sqrt{\frac{VC}{N}} + C_2\sqrt{\frac{\log(1/\varepsilon)}{N}}, \text{ for some absolute constants } C_1 > 0 \text{ and } C_2 > 0$$

In order to exploit the VC-theorem, the two members have to be in the same order of magnitude:

1. $\text{Err}_{test}$: the same order of performance during training must be obtained as the operational one.

2. **$\delta$**: the first part of the upper bound of **$\delta$** depends on the model complexity, and the number of samples in the training set N. Numerically, to obtain an upper bound of $\delta \approx 10^{-5}$ with a model where $VC = 10^5$ (deep learning example), the number of samples N must be at least around $10^{15}$. This result shows that this approach is very limited, but it also pushes to obtain more accurate theorems and properties than those used here. Research is on-going, and some tighter error bounds have been proved in some cases leading to non-negligible yet smaller values of $N$.

**Limitation 3:** the previous approach is theoretically correct but practically intractable. Thus, there is currently no solution to tackle the open question of linking probabilities of errors on the dataset to the reality. For instance, if we want to develop a pedestrian system which will have a Recall of 0.999999 over 1 000 000 samples, meaning that we miss 1 pedestrian

---

[39] Reminder, in this paragraph, the term error has no link with Laprie definitions described in §1.3 and §4.2.2.1, but correspond to the standard definition in theory of Machine Learning

every 1 000 000 cases, we have to make an assumption of how close our empirical error is from our operational error.

### 4.2.4.4 Interpreting probabilities

The relationship between the likelihood of the model performance and the "acceptable failure rates" of safety is established by the distribution over the time of the events to be observed.

For example, for a pedestrian recognition:

- AI model: images are given to be recognized, and the performance corresponds to a number of recognized images, vs. a total number of images
- Safety: a number of failures per hour of operation

Convergence of the two: it is necessary to model a number of solicitations per hour of operation; therefore, a number of obstacles to be recognized per hour.

Numerically, it gives: in the city, we can say that we have an obstacle per minute to recognize; a performance of $10^{-3}$, therefore gives $60 \cdot 10^{-3} = 6 \cdot 10^{-2}$ failures per hour of operation; in the countryside, we can say an obstacle every hour, therefore $10^{-3}$ failures per hour.

**Issue 4:** Mixing probabilities of the two worlds: importance of interpretation

**Result 4:** Papers [23] and [24] discuss "the meaning of probability in probabilistic safety assessment". While probability calculus is well established in the scientific community, the interpretation of probability is hotly debated to these days. Numerous interpretations of probability have been proposed and contested. Amongst the most popular ones, we can mention the classical, the frequentist, the subjective, or the propensity interpretations. While the classical or frequentist ones "make sense" for game theory (where outcomes can be enumerated and evaluated a priori), they are very debatable when applied to domains where events are very rare or unknown / unseen in the past (e.g. black swans, see §4.2.2.3). In these cases, a "subjective" interpretation of probability is more in tune with our intuition: it expresses degrees of belief of an agent w.r.t. possible events. While this interpretation has its own problems, we think that it is the most appropriate one to use in case of uncertainty quantification and propagation for a safety analysis. In any case, we highlight that any probabilistic assessment must state explicitly what type of interpretation is assigned to the probabilities used by the analysis. We think that the probability is not an intrinsic property of the world and should be used as a tool to build arguments in view of certification.

### 4.2.5 Conclusion

Software verification and validation methods typically used for certification purposes are limited in their abilities to evaluate ML-based software with respect to safety objectives.

We think that in order to embed with confidence ML models in safety critical software, the certification and ML communities must work together to develop new tools, methodologies and norms to provide *rigorous probabilistic assessment of a system integrating a ML-based software in a safety perspective*. In this context, we highlight the inherent stochastic nature of ML-based software and we argue that existing certification norms such as ISO26262:2018 or ARP4754A/ED-79A or DO178C/ED-12C do not properly address the new types of errors introduced by this type of software. In an effort to reconcile the concepts and objectives of the two communities, we draw parallels between them and formulate some of the major challenges that must be addressed from a probabilistic perspective. Those challenges are recalled hereafter.

### 4.2.6 Challenges for probabilistic assessment

For the Probabilistic Assessment theme, we identify the following challenges:

- Challenge #1.1 Definition of environment / context / produced outputs / internal state.
    - How to identify all the possible events (not only failures)?
    - How to estimate the probability of occurrence for each event?
    - How to estimate the harm associated with each event?
- Challenge #1.2: How to propose definitions of the risk incorporating more concepts (such as the estimation of harm effects) to express classical safety requirement such as "safety reserve" (safety margin, safety factor)?
- Challenge #1.3: How to propose appropriate loss functions taking into account safety objectives?
- Challenge #1.4: How to make the link between probabilities assessed on the datasets and on the real environment (to be used as safety indicators)? Or, what are the possible means to translate the performance of the model (evaluated on a validation dataset) to the operational performance of the system (evaluated during its operation)?
- Challenge #1.5: How to find tractable methods to assess the operation error w.r.t. the empirical error?

## 4.3 MAIN CHALLENGE #2: RESILIENCE

### 4.3.1 Introduction

In this White Paper, we use the term "resilience" as a synonym of "fault tolerance", i.e., "the ability for a system to continue to operate while an error or a fault has occurred". Note that the usual definition in the domain of dependable systems is slightly more specific: resilience is "the persistence of service that can justifiably be trusted when facing *changes*." [25], [26]. Changes include unexpected failures, attacks, increased load, changes of the operational environment, etc., i.e., "all conditions that are outside the design envelope". Considering the changes of the operational environment is also of major importance in the case of ML systems, especially for those interacting with a complex, evolving environment (e.g., roads).

Sticking to the first definition, what are the "faults and errors[40]" to be considered in a ML system?

A ML component may be subject to various accidental or intentional faults, all along its lifecycle, during data collection, model design, training, implementation… In this chapter, we focus on the following two categories of faults that can appear during operation:

- o Operating conditions inconsistent with those considered during training.
- o Adversarial inputs.

Those faults may also be considered as part of robustness analysis (see challenge #6 Robustness, §4.7) but, here, we consider that the limits of robustness have been exceeded and that mitigation means at system level are required.

### 4.3.2 Monitoring in ML and runtime assurance

Many of the challenges identified in this White Paper show the difficulty to ensure *design-time* assurance and this chapter points out the necessity to consider *runtime* assurance. The main idea is to surround the unsafe ML component with a set of low-level monitors and controllers (e.g., "safety nets") ensuring the safety properties by detecting and mitigating the hazardous behaviours of the unsafe component.

Monitors check if the system stays in its nominal functioning domain by observing its inputs (e.g., their ranges, frequencies…), its outputs, or some internal state variables. They can use assertions, memory access check, "functional monitoring", etc. Some methods like the one proposed in [27] propose an adaptation of the simplex architecture to the monitoring of a neural controller, using a decision block able to detect an error and to switch from the neural controller to a high assurance – but less performant – controller.

Monitoring the system inputs can be used to ensure that the operational conditions match the learning conditions. In [28], a technique is proposed to monitor the distance of the input/output distribution observed during operation with the one observed during the learning phase, and raise an alert when this distance exceeds some threshold. Others, such as [29], propose to collect neuron activation patterns during the learning phase and, thanks to online monitoring, detect the occurrence of an unknown activation pattern possibly indicating an erroneous prediction. However, such monitors (out-of-distribution detection, or abnormal activation patterns) are still on-going research topics, and their reliability is not guaranteed.

---
[40] A definition of those terms is given in §1.3.

Some on-going works [28], [29] propose a runtime verification approach based on an adversarial runtime monitor, i.e., a mechanism based on the generation of adversaries at runtime aimed at assessing whether the ML system is able to provide a correct answer.

### 4.3.3 Ensuring resilience

Most of the techniques applicable to classical (non-ML systems) are also applicable to ML systems; we present *some of them* in the next paragraphs.

**Redundancy**[41] can be used to detect or mask faults/errors. Redundant architectures rely on spatial redundancy (i.e., the component is replicated with or without dissimilarity) and/or temporal redundancy (i.e., the behaviour is replicated on the same component), depending on the faults/errors to be tolerated. Typical redundant architectures involve two (duplex) or three (triplex) components, depending on whether safety and availability is at stake: with two components, the system may detect an error and switch to a safe state whereas with three components, the system may stay operational (fail-op) and possibly detect and exclude the faulty component or possibly place it back into a non-erroneous state. To be applied for ML-based systems, the difficulty resides in the cost of these strategies in terms of computing resources and/or computation time and in the prevention of Common Mode Faults.

**Dissimilarity** may be used to prevent Common Mode Faults, but dissimilarity is difficult to ensure for ML components since it requires the capability to have multiple solutions to the same problem (multiple ML techniques, multiple model architectures, etc.) depending on the fault models (see Table 5). Unfortunately, ML technique are generally used to implement functions for which no non-ML solution is actually applicable *with the same level of performance*. Providing dissimilar ML solutions and being able to estimate to what extent Common Mode Faults are prevented, is clearly a challenge. Assuming that multiple and dissimilar solutions are available, **voting** may be used to mask errors. For example, if three components produce a binary output, the one provided by at least two out of the three components is selected.

As already presented, **monitoring** may be used to detect error by verifying some correctness properties on the system. Monitoring may be based on different forms of redundancies, from a complete functional redundancy (the monitoring component performs the same function as the monitored component) to some partial redundancy (verification of some invariant, computation of an inverse function, etc.). Again, systems using ML are usually difficult to monitor: independence of redundant ML solutions is difficult to demonstrate, inverse functions cannot be computed, invariants are difficulty to express, etc. In some cases, a solution consists in using a classical implementation (ensuring safety) to monitor a ML implementation (ensuring other performances).

---

[41] Generally speaking, "redundancy" refers to the property of some element (e.g., software or hardware component) that can be removed from a system without modifying the function performed by the system. Hence, a monitoring mechanism, or an error detection code is actually a form of redundancy.

*Table 5. Proposition of dissimilarities between ML components*

| Dissimilarity origin | Comment | Resilience with respect to… |
|---|---|---|
| Training dataset | MLs components are trained using different datasets | …training set corruption<br>Noise mitigation |
| Model | Different kind of models are used (ex : decision tree and neural net) | …specific model vulnerabilities |
| Implementation | Models are embedded using different framework | …faults due to development framework |
| Inputs | ML components can take as inputs different sensor chains | …local weaknesses of the ML component |
| HW | Models are not embedded in the same HW component | …HW failures |

### 4.3.4 Relation with other HLP and challenges

Monitoring being one of the system-level solution to ensure safety, resilience is tightly linked to the **monitorability** HLP. As the provision of explanations may be a means to monitor a system (in the sense that "if no explanation can be provided to a decision, the decision is deemed suspicious"), resilience could be also linked to **explainability** but, in practice, this approach seems hard to apply. Anyway, non-explainable models *may be* harder to monitor.

Resilience is also strongly coupled with the various forms of **robustness:**[42] intrinsic robustness – i.e., the ability to guarantee a safe functioning in the presence of input data different from training data –, robustness to attacks, or robustness to classical implementation problems. As explained in §4.3.1, resilience is a property required when input are beyond the nominal operating range of the ML component, as shown on Figure 15 below:

- The green area is the nominal work domain where the ML component delivers the expected service.
- The light orange area represents the extended work domain ensured by the robustness property.
- The dark orange area represents situations where the bounds of the robustness domain have been exceeded. The ML component fails but system-level mechanisms prevent a global system failure.
- The hatched red area represents situations where the bound of the resilience domain have been exceeded. The system fails.

---

[42] Robustness is addressed in details in Section 4.7.1.2. Note that, in this document, we only consider robustness at the level of the ML component.

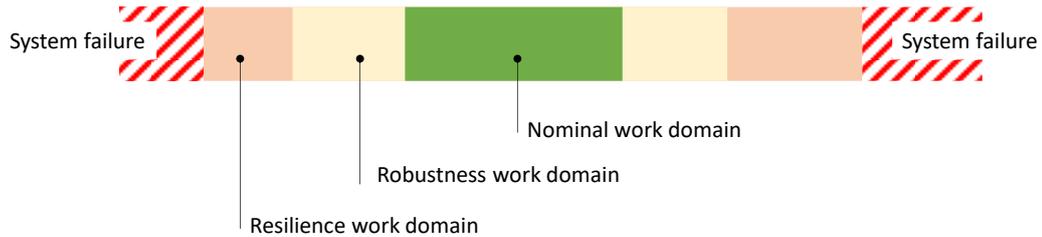

*Figure 15. Illustration of robustness and resilience*

### 4.3.5 Challenges for resilience

Resilience is certainly a required property of ML-based systems, in particular when considering the lack of robustness of some ML techniques (e.g., artificial neural networks) to some adversarial attacks. However, ensuring resilience for a ML-based system raises various challenges:

- Challenge #2.1: How to detect erroneous or out-of-domain inputs of a ML model, in particular when considering the possible failures of the acquisition system or the presence of intentionally corrupted data?
- Challenge #2.2: On what system architectures can we rely to ensure the safe operations of ML–based systems?
    - Challenge #2.2.1: Can we rely on Runtime Assurance? How to monitor ML-based systems?
    - Challenge #2.2.2: Can we rely on multiple dissimilar ML-systems? And how to assess dissimilarity?
- Challenge #2.3: How to create ML models "robust by design", to reduce the need for resilience requirement at system level?
- Challenge #2.4: How to create a confidence index characterizing the proper functioning of the ML component?
- Challenge #2.5: How to recover from an abnormal operating mode in an ML system?

## 4.4 MAIN CHALLENGE #3: SPECIFIABILITY

### 4.4.1 Introduction

For avionics systems, ARP4754A/ED-79A defines specification as "a collection of requirements which, when taken together, constitute the criteria that define the functions and attributes of a system, component, or item". The specification process supports the capture of the intended function of the system in terms of functional, safety, operational and environmental aspects as well as the implementation and verification processes. The concept of "intent", which was introduced by the working group on Overarching Properties for avionics [37], helps to answer to the following question: "are we sure that the development team has understood what they were supposed to develop?". The aim is to be sure that the specification is correct and complete with respect to the intended behaviour for the system.

Then Specifiability becomes the ability to capture the needs and constraints of a system.

### 4.4.2 Impact on certification

#### *4.4.2.1 Why does the specification of ML based Systems raise new challenges?*

The development of Machine Learning (ML) based Systems differs from traditional system development in several aspects of its lifecycle, the specification step in particular must be adapted accordingly.

In the development of classical systems, for the development of a new system, the specifications are usually set up using an iterative loop between the expression of system needs, and the choices of the implementation solutions. Like with the SOTIF [30] that proposes an iterative approach to reduce the occurrence of the unknown and unsafe operational scenarios, this loop remains for ML systems but with an increased "complexity".

One of the challenges of specification for a ML based system is to demonstrate that the expected behaviours and constraints (in terms of functional, safety, operational and environmental aspects) are actually contained in the specification. Pedestrian detection is one classical example: even though ML algorithms perform pretty well at detecting pedestrians (in some conditions), defining unambiguously what a "pedestrian" is remains an open issue.

Let us elaborate on that problem.

According to Merriam-Webster, a pedestrian is "a person going on foot". Clearly, this excludes persons on wheeling chairs.[43] So, the specification may be rewritten using the word "person" instead of "pedestrian" in order to reflect the actual intent. The word "person" now reflects the intent but is very ambiguous. Looking for "person" in the dictionary leads lead to "human", which does not help that much. At the end of the day, we end up replacing one term by (supposedly) more primitive terms, without ever reaching a fixed point. Note that this problem does not exclusively affects ML. Indeed, any non-mathematical definition suffers from this kind of ambiguity. But this issue shows up more particularly in domains

---

[43] Note that we may consider that, in this precise case, the specification was simply invalid and not only ambiguous.

where ML excels, i.e., domains where there exists a large semantic gap between the concept expressed in the specification (e.g., a "human" or some "opinion") and the data actually available to the system (an "image" or some "piece of text").

Faced to such difficulties, the training dataset may be used to complement a set of explicit specification items, possibly iterating with the expression of the system needs (for instance removing loopholes that the ML system may be able to learn). In some cases, data may even be the only way to specify the intended function (during the initial training phase and even iteratively during *future* trainings). In addition, the dataset may also introduce unexpected behaviours. For these reasons, the dataset must be considered as a part of the system specification. Therefore, the dataset shall at least comply with the quality requirements applicable to classical specification items: correctness, completeness, compliance, consistency, accuracy, and so on. But other properties such as fairness, representativeness of the real environment, etc. may be necessary. For instance, representativeness is necessary to ensure the completeness of the specification. Therefore, the specifiability challenge is tightly linked with challenges related to the data quality and representativeness.

In conclusion, if the completeness of a specification is usually difficult to achieve and to demonstrate – as shown with the "pedestrian detection problem" – this difficulty is even greater when dealing with some of the specific problems addressed by ML. Therefore, to validate the final specification, i.e., to give confidence that it complies with the system needs and with the safety constraints is definitively a specific challenge for the development ML-based systems.

### 4.4.3 Link with HLPs

"Specifiability" is related to other HLPs:

- Robustness, because the behaviour of the ML based system in front of abnormal conditions should be specified at system layer.
- Data quality, because the System specification should be the basis to assess the quality of the data and because the data could be part of the specification as explained in the previous section.
- Verifiability, because the System specification would be the basis to the verification.

### 4.4.4 Challenges for specifiability

Several challenges can be identified concerning specifiability:

- Challenge #3.1: How to identify the additional behaviours introduced during training in order to:
    - Complete the system specifications?
    - Assess the potential safety impact?
    - Accept or to reject those additions?
    - Assess robustness?

- Challenge #3.2: What criteria could be used to close the iteration loop on system specification during training to take into account the behaviour that could have been added during this phase?

## 4.5 MAIN CHALLENGE #4: DATA QUALITY AND REPRESENTATIVENESS

### 4.5.1 Introduction

Although Machine Learning through graphical and model-driven approaches is a promising field of research [31], most Machine Learning achievements are based on a data-driven approach [32]. Data-driven decision making refers to the practice of basing decisions on the analysis of data rather than on intuition [33] or on rules.

Since data must be adapted to the intended task [34], [35], the issue of guaranteeing the *quality* and the *representativity* of the data when creating a dataset for a given task is central. The following paragraphs address these two properties in details.

### 4.5.2 Data Quality

*Data quality* is the "fitness [of the data] for the intended purpose" [34]–[37]. This definition can be refined as "the absence of the defects/noise in the dataset (biases, outliers, erroneous data, imprecise data, etc.)" [38].

In the 1950s, several definitions of data quality were proposed: "the degree to which a set of inherent characteristics fulfil the requirements" (General Administration of Quality Supervision, 2008); the "fitness for use" [39]; the "conformance to requirements" [40]. Later, with the development of information technology, focus was placed on the evaluation criteria to assess data quality standards [41]. In the 90's, scholars proposed to characterize quality according to several "quality dimensions" aiming at decomposing the problem of verifying the quality of data with respect to some specific purpose [41], [42].

Quality dimensions concerned properties such as Accuracy (in data generation), Accessibility, Completeness, Consistency, Relevance, Timeliness, Traceability, and Usability. Other properties such as Provenance and Interpretability address the problem of providing reasons to trust data that can have different sources. The properties to be considered and the relative importance given to a specific property are definitively problem-dependent [36], [37].

But, once a specific subset of quality dimensions is retained, how do we measure and assess the quality of data?

#### 4.5.2.1 Metrics for data quality

Assessing the quality of data requires an evaluation framework for asserting expectations, providing means for quantification, establishing performance objectives, and applying an oversight process [42]. Note that the activities for data quality evaluation should be carried out before the selection of the subsets for a Machine Learning task. These activities should focus on different aspects of data collection, and on the representativeness of the dataset. The metrics, or means of quantification, depend on the dimension under scrutiny:

- **Accuracy** depends on data gathering/generation and measures the faithfulness to the real value. It also measures the degree of ambiguity of the representation of the information.
- **Accessibility** measures the effort required to access data [35].
- **Consistency** measures the deviation of values, domains, and formats between the original dataset and a pre-processed dataset [43].

- **Relevance and Fitness**, with two-level requirements: 1) the amount of accessed data used and if they are sufficient to realise the intended function and 2) the degree to which the data produced matches users' needs [35].
- **Timeliness** measures the "time delay from data generation and acquisition to utilization" [35]. If required data cannot be collected in real time or if the data need to be accessible over a very long time and are not regularly updated, then information can be outdated or invalid.
- **Traceability** reflects how much both the data source and the data pipeline are available. Activities to identify all the data pipeline components have to be considered in order to guarantee such quality.
- **Usability** is a quality bound to the credibility of data [35], i.e. if their correctness is regularly evaluated, and if data exist in the range of known or acceptable values.

#### 4.5.2.2 Means of quantification

The first step of the assessment process is to set the goals of the data collection and define the characteristics of the data for each quality dimension. This, in order to evaluate the data quality and eventually set a baseline that fits the desired goals [35].

Data quality and its evaluation depend on the application and its environment. For instance in social media data, timeliness (and sometimes freshness) and accuracy are the most relevant features to assess the quality of data.

To monitor and trace raw data, additional information are needed because multiple data are often queried from different sources, with no guarantee of coherence [44]. In particular, data quality shall vary depending on the degree of structure the data embed. It is intuitive that dimensions and techniques for data quality depends on the types of data: fully structured, semi structured (when data has a structure, which has some degree of flexibility, e.g. XML), or unstructured. Obstacles related to unstructured data are discussed later in the document.

The data processing methods and procedures involved in the implementation of a ML component must also comply with the applicable development assurance practices [9]. Obviously, data quality activities are also required for the development of ML components, from the acquisition of data to their eventual composition and correction. This active research field has proposed for instance composition algebra for data quality dimensions. Frameworks like the DaQuinCIS [10] provides support for data quality diffusion and improvement, and proposes detection of duplicates in datasets and data cleaning techniques. Finally, confidence can be increased by means of manual activity such as independent scrutiny of datasets, for example by measuring the same attribute in different ways or through reliability analysis. The empirical assessment of the data quality is another mean to increase trust in the dataset, even if it implies the challenge of collecting the "true" value of a data item, which may be unknown.

#### 4.5.2.3 Outliers

In [45], an outlier is defined as "a pattern in data that does not conform to the expected normal behaviour". Outliers are incorporated sometimes in the definition of noise (intended as data inaccuracy) [46]. Outliers can represent issues for some analysis methods [47], but can be relevant for others applications, or even can arise from correct data points, and be catered for with robust analysis techniques [38][48]. Thus, outliers detection is also application dependent.

*4.5.2.4 Missing data*

Missing data implies "repair" activities that must be scrutinised [49]; from data augmentation, to manually improving data collection. The direct maximum likelihood and multiple imputation methods have received considerable attention in the literature [50]. These methods are theoretically appealing with regard to traditional approaches because they require weaker assumptions about the cause of missing data, producing parameter estimates with less bias and more statistical power [51].

Missing data can occur in a dataset for several reasons; e.g. a sensor failing, or lack of response from sampled individuals. Response data bias, and low statistical power are the two main consequences of missing data [51].

- The response bias occurs when the results deriving from the collected data are different from the ones that would be obtained under a 100% response rate from sampled individuals.
- A low statistical power means that the data sample is too small to yield statistically significant results.

The degree of stochasticity involved in the data absence has an impact on the nature and magnitude of the deriving bias. For instance, the systematic absence of data generally leads to a greater bias in parameters estimates than a completely random pattern [51].

### 4.5.3 Data Representativeness

The concept of *data representativeness* refers in statistics to the notion of sample and population [52]. A sample is representative of a population if it contains key characteristics in a similar proportion to the population [53]. Transposed to the world of Machine Learning, the sample corresponds to the dataset available for the development of the model (training, validation, testing), and the population corresponds to all possible observations in the field of application.

The subject of data representativeness has been debated for many years and remains a subject on its own. As mentioned in the introduction, representativeness is a concept inherited from the statistical field, and is a qualifier of a sample (dataset). It appears that the definition of a representative sample depends on the nature of the sampling design (probabilistic or not) [54]. If the distribution of the population is not known, it is impossible to demonstrate that as sample drawn from this population is representative of this population.

This limit also naturally applies to Machine Learning and ensuring representativeness is still a challenge today: how to ensure that the characteristics of a learning, validation or test dataset strictly respect those of the environment from which they were extracted? We provide no complete answer in this document, but we propose to focus the discussion on three specific topics, important for future research:

- Assessment of representativeness.
- Obstacles related to unstructured data.
- Fairness.

*4.5.3.1 The challenge of assessing representativeness*

According to [54], the polysemy of "representativeness" makes its precise assessment impossible. Nevertheless, the assessment of the representativeness of a sample can be approached by (i) the evaluation of the sampling design, (ii) the construction of sampling plans, and (iii) the quality control of the sampling. Ensuring representativeness boils down to complying with certain key criteria in the realization of each of the following three steps [55]:

1. Determine data quality objectives regarding the problem being addressed.
2. Prepare the sampling plan.
3. Evaluate the quality of the sample plan.

The objective of the first step is to define a set of questions allowing to identify clear objectives, population, and issue to build a sampling plan. The definition of the population relies on the Operational Design Domain[44] in which the Machine Learning model is trained, validated, and tested is crucial. If the operational design domain is not available/ fully known, the usage of the model must be restricted.

The second step depends directly on the results of the first step. The type of confidence (statistical or core business) will also have an influence on the design of the sample [55], and on the generalization capabilities.

When statistical confidence is required, random sampling, or any other probabilistic sampling, may be used to build a sample that reflects the characteristics of the population. As explained in Section 4.5.3.2, this approach is difficult to implement in the case of unstructured data.

When the confidence is based on the judgment of an expert, the final quality of the sample will be the responsibility of the professional and will not require random sampling. Thus, it is an expert who validates the scope of the operational design domain and evaluates the relevance of the sample collected.

Depending on the knowledge of the population (see step 1), multiple sampling methods can be used. We describe them briefly below, based on [56]:

- Probabilistic methods (confidence based on statistics):
    - **Simple Random Sampling**: Observations (units of data to be analysed) are collected according to a random process. Each observation of the population has the same chance of being drawn in the sample. Depending on the answers to the questions in Step 1, sampling must take into account the spatial and temporal variability of the population (domain of use). Theoretically, the random sample should contain the same proportions of each type of observation as in the population, even the rarest ones. In practice, biases may occur, for example, because of an inability to achieve a truly random method, or because there is a probability of over-representation of certain groups.

---

[44] The concept of "operational design domain" (ODD) is defined in SAE J3016. It denotes the operating conditions under which a given system is specifically designed to function.

- **Clustered sampling** presupposes that the operational design domain (the population) is homogeneous in its composition. The domain is then segmented into "clusters", a subset of the clusters is randomly chosen and all observations from the selected clusters are sampled.

- **Stratified sampling** is based on dividing a heterogeneous population into mutually exclusive strata and on applying probability sampling in each stratum (the simple random method being the most common). Stratification is carried out with respect to a variable of interest that is often simple and categorical (in the case of societal studies: gender, age group, nationality, etc.). This method has the advantage of maintaining in the sample a proportionality identical to that of the population for a variable of interest.

- **Systematic sampling** assumes that the entire population is known, and defines sampling intervals according to the desired sample size. Thus, an observation will be sampled every k observation, k being the ratio between the number of possible observations in the population and the size of the desired sample. This method may be of interest to ensure the uniformity of the quality of a production line, for example.

- **Multiphase sampling** combines the methods presented previously, for instance by performing random sampling for each of the clusters obtained by clustered sampling.

- Non-probabilistic methods (confidence based on business expertise).

  Sampling based on "business knowledge", acquired through experience, can be as valid as the traditional probabilistic approach when it comes to studying a specific phenomenon.

  Among these methods, we will mention only two of them:

  - **Subjective sampling:** This method is based on the appreciation of the profession, depending on whether an observation has a typical or atypical characteristic. It is then possible to focus on anomalies, to the detriment of representativeness.

  - **Convenience sampling:** This method, which retains in the sample the data provided by individuals who voluntarily present themselves, is generally not applicable in an industrial context and is more appropriate in societal studies.

Note that each of these methods is likely to induce a bias that will affect the model's performance and robustness, and this needs to be taken into account.

In addition to the choice of the sampling method (answering "how"), the size of the sample itself plays a fundamental role in representativeness (answering "to what extent"). The question of sampling effort is particularly addressed in ecology, a discipline in which it is necessary to find a cost/completeness trade-off when studying environments or communities. One of the most commonly used methods is the rarefaction curve [57], which encourages continued sampling effort until an asymptote of a number of species in an environment is reached. Other tools involving coverage rates complement this approach and are aimed to reduce bias. Similar approaches are also found in econometrics. Saturation curves, for instance, reflect the evolution of the distribution potential of a good or service in the market. This evolution also tends here towards an asymptotic plateau, unknown in advance.

The third and final step is to validate the quality of the sampling plan, i.e., to assess its capability to generate samples that will represent the population. This can be done by identifying the potential errors produced by the sampling plan (human errors, systematic errors, and random errors) and by ensuring that they remain below an acceptable threshold. This type of consideration includes the use of replicas. Thus, this step does not necessarily aim to validate a sampling plan, but to invalidate it when the errors are too high.

### 4.5.3.2 Representativeness and unstructured data

When studying structured data, the description of a population from a sample is based on the collection of information relevant to the population. Thus, achieving representativeness implies comparing distributions in a semantically relevant space. However, the characteristic of unstructured data, such as images or text, is not to be projected into this kind of space.

In the field of natural language processing, for instance, the similarity between the semantic networks of the samples and the semantic network of the original document [58] can be used to evaluate representativeness.

The minimal and sequential representation of reality by a visual recording device makes it unlikely or even theoretically impossible to achieve a complete representation of a moving scene.

But even in the case of unstructured data, certain use cases are more suitable for the constitution of supposedly representative samples. In the example of the early detection of breast cancer [59], the nearly 90,000 mammograms, obtained through a rigorous capture and calibration process, involve the same anatomical area. Thus, the variability of the images is essentially driven by age, genetics, body size, hormonal dynamics, eating habits breastfeeding activity by the patients, and, finally, by the possible presence of a cancer. The implementation of stratified sub-sampling and the definition of objectives targeted on these different criteria can lead to the reduction of some of the biases caused by the first sampling (possibly a convenience sampling).

Other fields of application, involving for example video capture and object identification/classification, make a similar approach futile, if not by severely restricting the operational design domain.

An alternative approach consists in analyzing the input domains with respect to new inputs, or a sub-sample to reassure on the conformity of these with the training dataset. Conversely, any deviation from the input domain will warn of a risk of non-representativeness of the training dataset. Thus, in image analysis, it is not possible to guarantee a priori the representativeness of a sample, but it is possible to detect some cases of non-representativeness.

### 4.5.3.3 Fairness

Machine Learning methods exploit the information conveyed by the learning dataset, by looking at the different relationships between the observations, which includes their correlations. In particular, they inherit the biases present in these data.

Some of these biases enable to increase the overall accuracy of a model when they are related to properties that are relevant to the targeted problem, and can be generalised to other observations. On the contrary, some biases are very specific to the data which is used to build the model and are related to a mere correlation (e.g. the well-known snow background on the wolves and husky classification [60]). Learning such correlations could hamper the prediction of the algorithm outside of the training sample. As a consequence, the control of possible underlying biases is a crucial issue when trying to generalise machine learning based algorithms and certify their behaviour.

The source of biases are multiple. In industrial cases, collected data may first not be representative of the operational design domain. Another common issue comes from unbalanced training samples or errors in the acquisition and/or labelling of the data leading to misclassified examples.

In all cases, removing unwanted correlation will help recovering the standardised behaviour of the algorithm and increase its accuracy. Research on biases in machine learning has grown exponentially over the very recent year. It includes the detection and identification of biases as well as the design of new algorithms to bypass unwanted correlations in the dataset, either by removing them from the learning sample or by controlling the learning step by adding a constraint, which enforces the so-called fairness of the algorithm. Such methods are also closely related to domain adaptation or transfer learning algorithms.

#### 4.5.4 Discussion

We have presented a list of criteria pertaining to Data Quality and also pointed out that their method of evaluation is prone to the subjectivity of the research or implementation teams, and depends on the targeted application and the operational design domain. The trust that will be placed in the dataset strongly depends on them.

It is accepted that the achievement, even if only theoretically, of representativeness for a sample used as a basis for learning, validating and testing a Machine Learning model is at the very least uncertain or in some cases even impossible (e.g. image classification in an open environment) since such a principle is not particularly well defined. Representativeness being hard to assess, is it required for all the datasets used in the development of a ML system (learning, validation, test)? An approach, which is still very controversial within the working group, could be to impose the representativeness constraint only on the test dataset, used to evaluate performances for the demonstration of compliance. But alleviating this constraint on the training dataset is risky in several respects: it favours a late discovery of weak learning datasets possibly increasing the overall development costs; it increases the overall risk by removing one level of verification.

By moving from statistical theory to the practice of Machine Learning in a critical context, it becomes relevant to rephrase the question "how to assess the data quality or data representativeness of a sample?" to "**Are both Data Quality and Representativeness essential for the implementation of a Machine Learning model in a critical system?**"

Considering that that a model will not be trained on all the observations that may appear in the population/domain of use, or on a dataset that bring a total confidence about its quality, how to limit the impact of this gap on final safety? One solution consists to mitigate the risks introduced by the ML systems by appropriate monitoring means, redundancies, error masking and recovery mechanisms at system level. These solutions are discussed in Section 4.3, where the counterpart of difficulty to assess representativeness, lies in Out-Of-Distribution detection.

The Data Quality and Data Representativeness High Level Properties are linked to the concepts of Data Fairness and Auditability. Data Quality property also contributes to the definition of Accuracy and Numerical Accuracy, while Data Representativeness also contributes to Robustness, Fairness.

### 4.5.5 Challenges for data quality and representativeness

For the Data Quality & Representativeness theme, we identify the following challenges:

- Challenge #4.1: How can we prove that a dataset is representative of the Operational Design Domain for a given usage?
- Challenge #4.2: How can the representativeness of a dataset be quantified, in particular in the case of unstructured data such as texts or images?
- Challenge #4.3: Which of the training, validation and test datasets should possess the data quality and representativeness properties?

## 4.6 MAIN CHALLENGE #5: EXPLAINABILITY

### 4.6.1 Introduction

The production of explanations is a central problem for the analysis of reasoning, communication, mental model construction, and human-machine interactions. It has been widely investigated in various scientific fields including philosophy, linguistic, cognitive psychology, and Artificial Intelligence [61].

The automated production of explanations has been much more considered within the field of Artificial Intelligence through three principal domains [62]:

- Expert System (in the 90's).
- Machine Learning (since 2000).
- Robotics.

In this chapter, we address the concepts of explanation and explainability in AI with a focus on Machine Learning.

### 4.6.2 On the concept of explanation

#### 4.6.2.1 Explanation

Many definitions have been proposed [63]–[66]. A reasonable one can be found in Wikipédia (https://en.wikipedia.org/wiki/Explanation, 2019):

> *An explanation is a set of statements usually constructed to describe a set of facts which clarifies the causes, context, and consequences of those facts. This description of the facts may establish rules or laws, and may clarify the existing rules or laws in relation to any objects, or phenomena examined. The components of an explanation can be implicit, and interwoven with one another.*
>
> *An explanation is often underpinned by an understanding or norm that can be represented by different media such as music, text, and graphics. Thus, an explanation is subjected to interpretation, and discussion.*
>
> *In scientific research, explanation is one of several purposes for empirical research. Explanation is a way to uncover new knowledge, and to report relationships among different aspects of studied phenomena. Explanation attempts to answer the "why" and "how" questions. Explanations have varied explanatory power. The formal hypothesis is the theoretical tool used to verify explanation in empirical research.*

This definition emphasizes the three key aspects of an explanation:

- Description of an existing phenomenon/system: an explanation provides information about a phenomenon (or a system).
- An explanation is dependent on a domain knowledge and its representation.
- Explanation helps to complete the domain knowledge by uncovering new knowledge (concepts, relationship between concepts) required for validating some hypothesis not confirmed by the initial domain statement.

To summarize, explanations enable a human to complete its mental models for understanding a phenomenon/system. "Understanding" means that the human has the capability to do inferences (deduction, induction, abduction…) on the behaviour of the system without relying on experiments. It is worth noting that this definition leaves apart communication aspects. Indeed, the producer and the recipient of the explanation can be the same person.

An explanation expresses a causal chain linking events for making clear a phenomenon [67]. "Making clear" is an elusive term which hides the tradeoff between *interpretability* and *completeness* of an explanation [63]:

- **Interpretability** relates to the capability of an element representation (an object, a relation, a property...) to be associated with the mental model of a human being. It is a basic requirement for an explanation. For instance, watching a series of pictures of a falling ball (the representation) can be interpreted as the usual trajectory of a falling body (the mental model).
- **Completeness,** in the context of explainability, relates to the capability to describe a phenomenon in such a way that this description can be used to reach a given goal.
  This concept encompasses two main notions:
    - **Precision,** which indicates how much details must be provided to the human to let her/him execute mentally the inference in a right way with respect to her/his goal. For instance, there is no need to know the laws of general relativity or quantum mechanics to predict the trajectory of a ball.
    - **Scope** (or context) which defines the conditions in which the explanations are valid with the given precision.

### 4.6.2.2 Explanations and AI

In the AI domain, explanation generation has been considered for enabling the use of complex systems by their end users. The field of Expert systems has been one of the most active in this area.

The logical process of Expert Systems is expressed through deductive inference rules (a ⇒ b) which can be directly accessed, interpreted, and corrected by the domain expert using the system. This capability, which enables a quick transfer of knowledge from an expert to a machine without the intervention of an IT specialist, certainly justifies the strong interest for those systems in the 90's.

Unfortunately, the number of rules and the opacity of their selection in the inference process have shown that the expressiveness power of a language is not sufficient for bridging the gap between the system processing and the user mental model. A synthesis of the numerous deductive inferences performed for a particular decision must be achieved for allowing the user to understand the rationale of the decision with his/her own cognitive limitations (which are not often well known).

Several works [68] have shown that the synthesis cannot be just a reformulation of the triggered rules. Explanation generation must be structured through a dialog with the end user. Questions such as "what?", "why?", "how?" are the means for the user to "navigate" in the space of the synthetic information given by the system. This led to a second generation of expert systems whose joint purposes were to do inference that an expert could do, and to explain the decision that an expert would do by considering the unknowns of the user [69].

### 4.6.2.3 Explanations and Machine Learning

There is a wide consensus in the ML scientific and industrial community on the need to have the capability to explain the behaviour of a model produced by these technologies.

The need to get explanations was justified *a priori* by many authors for:

- Ensuring trust.
- Establishing causal relationships between the input and the output of the model.
- Catching the boundaries of the model.
- Supporting its use by providing information about how the decisions have been taken.[45]
- Highlighting undesirable biases.
- Establishing how confident a model is with its own decisions.
- Allowing the end users to identify faults/errors in the model.
- Checking that the model does not violate some privacy rule.

These objectives can be associated to three types of person profiles [65], [71]:

- **Regular users**, i.e., the users of the ML based system.
- **Model developers**, i.e., the data scientists in charge of developing the ML model and the domain expert who provides information about the domain.
- **External entities**, i.e., regulatory agencies in charge of certifying the compliance of the models with required properties mainly related to safety and transparency.

All these profiles are not expecting the same kind of explanations. Regular users need to get highly interpretable information but few details about the way the system has processed the information during the inference. Model developers need precise information about the relationship between the input and the output of the model in order to validate it with precision; completeness in this case is the key driver for explanations. The case of "external entities" (scope of this chapter) will be discussed in more details later (§4.6.3), but it is worth noting that, in some circumstances, the regulatory authority may also request demonstration of explainability for the regular users. Beforehand, let us consider the inherent difficulty of getting explanations from ML models.

ML models are not fully built by human. They are partially produced automatically from a set of observations (training set) meant to be representative of the objects/concepts to learn. The global strategy for building a ML model consists in selecting or building the features able to discriminate the observations with respect to their associated object/concept. These selections rely on various heuristics used to avoid an exhaustive exploration of combinations of features that would not be tractable even for medium size problem. The observations used in the training set are the representation of the world chosen by the developer of the ML model and/or the system engineer.

This process makes ML systems very different from classical software:

---

[45] This is for instance motivated by the compliance with GDPR which mandates a right for the end users to obtain such information [70, Secs 15 and 22].

- The algorithm to build the ML model considers the training observations (a subset of the domain elements) as a whole for capturing the rationale of the decision model.
- The algorithm to build the ML model is an *ad hoc* composition of mathematical algorithms and heuristics applied to the domain features brought by the training observations. Those algorithms and heuristics can be understood somehow by human beings (they have been established through the design of the ML algorithm), but their composition is not designed through a human analysis of the domain.

The ML model is thus a computed artefact, associated to a specific domain which is not directly interpretable by a human. The various processing performed during the training and their representations must be interpretable to be part of the explanations. In other words, they must be associated explicitly with some domain concepts belonging to the mental model enabling a human to do some mental inference (*cf.* §4.6.2). So, while Expert Systems focus on communicating explanations in a human friendly way, ML systems require to step back: the end user must *first* be able to interpret the relationship between inputs and outputs to understand it (for instance, to validate it).

This need is an important focal point for the academic and industrial ML communities. The kind of questions that must be answered to is basically:

- Why has this observation (not) been associated to this class?
- What is the role of a given input variables in the decision made by the model?

Such questions can be useful for the ML designer and the end user. For instance, they may be used by the ML developer to determine the parts of the system to be fixed or they may be used by the end-user to understand the rationale of the ML decision/output and use it in a more appropriate manner or, at least, feel more comfortable for using it. The ML designer could also ask for more details about the relationship between the input and the output variables. In that case, he/she may require information about the internal representation of the domain produced by the ML algorithm (the latent space), which holds the features created/selected/aggregated by the algorithm. For now, it is not clear if an end user must also access to this space (white box approach) or if explanations based on the input and the output variables are enough (black box approach).

Many approaches are currently being investigated for producing appropriate explanations [63], [65], [71], [72]. As of today, and as pointed out by many authors, the means for producing accurate explanation, i.e. based on the real decision of the model, are not fully reliable or not even feasible [64], [73].

### 4.6.3 Explanations for certification of ML systems

In the case of the certification of ML based systems, two general cases may have some requirements specific to the production of explanation:

- Compliance with safety requirements.
- Compliance with legal requirements.

#### 4.6.3.1 Compliance with safety requirements

For certification, explanations can have two purposes:

First, explanations can *give confidence to the authority*. Indeed, explanations are often enablers of confidence. If the behaviour of a ML model were explained with much precision and with clear and correct indications about its relationship to the target domain, it would provide the authority with the ability to assess the system with a proper understanding of the design details and motivations. Unfortunately, as aforementioned, there is no guarantee such explanations could be produced. Furthermore, concrete methods producing evidence that an explanation about a ML model is correct are still lacking. Correctness of explanations may be achieved by recreating a causal model or through the identification of confidence metrics associated to an explanation…

Second, explanations can *help the end-user operating the system*. In some situations, the end-user needs to understand the behaviour or the decisions made by a system, in order to control it in an effective manner. This can be referred as the controllability of the system. The need for explainability for controllability is discussed in more details later.

The *type of explanations* depends on the impact of a failure of the system on safety, and on the level of autonomy. Hereafter, we distinguish three main levels of autonomy:

- The system is fully autonomous
- The system is autonomous but it can leave the control to an operator.
- The system is monitored, and the control can be taken by an operator.

*Full autonomy*

A system is fully autonomous when the expected function can be achieved by the system without any intervention by a human. Transport systems are willing to reach that level, but the technology is still considered immature for that purpose.

The deployment of such systems will only be allowed if strong guarantees about their behaviour are provided for a very wide range of circumstances. Should a catastrophic failure occur, a post-mortem analysis aiming at identifying clearly the root cause of the failure and the way to fix the system will be performed. This analysis may be supported by explanations about the behaviour of the model in the circumstance of the failure.

Additionally, the specification of the fix will also require a thorough analysis of this abnormal behaviour to provide information to the data science expert (for the algorithmic aspects) and/or to the domain expert (if a causal interpretation of the model can be given). Even if this type of "debugging" does not fall within the scope of this White Paper, it can be assumed that evidences of the efficiency of the fix will have to be given. In this case, the usefulness of the explanations will depend on their ability to point out the root cause of the problem (as for any system). But more importantly, it will also depend on their ability to establish the scope of the failure which is not easily accessible since induced internally by the ML algorithms.

*Autonomy with the capability to transfer the control to a human operator*

In order to be able to transfer the control from a ML based system to a human, the system will require additional abilities to monitor the ML behaviour, which is also an active research field [74]–[76]. The idea is to add to the model some self-assessment capabilities that may detect situations:

- Which are far from the ones considered for training the model (novelty, or out-of-distribution detection).
- For which the system has not enough discrimination means [77] and cannot rely on its own confidence.

If such situations are detected, the system might trigger some mitigation operations such as transferring the control to a Fail Safe module [78] or to a human operator (such as the car driver, the pilot of the aeroplane, a remote-control center, …) as recommended by some standards (for instance, the SAE's level 3 expect a human may intervene per request of the autonomous car [79]).

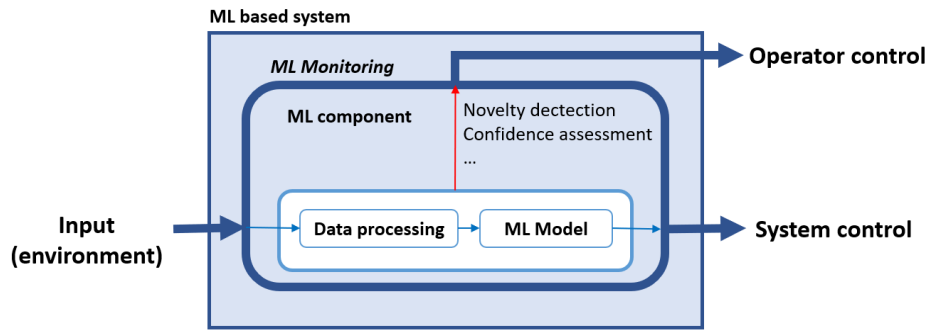

*Figure 16. The system integration of a ML model requires a monitoring able to deal with contexts of execution not aligned with the model scope.*

The effectiveness of such approach relies on the capability of the operator to take-over the control in the most appropriate way [80]. It means that he/she must, in a very limited timeframe determined by the context of use of the application:

- Be aware of the situation .
- Understand the reason he/she must take-over the control of the system.
- The nature of the problem which has not been solved by the system.
- The set of actions he/she can perform.

In such circumstances, controllability of a ML based system does not depend on the precise understanding of the relation between the input and output of the ML model. It depends more on the understanding of the function of the system (*i.e.* the logical relationship between the categories of input and output), its limitations, and the means that are available for continuing operation of the system.

It is not necessary to understand the internal logic of the model [81]. It may even be counterproductive since the situation may require a fast reaction. Understanding the rationale of the fail-safe strategy is more appropriate for the operator in determining the intent of the control transfer and the actions he/she must execute.

For now, this level of autonomy is implemented without any ML techniques. The generation of explanation in this context can be thus performed through explanation *planning*, which exploits models of the dynamic behaviours which occur in the domain [82]. The assessment of the efficiency of the explanations could be done with classical human-in-the-loop verification methods.

*Supervised autonomy*

Supervised autonomy is an approach followed by some industrials for tackling safety problems. It is close to the one described previously but, here, it is expected that the operator monitors the decisions of the system and bypass them in case of disagreement. The best-known example is Tesla which recommends to the driver to keep their hands on the driving wheel even when the autopilot is activated [83].

Other examples are ML models used in medical diagnosis for establishing the cause of symptoms. The system may establish a complete diagnosis but this one is reviewed by a doctor before a final decision. The notion of autonomy here can be debated in the sense that the system decision remains under the control of an expert. However, the ML model can achieve an accurate diagnosis (as a human expert might do with more effort) without requiring the intervention of the expert during the processing of the data. In this situation, explanations are definitively required as in the case of expert systems (cf. § 4.6.2.2): to take a decision, the expert must understand the reason why the ML model has given a certain diagnosis among all the possible ones.

In the case of autonomous vehicle, reaction time is much more critical than the completeness of the explanation. The first objective is to ensure situation and risk awareness for managing takeover phases in very limited timeframe [80]. It means the operator must understand instantaneously what is the decision taken by the system and what are the element of its environment involved in this decision. In other word, the interface between the user and the model must integrate as far as possible the operator's perception of the environment and the input/output managed by the system. For instance, "augmented reality" tools may be used to display the identified threats and the vehicle trajectory so as to limit the driver's cognitive load. From the explanations perspective, the challenge consists in identifying precisely and rapidly which input (or element of the environment) is responsible of the output (or the decision of the system). The causal interpretation of the processing achieved by the system is of less importance.

As discussed previously (*cf.* §4.6.2.3), it is very hard to provide an interpretable and complete explanation of a ML model. A precise and contextualized causal chain would be needed but, unfortunately, ML models are closer to Bayesian models than to causal models...

A causal interpretation of the model could be then an answer to the need of explanations but, then again, this interpretation is hard to obtain for any given ML model [84] and existing approaches and tools are still exploratory. Besides, as stated by [73], if such causal models were to be obtained with a high level of completeness, they would well replace models such as those produced by neural network.

### 4.6.3.2 Compliance with legal requirements

Explainability is not only required for the sake of safety; it is also required for ethical reasons. For instance, in an article one often called the "right to explanations" regarding automated data processing, the General Data Protection Regulation (GDPR, [70]), states:

> *The data subject shall have the right to obtain from the controller confirmation as to whether personal data concerning him or her are being processed, and, where that is the case, access to the personal data and the following information: […]*
>
> *(h) the existence of automated decision-making, including profiling, referred to in Article 22 (1) and (4) and, at least in those cases, meaningful information about the logic involved, as well as the significance and the envisaged consequences of such processing for the data subject.*

This highlights the social demand for an AI that people can understand and predict, but note that no specification is given about the explanation to give, in particular about their completeness [70, Sec. 5].

Similar requirements emerge in new AI ethics policies. For example, one of the seven AI trustworthiness criteria published by the AI High Level Expert Group of the European Commission is "Transparency" and includes explainability requirements.

It is too early to know what level of completeness is expected by these regulations and policies. In the domain of safety-critical systems, we may assume that explainability requirements will be mainly driven by safety requirements that probably exceed transparency requirements.

### 4.6.4 Conclusion

Providing explanations about ML models is a key feature for a better mastering of the systems using them. When these systems become critical, focus must be set on the explanations allowing a human operator to supervise the decisions taken by the system.

When these systems are subject to a certification process, challenges concern

- The *interpretability* of the information communicated, i.e. "is the operator able to associate them with his/she own domain knowledge.
- The *completeness* of the information communicated, i.e., "how much details must be provided to the operator with respect to the actions he/she must execute in a given timeframe?"

Additional work is certainly needed, both on the refinement of the needs of explanation for certification and on the tools to explain ML. The involvement of ML, certification, and human factor experts is crucial to achieve the right level of explainability.

### 4.6.5 Challenges for explainability

For the Explainability theme, we identify the following challenges:

- Challenge #5.1: How to ensure the interpretability of the information communicated to the operator? How to ensure that the operator is able to associate them with his/her own domain knowledge / mental model?
- Challenge #5.2: How to provide the necessary and sufficient information to the operator regarding the actions he/she must execute in a given situation/timeframe?
- Challenge #5.3: To what extent explainability is needed for certification?
- Challenge #5.4: When is an explanation is acceptable? How to define explainability metrics to assess the level of confidence in an explanation?
- Challenge #5.5: How to perform investigation on ML-based systems? How to explain what happened after an incident/accident?

## 4.7 MAIN CHALLENGE #6: ROBUSTNESS

### 4.7.1 Introduction

#### *4.7.1.1 The need for robustness*

Robustness is one of the major stakes of certification. The state of the art regarding the safety-critical development is that the software should be robust with respect to the software requirements such that it can respond correctly to abnormal inputs and conditions. For instance, DO-178C/ED-12C defines the robustness as *the extent to which software can continue to operate correctly despite abnormal inputs and conditions.* It requires several activities to meet the objectives:

- **Specify any robustness behaviour** of the software using derived requirements (specification of any emerging behaviour of the software which is not traceable to input requirements).
- **Review design and code** to verify that the robustness mechanisms are in place.
- **Develop test cases** using specific software features (e.g. equivalence classes, invalid values, input data failure modes, overflow of time-related functions…) to test the robustness of the embedded software.

Robustness can be seen as the system behaviour analysis in regard to any – known and unknown – environmental or operating perturbations. Considering the specific case of ML development, perturbations should be considered of a data-driven development process. The model is developed to fit the intended function by the use of appropriate datasets, learning process, and implementation process for target embodiment. While a human has the capability to adapt his/her behaviour (with respect to the intended function) whatever environmental or/and operating perturbations, the ML system has only memorized what it has captured from features during the learning phase. Therefore, the robustness of a ML system should logically relate to the robustness of the ML inference versus any variability of the input data compared to the data used during the learning process.

Perturbations can be natural (e.g. sensor noise, bias…), variations due to failures (e.g. invalid data from degraded sensors), or maliciously inserted (e.g. pixels modified in an image) to fool the model predictions. Perturbations can also be simply defined as true data locally different from the original data used for the model training and that might lead to a wrong prediction and an incorrect behaviour of the system. The model inference should be robust to all of these perturbations that were not considered during the training phase, i.e. the model should develop the ability to "correctly" generalize from the features that were encountered during training.

If these perturbations could be classified as "known" and therefore deemed more accessible to modelling, the trickier aspect comes from the "unknown" part. As far as the perturbations can be defined, we remain in the paradigm of the current certification approach where robustness is apprehended with respect to the requirements. With a part of unspecified behaviour (or behaviour untraceable to the upstream systems requirements) contained in the training dataset, data-driven development makes ML very attractive to model functions that cannot be humanly definable (e.g. pedestrian detection, physical phenomenon not properly mastered…) but makes robustness a difficult property to be possessed by safety-related system.

The robustness may also be tackled from the model design perspective. Indeed, the behaviour of the ML system may be sensitive to the perturbations of the model design and implementation characteristics. For instance, perturbations of hyper-parameters, model architecture, or even slight changes of the computed weights when the model is embedded on target can adversely change the ML system behaviour at inference time.

Possessing the robustness property remains a challenge in the frame of future ML system certification. However, such demonstration will have to stay commensurate to the encountered risks. Indeed, robustness approaches will have to be considered so that ML systems can be adopted for advisory functions. For more critical functions implementation, it will be the impact of lack of robustness on the system reliability that should be quantified (or bounded) in order to be acceptable versus the safety of the system in all foreseeable operating conditions (cf. §4.3).

### 4.7.1.2 General definition of robustness

Analysing the behaviour of a system under normal (i.e. expected) functional conditions is a necessary yet insufficient condition to qualify with confidence the functioning of the system under all possible operational scenarios. Appropriate tools and methodologies must be designed to analyse the behaviour of the system in abnormal conditions. One such tool is robustness analysis which aims to quantify the influence of small perturbations on the functioning of the system. Obviously, the results of such analysis strictly depend on the definitions of "small" and "perturbation".

"Small" requires the specification of metrics, e.g. estimating the similarity between normal and abnormal (yet possible) input values. It is generally acknowledged that designing a "good" similarity metric, especially for high dimensional domains, is very difficult, if possible at all. Thus, metrics are choices of the analyst and not intrinsic properties of the world (although they should reflect real phenomena).

"Perturbation" refers to alterations of numerical quantities[46] that may induce erroneous values. In the context of ML models, perturbations mainly affects the inputs. In addition, the implementation of the model can also be subject to numerical instabilities.

Generally, we distinguish between local and global robustness:

- **Local robustness** is concerned with the response of the system w.r.t. deviations (i.e. within a small "neighbourhood" of) from a given input value. A model is deemed robust if its responses to these deviations are similar or identical to the response provided by the original (i.e. unperturbed) input.
- **Global robustness** takes into account the combination of multiple "input" deviations with regards to the behaviour of the system. By "input" we refer to any type of factor that contributes to the behaviour of the system, be it input data or any design choice or internal (hyper-)parameter of the system.

Robustness can be tested against known or unknown conditions. When conditions are known (e.g. admissible values can be specified), particular efforts may be invested to ensure that the system will behave as intended in the presence of those abnormal conditions. When the dimension of the input space is too large to be specified (e.g. pedestrian detection), it is impossible to define exactly what an abnormal condition can be. In such conditions, robustness has to be evaluated against unknown conditions.

---

[46] for image or signal processing, but the perturbation can be of another nature for other data types

Robustness of a system (or model) may be estimated by two means. One relies on empirical procedures to estimate the effects of a given perturbation on the behaviour of the system. The other one offers provable guarantees about these effects.

Provable guarantees definitely are more desirable but are generally harder to obtain. In addition, depending on the hypotheses on which they rely upon, sometimes these guarantees may prove impractical (e.g. by providing bounds that are too conservative or too restrictive).

From the above considerations, we should note that, ultimately, all robustness analysis tools rely on some human decisions. Indeed, even if the perturbations are generated randomly, they are drawn from a domain (metrics, magnitude, structure) that is human defined, and that represents an approximation of the actual, real-world, phenomena. So, one has to be careful when interpreting the results given by these tools.

Hereafter, we focus on two type of robustness properties:

1. Ability of the system to perform its intended function in the presence of:

    a) Abnormal inputs (e.g. sensor failure).

    b) Unknown inputs (e.g. unspecified conditions).

2. Ability to ensure coherency in the sense that for similar inputs, it provides an equivalent response. e.g. "a network is delta-locally-robust at input point x (similar) if for every x' such that ||x-x'|| < delta, the network assigns the same label/output to x and x' (equivalent response)".

### 4.7.1.3 Link with other HLPs and challenges

The scope of robustness is very large and has many intersections with other HLPs and challenges that are mentioned in this section.

Firstly, as described in Section 4.7.1.1, robustness is closely linked to Specifiability and in particular the ability to specify the input requirements (§4.4).

In traditional software engineering, robustness is analysed with regard to the functional specifications. However, ML-based modeling poses serious problems with regard to the feasibility of formulating complete and unambiguous functional specifications. Specifying types of perturbations, characteristics of the model's inputs or outputs that are required by the intended functionality is far from trivial or often even impractical/impossible.

It has been seen in main challenge "Probabilistic Assessment" (§4.2), that performances commonly used in ML (e.g. accuracy) are often based on average evaluation over a dataset, which is definitely not a sufficient criterion to decide whether the system relying on is safe or reliable. Adversarial examples that will be described in next sections have shown that accuracy is not enough. Metrics to evaluate robustness must be considered and evaluated in order to assess the behaviour of a model (and ultimately a system relying on it).

The general aim of ML modeling is to build models that generalize well, i.e. learn from seen data and make good predictions on unseen data. Ensuring that a model is robust to certain perturbations helps against model overfitting.

The confidence in the predictions of a model is directly related to its capacity of handling uncertainties, e.g. unknown inputs. The response of the model to subsets of such unknown inputs (namely small deviations from known ones) is the central theme of the robustness analysis. In this sense, robustness analysis can be seen as a valuable tool pertaining to a larger toolkit required by any sound probabilistic assessment of the behaviour of a model (§4.2).

Robustness implicitly posits a question related to the Data Representativeness too (§4.5): do perturbation models are representative with respect to the possible operational conditions of the system? In other words, are they realistic? Do they address well enough edge cases of the intended functionality?

Besides, since lack of robustness can lead to failure, it is also linked to Resilience (§4.3). The latter will focus on capacity to enable system mitigations means, whereas this section will focus on algorithmic mitigation for robustness.

The property of robustness is related to the property of Provability (§4.8.1) whenever it can offer provable guarantees concerning the behaviour of the model, without the need to deploy the system in the target environment. On the other hand, given the limited scope (e.g. incomplete guarantees) of any robustness analysis methodology, it is often recommended that monitoring procedures are put in place to supervise the correct functioning of the system under real operational conditions, as evoked above.

To end with, security issues (cyber-attacks), or learning phase attacks, may also be a concern for robustness, but are out of scope of this White Paper.

### 4.7.1.4 From robustness to antifragility

A system (or model) is deemed to be robust if it can tolerate abnormal values. While this is a desirable property, it may not be the most desirable. According to [85] the response of a system with respect to perturbations can be characterized in the following ways: "A fragility involves loss and penalisation from disorder. Robustness is **enduring** to stress with no harm nor gain. Resilience involves **adapting** to stress and staying the same. And antifragility involves gain and benefit from disorder." (Figure 17).

The notion of antifragility has been introduced by Taleb [86] and is defined as "a convex response to a stressor or source of harm (for some range of variation), leading to a positive sensitivity to increase in volatility (or variability, stress, dispersion of outcomes, or uncertainty, what is grouped under the designation 'disorder cluster')". The concept draws parallels with "hormesis", when the stressor is a poisonous substance and cells or organisms affected by small doses of it become better overall (e.g. improved survival probability).

While many efforts have been made in the last decades to endow artificial systems with bio-mimetic capacities, current state of affairs acknowledge a vast distinction between organic and artificial intelligence. We are not aware of any current AI models that can be considered antifragile in the sense stated above.

There are nevertheless certain claims that some software design methodologies follow antifragile design (see e.g. [85]). Some may claim that stochastic learning (as used e.g. in deep learning) constitutes an example of antifragile design. We regard these claims as (at least for the time being) either debatable or incomplete.

While in this paper, we focus our discussions to the concepts of robustness and resilience for offline learnt systems, we fully agree that we should strive to design more antifragile systems in the long run.

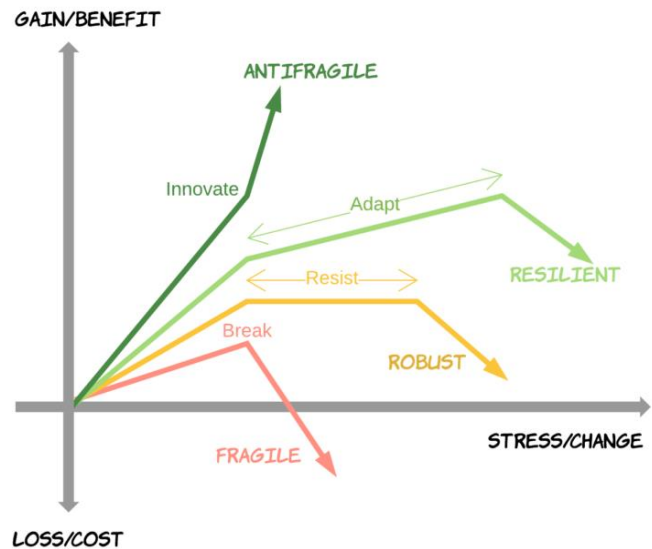

*Figure 17. Possible behaviour of a system in response to stress (from [2])*

#### 4.7.2 Scope of this main challenge

From the above considerations, we can conclude that robustness should not be seen as a standalone challenge, neither should it be considered as a sufficient property to ensure the well-functioning of a system. Besides, the problems raised by robustness are usually NP-hard problems, and in practice, the design of any system is a trade-off between complexity, robustness, scalability, developmental and operational costs, etc.

The main scope of this section will be about:

- Metrics for robustness assessment, and similarity measure.
- Algorithmic means for enhancing/guarantee robustness.
- Algorithmic means for detection abnormal or adversarial inputs.
- Adversarial attacks and defence.

#### 4.7.3 State of the Art

##### 4.7.3.1 Abnormal inputs and unknown inputs

Detecting abnormal or out-of-distribution (OOD) inputs is done classically on critical systems **where specification of the abnormal inputs and conditions can be given** (for instance temperature or speed sensors failure). But when dealing with high dimensional data (such as images, video or speech), and tasks where a part of unspecified behaviour exists (e.g. pedestrian detection), defining and dealing with abnormal or unknown inputs becomes very challenging (§4.4).

Furthermore, the whole machinery of Machine Learning technics relies on the assumption that the learning sample conveys all the information. Hence, the algorithm is only able to learn what is observed in the data used to train the algorithm. So, no novelty can be forecast using standard methods, while in practice new behaviour may appear in the observed data. The incapacity of Machine Learning algorithms to process experiences outside the training domain is an important issue.

In this domain, several topics are studied, first the detection of unknown/unknown inputs (not inside the train and test dataset distributions) in order to reject their processing or raise an alarm. Another research topic focus on the ability to learn these unknown examples (one-shot learning, lifelong learning…) to be able to adapt to new ones without a long and expensive retraining but is out of scope of this White paper which is restricted to offline learning.

In all cases, the system should handle these unknown/unknown inputs to ensure that outputs are based on some knowledge acquired during the learning process. The ML subsystem should be able to detect these unknown observations.

In recent years, likelihood–based generative models have been proposed in order to capture the distribution of the training data. Three main categories are described in [87]: Autoregressive models [88], variational autoencoders [89], and flow based models [87], [90], [91]. The latter have been developed to explicitly estimate the density function of the dataset, and could be seen as good tools to estimate experiences outside of the distribution. However, recently [92], it has been shown that, even on some classical dataset, some flow based-model could estimate that an experience outside of the distribution has a high probability to belongs to the training distribution.

Recent papers tackle the problem of detecting Out of distribution sample for Deep Neural Network. Liang et al. [93] propose a first out-of-distribution detector, called ODIN, applied on pre-learnt models, using temperature scaling of the output softmax layer, and input preprocessing in order to separate the output score of in and out-of-distribution. Lee et al. [94] propose a confidence score based on an induced generative classifier under Gaussian discriminant analysis (GDA), using the Mahalanobis distance between test sample x and the closest class-conditional Gaussian distribution. Same authors [95] propose also a framework to train a DNN with a confidence loss to map the samples from in- and out-of-distributions into the output space separately, using a generator of out-of-distribution inspired by generative adversarial network.

The authors of [96] propose a probabilistic definition of robustness, which requires the model to be robust with at least (1−epsilon) probability with respect to the input distribution.

### 4.7.3.2 Local robustness

Literature on ML systems attacks and defences is huge, and many surveys gather references [97], [98]. Three main categories of attacks are listed, depending on when and why such an attack may take place [98]:

- **Poisoning attacks** trying to modify the learning data to influence the ML outcome.
- **Exploratory attacks** trying to gather as much information as possible on a learnt system to understand or copy the system.
- **Evasion attacks** trying to construct malicious inputs changing the decision of the learnt ML system.

Poisoning and exploratory attacks, which respectively concern security and industrial property protection, are not covered in this White Paper (§1.3). So, we focus on evasions attacks.

In 2014, Szegedy et al [99] pointed out the lack of robustness of deep nets proposing a first method to find adversarial samples. For a learnt ML model $f$, an adversarial input is defined by:

> Given an input $x \in \mathbb{R}^m$, and given a target label $l \neq f(x)$, find $r \in \mathbb{R}^m$, minimizing $\|r\|$, such that $f(x + r) = l$

Szegedy et al [99] have shown that for images, adversarial examples can be found adding noise that is invisible for the human.

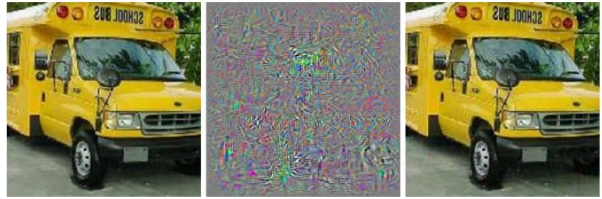

*Figure 18. (left) original image, (center) adversarial noise, (right adversarial example classified as an ostrich*

The equation is linked to the property of local robustness that critical systems must ensure where one may want to ensure that around a sample x, the output of the ML system should remain the same:

$$\forall\, r \in \mathbb{R}^m, \text{ such that } \|r\| < \epsilon, f(x + r) = f(x)$$

Many methods have been proposed in literature to search for adversarial examples, depending on adversarial capabilities and goals.

Adversarial capabilities depend on the knowledge the adversarial method may know and learn about the ML system:

- **White-Box attacks**: adversarial attack system knows ML type (NN, SVM,…) and has access to the internal states (weights, gradients…), and/or to the training set.
- **Black-box attacks**: adversarial system has no knowledge on the internals of the ML system, but may have access to the training dataset distribution (non-adaptative), or may be able to use the ML system as an oracle (adaptative), or only knows ML output on a restricted dataset (strict).

Goals of the attack systems can also have different goals [97]:

- **Confidence reduction.**
- **Misclassification**: changing the decision for a given example to any class different from the original one.
- **Targeted misclassification**: changing the decision for a given example to a target class.

[97] describes the complexity of the attack with regard to the capability and goal of the attack system, in the Figure 19.

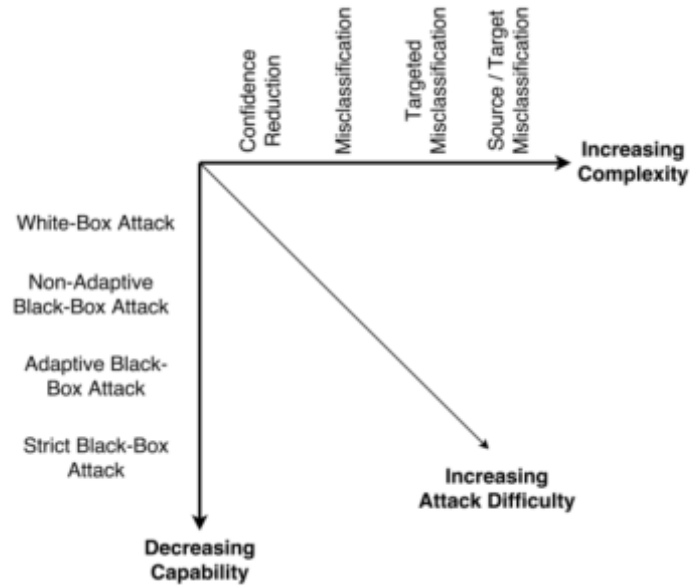

*Figure 19. Taxonomy of adversarial attacks and complexity*

With regard to the trustworthiness objective, robustness to any type of attack may be required, and so the easiest ones, i.e. white-box attacks for confidence reduction or misclassification. Besides, since most of white-box attacks rely on gradient computation, Black-box attacks have also been developed to find adversarial examples on ML methods (such as Decision Trees) where gradients cannot be computed (transferability of attacks).

The last axis for attacks taxonomy is the norm used for defining the locality of the adversarial example $\|r\|$. Classical norms used are:

- L2 Norm $\sum r_i^2$: L-BFGS [99], Deepfool-L2 [100], FGSM [101], JSMA [102], Carlini [103].
- L1 Norm $\sum |r_i|$ Deepfool-L1 [100], Carlini [103].
- $L\infty\ norm$ ($\max\{|r_i|\}$) Deepfool [100].
- L0 ($\text{count}\{r_i! = 0\}$): Single pixel [104].
- Affine or perceptual transforms: Adef [105] Spatial attack [106][107].

But, as indicated in [103], for image processing system, "No distance metric is a perfect measure of human perceptual similarity, and we pass no judgement on exactly which distance metric is optimal. We believe constructing and evaluating a good distance metric is an important research question".

Since 2016, an arms race has occurred between researchers working on defence strategies to make ML systems more robust and those working on adversarial attacks. Hundreds of papers are published every year on defence strategies, and a full survey is out of scope of this paper. A survey on attacks and defences can be found in [108]. [109] categorizes defence strategies in three main categories: modifying the training phase to include robustness to adversarial [110], [111], modifying the inference phase to defend [112], detecting adversarial attacks [113]. In this latter category, as mentioned in the previous paragraph, some methods tackle the out-of-distribution detection. By considering adversarial inputs as out-of-distribution examples [93], [114]–[116].

However, each published defence strategy has been broken within few weeks by new attacks (see for instance [117] breaking 6 defences). Up to now, no defence strategy is able to resist to any kind of attacks. It has also been shown in [118], [119] that there exist a trade-off between the standard accuracy of a model and its robustness to adversarial perturbations. So, neither applying defence strategies, nor failing to find an adversarial example will give a proof for local robustness.

Research for mathematical guarantees using special kind of ML (for instance Lipschitz networks, or monotonous network), or formal proof of robustness (§4.8.1) could give some evidences, or at least define classes of equivalence around a given example.

*4.7.3.3 Robustness with regard to annotation*

Arguably the most successful results in recent Machine Learning (deep learning) were obtained using what is called supervised learning, where a sample dataset of tuples (input, expected output) is provided for training and testing purposes. This learning strategy, which is also the focus of our analysis in this document, necessarily relies on human annotations. The quality of human annotations can thus be seen as a meta-parameter of the training methodology and one may legitimately ask whether the performance of the model is robust to this factor.

This is of utmost importance when applied to visual object recognition, where critical decisions may directly depend on the correct identification of the objects. The authors of [120] aim to answer this question concerning a well-established challenge in Computer Vision, i.e. object recognition on CIFAR-10 [121] and ImageNet [122] datasets. They very carefully follow the data collection methodology to build new datasets for validation. In spite of their particular attention, they discover significant performance drops for all state-of-the-art object recognition models on the new test datasets: 3-15% for CIFAR-10 and 11-14% for ImageNet in terms of accuracy. According to their analysis, the performance gap cannot be justified either by the adaptativity gap (hyper-parameter tuning) or generalization gap (induced by random sampling error). No visible signs of overfitting on original test sets could be identified. The authors conjecture that the "accuracy drop stem from small variations in the human annotation process. [...] Models do not generalize reliably even in a benign environment of a carefully controlled reproducibility experiment". While the conclusion seems disconcerting, it calls for a careful robustness analysis with respect to choices and parameters concerning the whole methodology used for training and validation of ML models.

### 4.7.4 Challenges for robustness

With regards to the need for robustness for Certification purpose (§4.7.1.1), the definition and scope of this paper (§4.7.2) and the short review of state of the art methods (§4.7.3), we propose a list of challenges organized into four objectives

- Challenge #6.1: How to assess robustness of ML systems? How to measure similarities in high dimension (image)?
  - Metrics for similarity: as explained before most of papers on local robustness are dealing with ball in Lq norms (q=1,2,∞). But in high dimension, and particularly for images, similarity between two inputs cannot be reduced to a distance in Lq. Recent papers are focusing on Optimal transport/ Wasserstein distance [123], or mutual information [124], but defining an acceptable similarity measure is still a challenge.

- - Metrics for robustness: comparing robustness of ML systems is also a challenge. Attacks and defence papers often report the number of adversarial example founds for a given distance to samples in a dataset, but this is not acceptable for critical systems. Metrics for comparing robustness of ML systems should be defined.
- Challenge #6.2: How to detect abnormal or adversarial inputs?
  - Out-of-distribution detection: As seen in §4.7.3.1, out-of-distribution detection or rejection in high dimension is still an open challenge [92] and have to be addressed for critical system implementation, to enable alarm raising when abnormal or out-of-distribution or even adversarial inputs are processed (e.g. [93], [114]).
- Challenge #6.3: What kind of ML algorithm can enhance or guarantee robustness?
  - Robustness by construction: Since the race between attacks and defence is endless, a challenge is to propose ML systems or loss that will provide local robustness by construction in specific cases (e.g., Lipschitz network [125], [126] or SafeAI loss [127] to enhance local robustness).
  - Guarantee of robustness: If global robustness in high dimension is a hard challenge, research can be done on special cases in lower dimension, or with special regularity for the intended function to prove (mathematical or formal proof) that the intended function is always respected.
  As a general note, we remind that robustness usually raises NP-hard problems. However, in certain contexts, knowledge about the application domain and the properties of the model can be exploited to develop practical solutions to the problems of robustness. In order for these methods to be useful for a certification endeavor, it is necessary that they are computationally efficient. This is a general challenge that is not proper to the robustness analysis but is common to many challenges listed in this document.
  - Robustness assessment with respect to (hyper-)parameters of a model or methodology.
  While the paper is mainly focused on the robustness of a model with respect to its inputs, this is not the only type of robustness one should consider in order to be confident about the behaviour of the model. Many applications require for example that the model's behaviour is robust to small perturbations of its parameters (e.g. the weights of a deep network) for example due to numerical truncations required by the target hardware on which the model is embedded. Others may require stability with regard to methodological parameters or hyper-parameters. Defining appropriate metrics in parameters space poses serious challenges.
- Challenge #6.4: How to identify and take into consideration corner cases?
  - Corner case detection: robustness can also be qualified with regard to known input domain where a ML system works correctly, and where it fails. Providing tools for searching corner case domains is also a challenge.
  - Identifying and adding corner case examples that are slightly out of domain can also be a way to enhance robustness.

## 4.8 MAIN CHALLENGE #7: VERIFIABILITY

In this chapter, we explore the concept of verifiability, i.e., *the ability of an artefact (an algorithm, a computer code) to be shown compliant with the requirements that it implements*. We set the focus on two specific properties[47] related to verifiability: *provability* and *testability*. And, we explore how these properties can contribute to the confidence about a ML algorithm.

### 4.8.1 Provability

Provability of a machine-learning algorithm is the extent to which a set of properties[48] on this algorithm can be guaranteed *mathematically*.[49] The larger this set, the higher the degree of provability. Guarantees can come from any field of mathematics: optimization, mathematical logic, functional analysis etc.

In this section, we will distinguish two different situations (see Figure 20):

- *A priori*-provability or *by-design* provability: The desired property is mathematically "transferable" as a design constraint to the ML algorithm. Then, to prove the property, it is necessary to demonstrate the validity of this transfer (i.e., if the design constraint is satisfied *then* the property holds on the model) and to demonstrate compliance with the design constraint.
- *A posteriori*-provability: The desired property is verified on the model after training. This approach may also rely on some assumptions on the ML algorithm (e.g. the architecture, the size of the network, the activation function type for a NN...), but these assumptions depends on the problem (e.g. [128]).

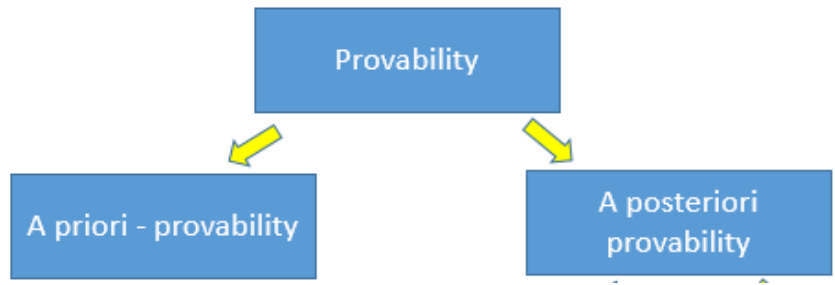

*Figure 20. Provability hierarchy*

#### 4.8.1.1 A priori-provability

"A priori provability" (or by-design) relies on appropriate design choices to ensure the capability to verify formally the model against some properties.

Gaussian processes [129] are well known models constrained by a priori knowledge or assumption about the problem at hand. For example, smoothness of the model can be enforced by choosing the co-variance matrix form properly.

Inspired by the work of Daniels, M. Velikova [130], F. Malgouyres et al. propose to ensure by design overestimating predictions in the case of a regression task of a monotonous function [3]. This approach is of particular interest when underestimating the quantity to predict leads to safety concerns.

---
[47] Other properties could also be considered, such as « demonstrability » or « inspectability » if we refer to the usual verification means.
[48] Or "requirements" in order to be consistent with the definition of "verifiability".
[49] In that sense, and despite its name, "provability" is not strictly related to deductive *proof*. It related to any verification method relying on a mathematical basis, including abstract interpretation, model-checking, etc.

Monotonic networks allow to introduce domain knowledge (monotonicity of the function to learn) and offer guarantees on outputs, or provide invertible function [131]. More information can be found in [131]–[133].

"Lipschitzness" is another design property providing guarantees about the network behaviour, especially in presence of perturbations. In [134], the authors explain that a small Lipschitz bound for a network implies some robustness to small perturbations (see also §4.7). Even if an upper bound of the Lipschitz constant of a Neural Network can be evaluated after training [134], several methods propose to enforce model's regularity during training, for instance using a regularization term [135], [136], or as part of the layers [137]. Lipschitz constant is a natural counter measure to adversarial examples (outputs can only change in proportion to a change in its inputs), but it is not the only benefit of this property (e.g. optimal transport) [138].

### 4.8.1.2 A posteriori-provability

A posteriori provability relies on the formal verification of the model. This approach is largely illustrated in the context of defences against adversarial attacks [139]. Recent works intend to prove the network robustness with respect to any kind of attack with a bounded perturbation amplitude.

*Abstract interpretation* may be used to prove the correct behaviour of a network in the neighbourhood of a particular input. [128] gives an example of instance-wise provability.[50] In this work, the input domain is represented by a so-called *abstract domain* (e.g., intervals, boxes, or zonotops) whereas the neural network behaviour is represented by a set of abstract transformations in the abstract domain. This technique is sound but incomplete in the sense that true properties may be unproved due to over-approximation of the input domain by the abstract domain. The authors use this approach to prove robustness to pixel brightening attacks.

Other techniques have the capability to prove property for any input, and not only in the neighbourhood of some input. Reluplex [13], for instance, uses a SMT solver[51] to prove properties or find counter examples. Unfortunately, Reluplex is still not scalable enough to address state of the art networks and some numerical limitations are also pointed out in [15]. So variants are proposed to overcome some of the limits of Reluplex [140], [141].

Concerning generalization capability, we can cite [142] where the authors derive a bound on the generalization capability of a network with respect to a given training set. This work is part of a long list of works [143], [144].

One of the great problems of a posteriori techniques is their lack of scalability to deep neural networks. In general, papers address small MLP or small CNNs. We can guess that technics actually used in practical applications will be among ones able to address complex (that is large and with complex architectures) networks.

---

[50] Instance-wise provability: Capacity to prove a property with respect to a particular input.
[51] SMT stands for "Satisfiability Modulo Theory", a category of decision problems for logical formulas with respect to combinations of background theories expressed in classical first-order logic with equality [Wikipedia].

### 4.8.2 Testability

Testability of a machine-learning algorithm is the extent to which a set of properties on this algorithm can be verified by testing. Testing, a method based on the execution of the system in specific *test scenarios*, is one of the major verification means in current certification practices.

Many testing methods have been proposed for systems implemented using non-ML techniques. Some of them are obviously applicable to ML-based systems, in particular when they concern the compliance of an implementation with its low-level requirements. This concerns for instance the verification of the implementation of a ReLU function, the verification of the absence of overflow on convolutions, the correctness of floating-point computations, etc.

The new difficulty with ML is to characterize the proper behaviour of the algorithms with respect to its high-level specifications, potentially huge input domain and with quite unknown decision logic executed by the algorithm. Today, it is not clear how to perform this verification and if all the algorithms (algorithm associated to intended behaviour specifications and input domain specifications) can be tested properly.

When a ML Machine Learning algorithm is tested, it is most of the time through some massive testing strategy. ML algorithm are made of one or several complex functions and they are employed in very complex environment. Functions and environments require specific testing strategies.

In addition to this work, an interesting survey can be found in [145].

#### 4.8.2.1 Testing the learnt function

The behaviour of ML algorithms, in particular DNNs, is strongly data dependent and is controlled by a huge set of parameters. Those parameters control the decision frontiers of the algorithm, which can be very complex. This leads to the following consequences:

- Since the behaviour of a ML component depends essentially on data, most inputs leads to the execution of the same code. Test coverage criteria based on the control structure of the implementation (e.g., coverage of each instruction, each decision, or each path…) do not bring that much information on the correctness of the *behaviour* ML component.
- Ensuring the coverage of each parameter is neither feasible nor useful since the behaviour of the ML component depends on the combined effect of all parameters.
- Exhaustive testing is usually impossible, and equivalence-classes on the inputs are extremely difficult to define (e.g. when the input is a set of pixels).

So, an appropriate testing strategy should provide

- A test vector selection process to explore "efficiently" the input space (both in time and in the gain in confidence).
- Test coverage metrics taking into consideration that the behaviour of ML algorithm is essentially determined by data.

In [146], testing is based on the study of the problem domain (e.g. size and ranges of input values, expected precision of floating point numbers), analyzing runtime options and their associated behaviour coherence to prevent bugs potentially caused by a misuse of the input options, and analyzing the algorithm itself with creation of synthetic data design.

A strategy of creating test images is also proposed in DEEP Xplore [147]. The authors describe how to optimize input images in order to make several networks to have a different behaviour (e.g. different classification for the same input image). These images are corner cases for the tested networks. In addition, a neuron coverage metric is proposed. This metric measures how many neurons in the network are significantly activated. The authors argue that this metric is more useful than code coverage. Indeed, code coverage is 100% with almost all input images then not representative of the network activation level. A good activation coverage rate aims to reduce the risk of unobserved behaviour of the algorithm in operation (e.g. network weights).

In [15], the authors adapt and mix well known Coverage-Guided Fuzzing and AFL testing procedures to test and debug neural networks. They point out why using standard CGF procedure in a naïve way will not perform good testing and they propose a new coverage definition, better suited to neural networks, based on the evaluation of the changes of the network state with respect to previous corpus data.

[148] approach consists in creating test images by applying "mutations" to some seed images. If this mutated image causes enough coverage modification it is added to the image corpus, in addition if it validates some objective function, it is added to the test set. The objective function represents the feared event like numerical errors or behaviour inconstancies introduced by weight quantization.

In addition, a *test oracle* is also needed to carry a test campaign. The oracle gives the *expected* output (the "ground truth") that is compared to the *actual* output produced by the system under test. This oracle must be automated if *massive* testing is to be performed. Here again, ML raises strong difficulties because such an oracle is hard to implement without relying on… machine learning. Another strategy may be to generate test scenarios where both the input and the output is known by construction, for instance using simulation.[52] But, then, the representativeness of the simulation must be questioned.

#### 4.8.2.2  Scenario testing

The previous paragraph presents methods to test the robustness of an ML component, but adequate testing strategies are also required to cover the "normal situations".

The ML algorithm learns its behaviour from a finite set of examples (see §4.5 for a discussion about these examples). As of today, control on what is actually learnt by the model can be considered from a certification point of view as too weak. Then, an intensive testing of the algorithm behaviour may be necessary (as a part of "Learning Assurance" set of tools [15]). The size of the input space, the complexity of the ML models implementation make mandatory the development of systematic (to explore every possible case) and precise (to provide guarantees on the testing procedure) testing strategies.

VERIFAI [149] is a toolkit for the design and analysis of artificial intelligence-based systems. It pertains to the massive testing strategy family. The system should be associated with a simulator that generates the environment embodiment. To facilitate the exploration of a large amount of scenarios a probabilistic language, SCENIC [150], is used to describe the objects and distributions over their parameters and behaviours in the environment. These distributions somehow contain some assertion about the environment.

---

[52] For instance, if system generates a synthetic image of a « stop sign », it knows *by construction* that it *is* a stop sign. No oracle is needed.

Approaches have been proposed to extend the first version of VERIFAI, in order to generate problematic scenarios (the one that violates the system and environment properties).

Commercial companies like NVIDIA[53] propose photorealistic scenario testing tools for autonomous vehicle.

*4.8.2.3 Cross platform validation*

As seen in the previous paragraphs, and in the absence of more efficient strategies, testing ML algorithm currently requires a huge computational power to execute "enough" scenarios, to cover the most problematic inputs for the algorithms, etc. Solutions may be found in cloud computing, Hardware In the Loop simulators, etc., but whatever the solution, a clear cross platform validation strategy must be developed.

In addition, great care must be taken when transferring the model (possibly developed on a desktop computer) to the target platform. The effects of the modifications made to cope with the limited processing and memory of embedded platforms such as, for instance, weight quantization,[54] must be carefully analysed.

### 4.8.3 Provability and testability in the development process

A ML-component will eventually be implemented by a piece of software and / or hardware, so (most of) the usual V&V objectives expressed in the standards (DO-178C, DO-254, ISO 26262, etc.) remain applicable. However, the presence of bias introduced to the design of an AI algorithm, or other faults introduced during the learning phase that would lead to an unacceptable behaviour of the software with respect to the intended function and the safety, must be detected by an appropriate strategy integrated in the ML system development process.

Therefore, we propose to map the provability and testability properties to a classical V-model (Figure 21). Note that organizational questions (e.g. who realize the tests?) are currently not to be addressed.

---

[53] https://www.nvidia.com/en-us/self-driving-cars/drive-constellation/
[54] It has been shown that quantization of a neural network can improve is generalization capability. This surprising effect illustrates the need of particular attention during cross platform validation.

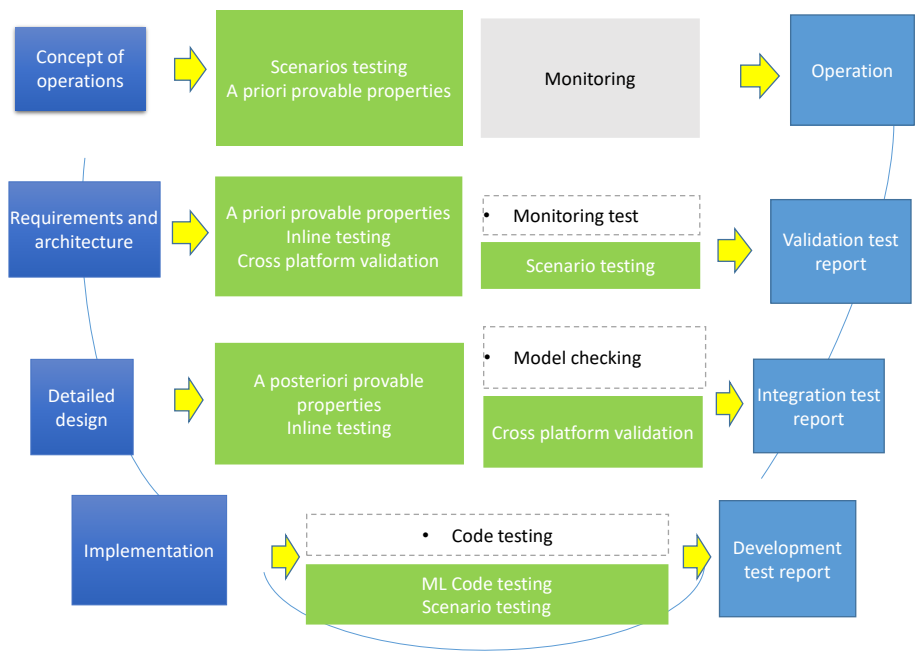

*Figure 21. Provability and testability mapped in the V-model (green boxes are ML-dedicated activities).*

#### 4.8.3.1 Concept of operations

A concept of operations document describes the characteristics of a system from the "user" point of view. The Operational Design Domain (ODD) is defined at this stage.

Assumptions about this domain must be studied and carefully integrated to the world representation used during scenario testing (massive testing). This work shall be performed by a team of domain experts, simulation experts and Machine Learning experts in order to ensure a high degree of accuracy and pertinence of the scenario testing procedure. The set of verification algorithms must be considered to show how confidence will be built in addition to the scenario testing tools.

During concept of operation definition, mandatory properties of the ML algorithms are also defined. A priori provability of these properties must be studied during this phase.

#### 4.8.3.2 Requirements and architecture

During the overall architecture design, a priori provable properties must again be carefully considered in order to define the architecture that will meet or enforce wishful/mandatory behaviour of the system.

Cross platform validation requirements must be specified during system architecture definition. In particular, it must be carefully verified that a priori properties, e.g., Lipschitzness, will actually hold on the final platform despite some changes on the floating-point numbers representation (e.g., from 64-bit to 16-bit).

Depending on the technical choice, monitoring requirements are studied with the rest of the system requirements. In reality, monitoring is a part of the system himself. It is not an exogenous property.

### 4.8.3.3 Detailed design

We propose to map *a posteriori* provable properties with the detailed design. This may appear artificial and one can consider that *a posteriori* provable properties can be useful to study during overall architecture design. That is true. In the detailed design phase, these properties can be used as nice to have properties to facilitate the certification process. For example, regularity properties can be shown to reinforce robustness of the algorithm. Then, regularity may not be required by to perform the function but will be very useful to certification. Including these points of attention during detailed design is highly valuable.

Monitoring techniques are also introduced at this stage.

### 4.8.3.4 Implementation

Low level tests must be designed and performed during the implementation stage. The constraint of testing on the final platform and on the scenario-testing platform may be stronger than before. Indeed, the consequences of the massive usage of floating-point computations must be observed carefully. ML code testing tools must be defined and used at this stage.

### 4.8.3.5 Integration test

For complex systems, Model Checking consists in verifying that a system (or a model thereof) satisfies a property by exploring all its reachable states [151]. Model checking are now routinely employed in verification of digital hardware. The high running cost of such general model checking remains a major challenge.

During integration test, cross platform validation tests complement the ML testing procedures.

### 4.8.3.6 Validation test

Traditionally, monitoring allows the detection of abnormal behaviours via the observation of specific observable variables quantities such as the mean response time of a function, the number of requests to some API, the update time of some variable, the data recovery time, etc. Associated with user experience feedback, we obtain in a way a continuous improvement of the model: a resilience (see §4.3). Automating this feedback speeds up the process.

Similarly, concerning ML deployment, validation goes beyond pure performance measurement. Several points must be checked like dedicated robustness checking (adversarial, OOD etc.). Explainability techniques can also be applied to validate on what information a model bases its decisions.

### 4.8.3.7 Discussion

The mapping proposed intends to help to identify where specific actions must be performed to adapt the "classical" V-Cycle to the new challenges of ML based critical systems. This is a first proposition and it will evolve with new tools/methods developments and will actual certification of ML based systems work.

In [151], the authors propose to put testing challenges in parallel with Complex Software challenges. Complex systems share with some ML algorithms the difficulty of being able to fully master the behaviour of the system by a human being. The authors present a set of methods like monitoring, model checking, data testing etc., some very close to the ones above, and deliver an optimistic message.

#### 4.8.4 Challenges for verifiability

For the verifiability theme, we identify the following challenges:

- Challenge #7.1: How to give a mathematical definition to the properties to be verified?

    Indeed, proofs can be given if the properties at hand are mathematically well defined. For example in the case of adversarial robustness, proof are given for particular kind of attacks or in the case of bounded perturbations [132].
- Challenge #7.2: How to make formal methods and tools applicable (and scalable) at an acceptable cost?
    For example, when convolutional networks become deep, Lipschitz bound becomes very difficult to evaluate accurately. Then the estimated bounds become uninformative. The same problem appears with Abstract Interpretation.
- Challenge #7.3: How to provide universal guidelines for the algorithm construction, e.g. network design to guarantee (or at least reinforce) mathematical properties by construction?
    For instance, we have described particular networks like monotonous function having by design useful properties. Unfortunately, there are still very few guidelines to construct neural networks showing such a property. Nowadays performance first research trend should evolve to property first.
- Challenge #7.4: How to define appropriate coverage metrics for the testing of ML components?
    Concerning testing, a universal difficulty is to develop a useful fault/error model of the system in consideration. This fault/error model can be organized in different levels. These levels must be revisited to cope with ML based systems specificity. The already long history of testing as provided a set of tools and practices that must be applied to ML based system as is. For example, unitary testing of each functions are mandatory to prevent wrong component behaviour (stack overflow) and to ensure that the mathematical properties of the algorithm hold in its implementation. However, intermediate level of tests must be defined to take into account that ML algorithms behaviour is massively driven by some data (e.g. weights of a neural network). These tests definition is still an open problem.
    Challenges that are even more specific appear with imprecise specifications, from a certification point of view, of the function to provide, and size of the parameters space. As developers use different exploration strategies: explicitly constructing difficult inputs or by massive exploration of the input domain to find wrong situations for the system, to give guarantees on validity in definition and coverage of the test is highly non-trivial.
- Challenge #7.5: How to take credit of tests done on simulations?
    When simulations are massively used for testing, at least at the early stages of development, domain gap between simulated world and real world must be fully handled.
- Challenge #7.6: How to specify and build test sets
    - To take into account that the behaviour of ML algorithms is essentially driven by the training data?

- - o   When the input space (and often decision space) are high dimensional?
  - o   When the decision boundaries of the algorithm are complex?
  - o   Considering the existence of corner cases?
- Challenge #7.7: When shall we stop testing? Does the size of the test dataset depend on the problem or the model size?

# 5  Conclusions

In this White Paper, we have presented the objectives of the ML Certification Workgroup set up at IRT Saint Exupéry in the context of the ANITI and DEEL projects. An overview and taxonomy of ML techniques are given, followed by the needs for ML in the different safety critical domains of the project partners (aeronautics, automotive and railway). We have proposed three types of analyses: a detailed analysis of a typical ML-system development process, a similarities analysis between ML techniques and techniques already implemented in certified systems, and a backward analysis to point out ML-techniques and applications that do not show *some* of the other techniques' challenges. The working group has identified a list of "High Level properties" (HLPs) that, if possessed by a ML technique, are considered to have a positive impact on the capability to certify the ML-based.

The proposed analysis and the HLPs led the working group to define and work on seven "main challenges" for ML certification:

1. Probabilistic assessment
2. Resilience
3. Specifiability
4. Data Quality and Representativeness
5. Explainability
6. Robustness
7. Verifiability

Each of these topics has been studied by the working group to identify challenges raised by ML with respect to the current certification practices with the aim to become concrete scientific objectives for the core team of the DEEL project.

The synthesis of all challenges is given hereafter:

- **Main challenge #1: Probabilistic assessment**
    - Challenge #1.1 Definition of environment / context / produced outputs / internal state. More precisely, how to identify all the possible events (not only failures)? How to estimate the probability of occurrence for each event? How to estimate the harm associated with each event?
    - Challenge #1.2: How to propose definitions of the risk incorporating more concepts (such as the estimation of harm effects) to express classical safety requirement such as "safety reserve" (safety margin, safety factor)?
    - Challenge #1.3: How to propose appropriate loss functions taking into account safety objectives?
    - Challenge #1.4: How to make the link between probabilities assessed on the datasets and on the real environment (to be used as safety indicators)? What are the possible means to translate the performance of the model (evaluated on a validation dataset) to the operational performance of the system (evaluated during its operation)?
    - Challenge #1.5: How to find tractable methods to assess the operation error w.r.t. the empirical error?

- **Main challenge #2: Resilience**
    - Challenge #2.1: How to detect erroneous or out-of-domain inputs of a ML model?
    - Challenge #2.2: On what system architectures can we rely to ensure the safe operations of ML–based systems?
        - Challenge #2.2.1: Can we rely on Runtime Assurance? How to monitor ML-based systems?
        - Challenge #2.2.2: Can we rely on multiple dissimilar ML-systems? And how to assess dissimilarity?
    - Challenge #2.3: How to create ML models "robust by design", to reduce the need for resilience requirement at system level?
    - Challenge #2.4: How to create a confidence index characterizing the proper functioning of the ML component?
    - Challenge #2.5: How to recover from an abnormal operating mode in an ML system?

- **Main challenge #3: Specifiability**
    - Challenge #3.1: How to identify the additional behaviours introduced during training in order to:
        - Complete the system specifications?
        - Assess the potential safety impact?
        - Accept or to reject those additions?
        - Assess robustness?
    - Challenge #3.2: What criteria could be used to close the iteration loop on system specification during training to take into account the behaviour that could have been added during this phase?

- **Main challenge #4: Data quality and representativeness**
    - Challenge #4.1: How can we prove that a dataset is representative of the Operational Design Domain for a given usage?
    - Challenge #4.2: How can the representativeness of a dataset be quantified, in particular in the case of unstructured data such as texts or images?
    - Challenge #4.3: Which of the training, validation and test datasets should possess the data quality and representativeness properties?

- **Main challenge #5: Explainability**
    - Challenge #5.1: How to ensure the interpretability of the information communicated to the operator? How to ensure that the operator is able to associate them with his/her own domain knowledge / mental model?
    - Challenge #5.2: How to provide the necessary and sufficient information to the operator regarding the actions he/she must execute in a given situation/timeframe?
    - Challenge #5.3: To what extent explainability is needed for certification?
    - Challenge #5.4: When is an explanation is acceptable? How to define explainability metrics to assess the level of confidence in an explanation?
    - Challenge #5.5: How to perform investigation on ML-based systems? How to explain what happened after an incident/accident?

- **Main challenge #6: Robustness**
    - Challenge #6.1: How to assess robustness of ML systems? How to measure similarities in high dimension (image)?
    - Challenge #6.2: How to detect abnormal or adversarial inputs?
    - Challenge #6.3: What kind of ML algorithm can enhance or guarantee robustness?
    - Challenge #6.4: How to identify and take into consideration corner cases?

- **Main challenge #7: Verifiability**

    Provability

    - Challenge #7.1: How to give a mathematical definition to the properties to be verified?
    - Challenge #7.2: How to make formal methods and tools applicable (and scalable) at an acceptable cost?
    - Challenge #7.3: How to provide universal guidelines for the algorithm construction, e.g. network design to guarantee (or at least reinforce) mathematical properties by construction?

    Testing

    - Challenge #7.4: How to define appropriate coverage metrics for the testing of ML components?
    - Challenge #7.5: How to take credit of tests done on simulations?
    - Challenge #7.6: How to specify and build test sets…
        - To take into account that the behaviour of ML algorithms is essentially driven by the training data?
        - When the input space (and often decision space) are high dimensional?
        - When the decision boundaries of the algorithm are complex?
        - Considering the existence of corner cases?
    - Challenge #7.7: When shall we stop testing? Does the size of the test dataset depend on the problem or the model size?

The DEEL project is now tackling part of these challenges, either from a scientific point of view within the Core team (e.g. Fairness, Robustness, Explainability challenges), or following a bottom-up approach towards industrial use cases (Acas-XU for collision avoidance, railway signals images classification) within the certification workgroup. For any further information please contact the heads of the workgroup (see page iii)

## Glossary

| | |
|---|---|
| AEB | Automatic Emergency Braking |
| AI | Artificial Intelligence |
| AVSI | Aerospace Vehicle Systems Institute |
| CAT | Catastrophic |
| CNN | Convolutional Neural Network |
| CSM | Common Safety Method |
| DAL | Design Assurance Level |
| DNN | Deep Neural Network |
| EASA | European Union Aviation Safety Agency |
| EUROCAE | European Organisation for Civil Aviation Equipment |
| FAA | Federal Aviation Administration |
| FC | Failure Condition |
| FE | Feared Event |
| FHA | Functional Hazard Assessments |
| FM | Formal verification Methods |
| FMEA | Failure Mode Effects Analysis |
| GAME | Globalement Au Moins Equivalent |
| GDPR | General Data Protection Regulation |
| HLP | High-Level Properties |
| KNN | K-Nearest Neighbours |
| ML | Machine Learning |
| MLE | Maximum Likelihood Estimation |
| MLM | Machine Learning Model |
| MLP | Multi Layer Perceptron |
| NN | Neural Network |
| NSE | No Safety Effect |
| OP | Overarching Properties |
| ReLU | Rectified Linear Unit |
| SIL | Safety Integrity Level |
| SOTIF | Safety Of The Intended Functionality |
| SVM | Support Vector Machine |
| TSI | Technical Specifications for Interoperability |
| UAV | Unmanned Aircraft Systems |
| WG | Working Group |